\documentclass[journal]{IEEEtran}
\usepackage{algorithm}
\usepackage{algorithmic}
\usepackage{amsmath}
\usepackage{amssymb}
\usepackage{array}
\usepackage{bm}
\usepackage{booktabs}
\usepackage{cases}
\usepackage[pdftex]{graphicx}
\usepackage{indentfirst}
\usepackage{mathrsfs}
\usepackage{multirow}
\usepackage{url}
\usepackage{xcolor}
\usepackage{makecell}
\usepackage{pifont}
\usepackage{subfig}

\newcommand{\eg}{\textit{e.g.},~}
\newcommand{\ie}{\textit{i.e.},~}
\newcommand{\etal}{\textit{et. al.}~}
\begin{document}
\title{Learning Adaptive Discriminative Correlation Filters via Temporal Consistency Preserving Spatial Feature Selection for Robust Visual Object Tracking}
\author{Tianyang~Xu,
        Zhen-Hua~Feng,~\IEEEmembership{Member,~IEEE}
        Xiao-Jun~Wu,
        and~Josef~Kittler,~\IEEEmembership{Life~Member,~IEEE}% <-this % stops a space
\thanks{T. Xu is with the School of Internet of Things Engineering, Jiangnan University, Wuxi, P.R. China and the Centre for Vision, Speech and Signal Processing, University of Surrey, Guildford, GU2 7XH, UK. (e-mail: tianyang.xu@surrey.ac.uk, tianyang\_xu@163.com)}
\thanks{Z.-H. Feng and J. Kittler are with the Centre for Vision, Speech and Signal Processing, University of Surrey, Guildford, GU2 7XH, UK. (e-mail: \{z.feng; j.kittler\}@surrey.ac.uk)}
\thanks{X.-J. Wu is with the School of Internet of Things Engineering, Jiangnan University, Wuxi, P.R. China. (e-mail: wu\_xiaojun@jiangnan.edu.cn)}% <-this % stops a space
\thanks{\copyright 2019 IEEE. Personal use of this material is permitted. Permission from IEEE must be obtained for all other uses, in any current or future media, including reprinting/republishing this material for advertising or promotional purposes, creating new collective works, for resale or redistribution to servers or lists, or reuse of any copyrighted
component of this work in other works.

DOI: 10.1109/TIP.2019.2919201}
}
\maketitle
\sloppy
\begin{abstract}
With efficient appearance learning models, Discriminative Correlation Filter (DCF) has been proven to be very successful in recent video object tracking benchmarks and competitions. However, the existing DCF paradigm suffers from two major issues, \ie spatial boundary effect and temporal filter degradation.
To mitigate these challenges, we propose a new DCF-based tracking method.
The key innovations of the proposed method include adaptive spatial feature selection and temporal consistent constraints, with which the new tracker enables joint spatial-temporal filter learning in a lower dimensional discriminative manifold. 
More specifically, we apply structured spatial sparsity constraints to multi-channel filers. Consequently, the process of learning spatial filters can be approximated by the lasso regularisation. 
To encourage temporal consistency, the filter model is restricted to lie around its historical value and updated locally to preserve the global structure in the manifold. Last, 
a unified optimisation framework is proposed to jointly select temporal consistency preserving spatial features and learn discriminative filters with the augmented Lagrangian method. 
Qualitative and quantitative evaluations have been conducted on a number of well-known benchmarking datasets such as OTB2013, OTB50, OTB100, Temple-Colour, UAV123 {\color{black}{and VOT2018}}. The experimental results demonstrate the superiority of the proposed method over the state-of-the-art approaches.
\end{abstract}
% \fi
\IEEEpeerreviewmaketitle
\section{Introduction}
Visual object tracking is an important research topic in computer vision, image understanding and pattern recognition.
Given the initial state (centre location and scale) of a target in {\color{black}{the first frame of}} a video sequence, the aim of visual object tracking is to automatically obtain the states of the object in the subsequent video frames.
With the rapid development of the research area during the past decades, a variety of tracking algorithms have been proposed and shown to deliver promising results~\cite{Yilmaz2006Object,Yang2011Recent,Wu2013Online,Smeulders2014Visual,Wu2015Object,Li2016NUS}. 
The advances opened a wide spectrum of applications in practical scenarios, such as intelligent surveillance, robot perception, medical image processing and other visual intelligence systems.
Despite the great success, robust and real-time visual object tracking remains a challenging task, especially in unconstrained scenarios in the presence of illumination variation, changing background, camera jitter, image blur, non-rigid deformation and partial occlusion.

Effective and reliable modelling of the appearance for a target and its surroundings is one of the most important keys to robust visual object tracking under unconstrained scenarios.
To this end, both generative and discriminative methods have been studied.
A generative method usually uses a parametric model, \eg the probability density function, to describe target appearance.
The most plausible candidate is selected as the tracking result by maximising its similarity to a generative model or minimising its {\color{black}{corresponding}} reconstruction error.
In contrast, a discriminative method exploits the background information to improve the representation capacity of an appearance model. 
Discriminative methods usually consider a tracking task as {\color{black}{a}} classification or regression problem hence directly {\color{black}{infers}} the output of a candidate by estimating the conditional probability distribution of labels for given inputs.
The optimal candidate with the highest response/score is selected as the tracking result.
Recently, considering the joint circular structure of sliding candidates, Discriminative Correlation Filter (DCF) based tracking methods~\cite{Bolme2010Visual,Henriques2015High} have achieved outstanding performance in many challenging benchmarks and competitions~\cite{Wu2015Object,Kristan2015The,Kristan2016The}. 
The main advantages of DCF include the effective use of circulant structure of original samples and the efficient formulation of the {\color{black}{learning task}} as ridge regression.
Besides, DCF employs the Fast Fourier Transform (FFT) to accelerate the computation of closed-form solutions for all circularly shifted candidates in the frequency domain.

Despite the effectiveness and efficiency of DCF, its performance is affected by two major issues: spatial boundary effect and temporal filter degradation. 
As the circular shift operation of an image patch results in discontinuity around original boundaries, such a boundary effect leads to {\color{black}{a}} spatial distortion, hence decreases the quality of {\color{black}{the}} training examples. 
{\color{black}{On the other hand}}, filter degradation reduces the {\color{black}{modelling}} effectiveness for lack of integrating historical appearance information. 
To eliminate these problems, we advocate the joint use of temporal information and spatial regularisation for adaptive target appearance modelling in a DCF-based framework.

The key to solving the first issue, \ie spatial boundary effect, is to enhance the spatial appearance model learning framework. 
{\color{black}{Generally}}, DCF-based tracking methods aim to optimise the correlation filter so that it is able to efficiently locate a target {\color{black}{in}} a search window.
To use the cyclic structure~\cite{Henriques2012Exploiting}, the DCF paradigm typically represents the target by extended padding from the target box to the entire search region.
As the correlation operator is defined by the inner product, features from different locations in the search region contribute to the final response.
{\color{black}{Therefore}}, a large search region contributes more clutter from {\color{black}{the background,}} while a small search region may lead to {\color{black}{a}} drift and target underdetection.
To address this issue, {\color{black}{different spatial regularisation techniques have been widely used~\cite{Galoogahi2015Correlation,Danelljan2015Learning,Lukezic2017Discriminative,Galoogahi2017Learning}}}.
However, existing spatial regularisation methods only regularise the filter with simple pre-defined constraints, {\color{black}{such as a pre-defined binary mask}}, without considering the diversity and redundancy of the entire feature input.

\begin{figure}[!t]
\begin{center}
   \includegraphics[trim={0mm 10mm 95mm 0mm},clip,width=1\linewidth]{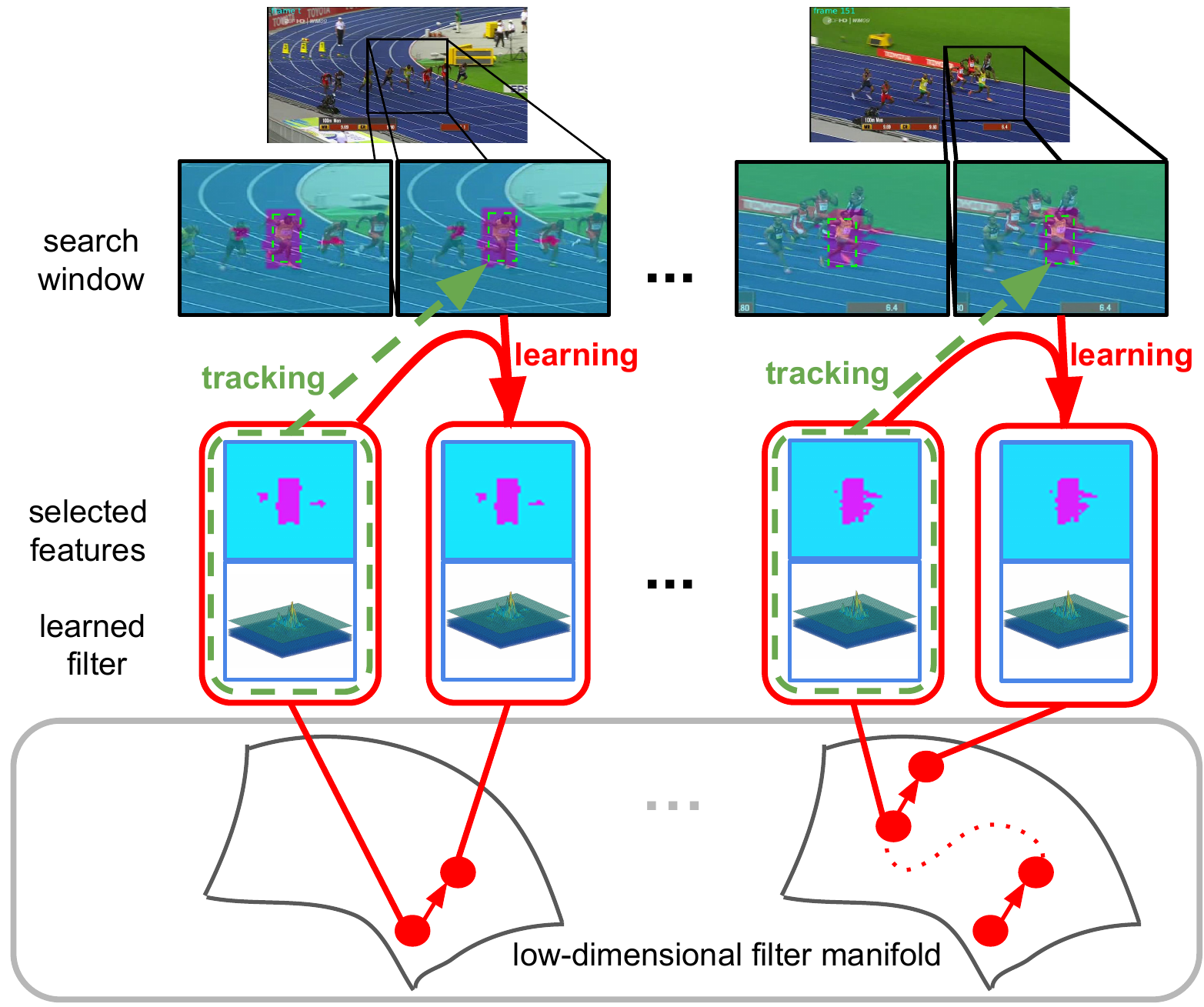}
\end{center}
   \caption{Illustration of the proposed method in sequence \textit{Bolt}. Only the selected spatial features of filters (annotated in the purple colour) are active in the tracking stage to achieve robust visual object tracking. Both sparse constraint and temporal consistency are jointly considered in the learning stage. {\color{black}{The spatial configurations (selected spatial features) are smoothly and robustly updated to form a low-dimensional discriminative manifold, enhancing the capability of the filtering system in dealing with appearance variations.}} }\label{illustration}
\end{figure}

In this work, a more informative spatial appearance leaning formulation is proposed to perform adaptive spatial feature selection and filter learning jointly.
Spatial features are regularised and selected with the {\color{black}{group}} lasso regularisation that adaptively preserves the structure of the discriminative manifold {\color{black}{encapsulating the variation of the target}} and its background.
With both, the centre and surrounding regions being exploited to support discrimination, spatial features with positive reference values are selected, including those from the background with {\color{black}{a}} similar and stable motion with respect to that of the target. 
As illustrated in Fig.~\ref{illustration}, our feature selection strategy enables compress sensing by adaptively selecting an optimal discriminative spatial mask, avoiding boundary distortion and restraining the impact of the distractive information inherent in the  original representations.

For the second issue, \ie temporal filter degradation, traditional DCF methods are vulnerable to the vagaries of extensive appearance variations of a target caused by the use of unstable temporal updating models for filters.
This instability is precipitated by two factors: i) single frame independent learning, and  ii) fixed-rate model update.
Although numerical visual features are integrated with appearance models, the existing tracking methods are not able to represent and store dynamic appearance of targets as well as surroundings.
{\color{black}{Instead}}, a single frame learning scheme is used in the traditional DCF paradigm to accelerate the optimisation process.

On the other hand, a fixed-rate model update of the moving average form, \ie $\theta=(1-\alpha)\theta+\alpha\theta'$, ignores the variation between different frames, hence reduces the adaptability to appearance variations.
In this work, temporal consistency is forced to interact with {\color{black}{the}} spatial feature selection in appearance modelling to generate a manifold structure capturing the global dynamic appearance information ({\color{black}{with stable spatial configurations}}, as shown in Fig.~\ref{illustration}). 
In particular, we propose an online temporal consistency preserving model to construct a generative manifold space that identifies an effective feature configuration and filtering function. 
Such an online adaptation strategy is capable of preventing filter degradation, thus {\color{black}{enhancing}} temporal smoothness.

Last, we present a unified optimisation framework that can efficiently perform spatial feature selection and discriminative filter learning.
To be more specific, the augmented Lagrangian method is used to optimise the variables in an iterative fashion.
We perform feature selection in the spatial domain and filter learning in the frequency domain.
The transformation between different domains is realised by FFT.
It should be highlighted that, for feature extraction, hand-crafted and deep neural network features can be used in conjunction with our method. 
{\color{black}{The experimental results obtained on a number of well-known benchmarks, including OTB2013~\cite{Wu2013Online}, OTB50~\cite{Wu2015Object}, OTB100~\cite{Wu2015Object}, Template-Colour~\cite{Liang2015Encoding}, UAV123~\cite{mueller2016benchmark} and VOT2018~\cite{kristan2018sixth},}}  demonstrate the efficiency and advantages of the proposed method in learning adaptive discriminative correlation filters (LADCF) via {\color{black}{the}} temporal consistency preserving spatial feature selection, over the state-of-the-art approaches.
The main contributions of the proposed LADCF method include:
\begin{itemize}
\item A new appearance model construction technique with adaptive spatial feature selection. In our experiments, only about $5\%$  hand-crafted and $20\%$ deep features are selected for filter learning, yet achieving better performance, {\color{black}{as compared to exploiting all the features}}. The proposed method also effectively addresses the issues of spatial boundary effect and background clutter.

\item A novel methodology for designing a low-dimensional discriminative manifold space by exploiting temporal consistency, which realises reliable and flexible temporal information compression, alleviating filter degradation and preserving appearance diversity.

\item A unified optimisation framework, proposed to achieve efficient filter learning and feature selection using the augmented Lagrangian method. 
%The experimental results on OTB2013, OTB50, OTB100, Temple-Colour and UAV123 demonstrate the superiority of our LADCF over state-of-the-art approaches.
\end{itemize}

%{\it Paper outline}
The rest of this paper is organised as follows. Section~\ref{related_work} discusses the prior work relevant to the proposed LADCF method. In Section~\ref{TD}, we introduce the classical DCF tracking formulation. The proposed temporal consistency preserving spatial feature selection method is introduced in Section~\ref{PA}. The experimental results are reported and analysed in Section~\ref{experiment}. {\color{black}{Conclusions are}} presented in Section~\ref{conclusion}.

\section{Related work}
\label{related_work}
For a comprehensive review of existing tracking methods the reader can refer to recent surveys~\cite{Yilmaz2006Object,Yang2011Recent,Wu2013Online,Smeulders2014Visual,Wu2015Object,Li2016NUS}. In this section, we focus on the most relevant techniques that define the baseline for the research presented in this paper. The review includes basic generative and discriminative tracking methods, DCF-based tracking methods and embedded feature selection methods.

\textbf{Generative methods}: Since the proposal of the Lucas-Kanade algorithm~\cite{Lucas1981An,Baker2004Lucas}, generative methods have become widely used techniques in many computer vision tasks, including visual object tracking. 
Subsequently, mean-shift~\cite{Comaniciu2002Real} was proposed for visual object tracking with iterative histogram matching. 
Accordingly, the spatial appearance information is summarised in the form of {\color{black}{a histogram}} for a target in {\color{black}{each candidate location}}, and the tracking task is achieved by minimising the Bhattacharyya distance.
To improve the robustness of visual object tracking with generative models, Adam \etal proposed fragments-based tracking by matching an ensemble of patches~\cite{Adam2006Robust}, in which the robustness of the use of histograms was improved by exploiting {\color{black}{aggregate}} appearance information conveyed by collaborative fragments. 
Later, the adoption of subspace-based tracking approaches offered a better explanation for the appearance model and provided an incremental learning method for appearance model update~\cite{Ross2008Incremental}. 
Recently, trackers based on sparse and low-rank representations had also achieved robust results by considering structural constraints and manifold information~\cite{Mei2009Robust,Ji2012Real,Jia2012Visual,Zhang2012Real,Zhang2013Robust}. Though generative models have achieved considerable success in constrained scenarios by faithfully modelling the target appearance, they are vulnerable to extensive appearance variations and unpredictable {\color{black}{target movements}}. 
Thus, more attention has been paid to discriminative approaches.

\textbf{Discriminative methods}: Unlike generative methods that focus on exploring the similarities between candidates and a target model, discriminative methods aim to construct a classifier or regressor to distinguish the target from its background. 
To this end, an online boosting tracker~\cite{Grabner2006Real} was proposed by Grabner \etal by fusing multiple weak classifiers.
In addition, a multi-instance learning tracker~\cite{Babenko2011Robust} was proposed to learn a discriminative classifier from the extracted positive and negative samples. 
To address temporal appearance variations of a target, the tracking-learning-detection method~\cite{Kalal2012Tracking} was proposed to handle short-term occlusion using a dual tracking strategy.
Moreover, in order to exploit the labelled and unlabelled data more effectively, Struck~\cite{Hare2016Struck} proposed to train a structural supervised classifier for visual object tracking.

More recently, taking the advantage of offline-learning and online-tracking deep models, end-to-end learning approaches have been introduced to visual object tracking with GPU acceleration. 
In this category, GOTURN~\cite{held2016learning} used consecutive sample pairs for regression model training and achieved efficient tracking results.
For robust comparison, SINT~\cite{tao2016siamese} proposed to consider the tracking task as a verification matching problem solved by a Siamese network. 
The same strategy was applied in SiamFC~\cite{bertinetto2016fully} and CFNet~\cite{valmadre2017end}. 
SiamFC established a fully-convolutional Siamese network by cross-correlating deep features, while CFNet learned {\color{black}{a}} correlation filter as a differentiable layer in deep architecture and achieved good results on standard benchmarks~\cite{Wu2015Object}. 
In addition, residual learning for spatial and temporal appearance was proposed by CREST~\cite{song-iccv17-CREST}, in which feature extraction, response calculation and model updating were fused in a single layer of a Convolutional Neural Network (CNN). 

\textbf{DCF-based tracking methods} have attracted wide attention recently. DCF employs circulant structure to solve a ridge regression problem in the frequency domain. 
Based on the proposals of Normalised Cross Correlation (NCC)~\cite{Briechle2001Template} and Minimum Output Sum of Squared Error (MOSSE) filter~\cite{Bolme2010Visual}, Henriques \emph{et al.} improved MOSSE by introducing circulant structure~\cite{Henriques2012Exploiting}, which enabled efficient calculation of filter learning with element-wise operations. 
Other improvements in DCF focus on exploring robust feature representations, scale detection and spatial regularisation.

For feature representation, contextual feature information was exploited in~\cite{Zhang2014Fast} to achieve spatial-temporal learning. 
Colour names~\cite{Weijer2009Learning} were fused with the correlation filter framework by Danelljan~\etal\cite{Danelljan2014Adaptive} to better represent an object.
Staple~\cite{Bertinetto2016Staple} employed colour histograms from foreground and background to generate a response map, which improved the reliability of the final response. Later, CNNs have been used to provide better feature representation of an object, as in deepSRDCF~\cite{Danelljan2015Learning}. 

For scale detection, SAMF~\cite{li2014scale} and DSST~\cite{danelljan2014accurate} were proposed to handle scale variations by performing scale selection in a scale pool after the tracking stage. 
On the other hand, fDSST~\cite{danelljan2017discriminative} proposed to perform scale detection in the tracking stage. This improves the efficiency by joint scale and location estimation. 

For spatial regularisation, a pre-defined filter weighting strategy was proposed in SRDCF~\cite{Danelljan2015Learning}, concentrating the filter energy in the central region of a search window.
In CSRDCF~\cite{Lukezic2017Discriminative}, the filter was equipped with a colour histogram based segmentation mask, with which only the discriminative target region was activated. 
A similar approach was employed in CFLB~\cite{Galoogahi2015Correlation} and BACF~\cite{Galoogahi2017Learning}, forcing the parameters corresponding to background to be exactly zero.  
In addition, DCF-based tracking methods have also been extended to support long-term memory~\cite{Ma2015Long}, multi-kernel method~\cite{tang2015multi}, structural constraints~\cite{Liu2016Structural}, support vector representation~\cite{Zuo2016learning}, sparse representation~\cite{Zhang2016In} and enhanced robustness~\cite{Liu2015Real,Bibi2016Target,Sui2016Real,Hong2015MUlti,chen2018convolutional,liu2018robust}.
Furthermore, adaptive decontamination of the training set~\cite{Danelljan2016Adaptive} was proposed to achieve adaptive multi-frame learning in the DCF paradigm, which improved the generalisation performance. Danelljan \emph{et al.} proposed sub-grid tracking by learning continuous convolution operators (C-COT)~\cite{Danelljan2016Beyond}. 
Efficient Convolution Operators (ECO)~\cite{Danelljan2016ECO} were proposed to achieve a light-weight version of C-COT with a generative sample space and dimension reduction mechanism. 

\textbf{Embedded feature selection} has been proven to be very effective in processing high-dimensional data. Thus it has been a widely studied topic in many pattern recognition and computer vision applications, \eg image classification~\cite{Dadkhahi2016Masking} and image compression~\cite{Barni2001Improved}. 
Embedded feature selection methods are regularisation models with an optimisation objective that simultaneously minimises classification (regression) errors and forces the variables to satisfy specific prior properties, \eg lying in a $\ell_p$-norm ball around $0$~\cite{James2013An,gui2017feature}. These approaches enhance model generalisation by reducing over-fitting with the advantage of interpretability. 

Basic DCF-based trackers~\cite{Henriques2012Exploiting,Henriques2015High} utilise $\ell_2$-norm to regularise the coefficients. 
Such a penalty shrinks all the filter coefficients towards zero, but it does not set any of them exactly to zero. 
A weighted $\ell_2$-norm regularisation was exploited in SRDCF, C-COT and ECO, in which the coefficients in the background shrink more than those in the target region. 
Besides, {\color{black}{masking strategy}} has been applied in the DCF paradigm by CFLB, CSRDCF and BACF. 
Only the coefficients in the target region are activated in CFLB and BACF.
CSRDCF utilised a two-stage feature selection method to first pre-define the selected region by discriminative colour information and then train the mask-constrained filters. 
In addition to the $\ell_2$-norm regularisation, lasso regularisation, based on the $\ell_1$-norm, is a popular method to achieve embedded feature selection in classification (regression).
It has been widely explored in sparse representation based tracking approaches~\cite{zhang2015robust,Jia2012Visual,Zhang2013Robust}.
{\color{black}{Lasso regularisation}} is able to {\color{black}{achieve}} sparse estimation with only a small number of features activated. 
However, the basic lasso regularisation methods ignore the structure information as they assume the variables are independent.
To this end, structured feature selection methods~\cite{yuan2006model,nie2010efficient} have been proposed to integrate group knowledge, improving the accuracy and robustness.

Our method employs the DCF paradigm to formulate the tracking problem, with {\color{black}{an appearance model constructed using an embedded, temporal consistency-preserving spatial feature selection mechanism. }}
The proposed DCF learning scheme and spatial feature selection achieve efficient discriminative filter learning.
Adaptive spatial features are activated by lasso regularisation, that can be efficiently optimised by iterative threshold shrinkage. 
Last, {\color{black}{a temporal consistency constraint is imposed on the dynamic appearance model}}  to enhance the robustness of the selected features over time.

\section{Tracking Formulation}
\label{TD}
{\color{black}{Consider}} an $n\times n$ image patch $\bm{x}\in\mathbb{R}^{n^2}$ {\color{black}{as a base sample for the DCF design.}} The circulant matrix for this sample is generated by collecting its full cyclic
shifts, $\bm{X}^\top=\left[\bm{x}_1, \bm{x}_2, \ldots, \bm{x}_{n^2}\right]
^\top \in\mathbb{R}^{n^2\times n^2}$ with the corresponding gaussian shaped regression labels $\bm{y}=\left[y_1,y_2,\ldots,y_{n^2}\right]$~\cite{Henriques2015High}. Our goal is to learn a discriminative function $f\left(\bm{x}_i;\bm{\theta}\right)=\bm{\theta}^\top\bm{x}_i$ to distinguish the target from background, where $\bm{\theta}$ denotes the target model in the form of DCF.
{\color{black}{This type of}} data augmentation method is widely used in the DCF paradigm with the calculation convenience in the frequency domain:
\begin{equation}\label{fx}
f\left(\bm{X};\bm{\theta}\right)=\bm{\theta}^\top\bm{X}
=\bm{\theta}\circledast\bm{x}=\mathcal{F}^{-1}\left(\hat{\bm{\theta}}\odot\hat{\bm{x}}^\ast\right),
\end{equation}
where $\mathcal{F}^{-1}$ denotes the Inverse Discrete Fourier Transform (IDFT), and $\hat{\bm{x}}^\ast$ is the complex conjugate of $\hat{\bm{x}}$ in the frequency domain. $\hat{\bm{x}}$ is the Fourier representation of $\bm{x}$, \ie~$\mathcal{F}\left(\bm{x}\right)=\hat{\bm{x}}$. $\circledast$ denotes the circular convolution operator~\cite{Henriques2012Exploiting} and $\odot$ denotes the element-wise multiplication operator. It should be noted that the computational complexity of $f\left(\bm{X}\right)$ is significantly decreased from $\mathcal{O}\left(n^4\right)$ to $\mathcal{O}\left(n^2\log n\right)$ with the help of convolution theorems. We use the following tracking-learning-updating framework to formulate our tracking method.

\noindent \textbf{Tracking:} Given the model parameter $\bm{\theta}_{\textrm{model}}$ estimated from the previous frame, we aim to find the optimal candidate that maximises the discriminative function in the current frame $\bm{I}$:
\begin{equation}\label{tra} \bm{x}_\ast=\arg\underset{\bm{x}_i}{\max}f\left(\bm{x}_i;\bm{\theta}_{\textrm{model}}\right),
\end{equation}
where the candidates are generated by the circulant structure of a base sample $\bm{x}$, which is the image patch centred around the tracking result in the previous frame. Consequently, the {\color{black}{results can efficiently be calculated }}in the frequency domain.

\noindent \textbf{Learning:} After the tracking stage, a new model is trained by minimising the regularised loss function:
\begin{equation}
\bm{\theta}_\ast = \arg\underset{\bm{\theta}}{\min}
\mathcal{E}\left(\bm{\theta},\bm{\mathcal{D}}\right)+ \mathcal{R}\left(\bm{\theta}\right),
\label{le}
\end{equation}
where $\mathcal{E}()$ is the objective and $\mathcal{R}()$ is the regularisation term. 
$\bm{\mathcal{D}} = \left(\bm{X},\bm{y}\right)$ represents the labelled training samples generated by the circulant matrix with the base sample $\bm{x}$ centred around the tracking result in the current frame. 
In the traditional DCF paradigm, the quadratic loss and $\ell_2$-norm penalty are used in the learning stage to form a ridge regression problem, \ie  $\mathcal{E}\left(\bm{\theta}, \bm{\mathcal{D}}\right)=\|{\bm{\theta}}^\top \bm{X}-\bm{y}\|_2^2$ and $\mathcal{R}\left(\bm{\theta}\right)=\|\bm{\theta}\|_2^2$~\cite{Henriques2015High}. 

\noindent \textbf{Updating:} {\color{black}{Considering potential variations in the target appearance}}, an incremental model update strategy~\cite{Henriques2012Exploiting} is used in DCF: 
\begin{equation}
\bm{\theta}_{\textrm{model}} =(1-\alpha)\bm{\theta}_{\textrm{model}}+\alpha\bm{\theta}_\ast,
\label{LU}
\end{equation}
where $\alpha \in [0,1]$ {\color{black}{controls the trade-off}} between the current and historical information.

\section{The Proposed LADCF Algorithm}
\label{PA}
In this section, we first present our temporal consistency preserving spatial feature selection method for appearance modelling using single-channel features. Then we extend the method to multi-channel features.
The optimisation process is designed using the Alternating Direction Method of Multipliers (ADMM). 
Last, we depict the proposed LADCF tracking algorithm in more details.

\subsection{Temporal Consistency Preserving Spatial Feature Selection Model}\label{FS}
\begin{figure}[t]
\begin{center}
   \includegraphics[trim={0mm 5mm 95mm 10mm},clip,width=0.65\linewidth]{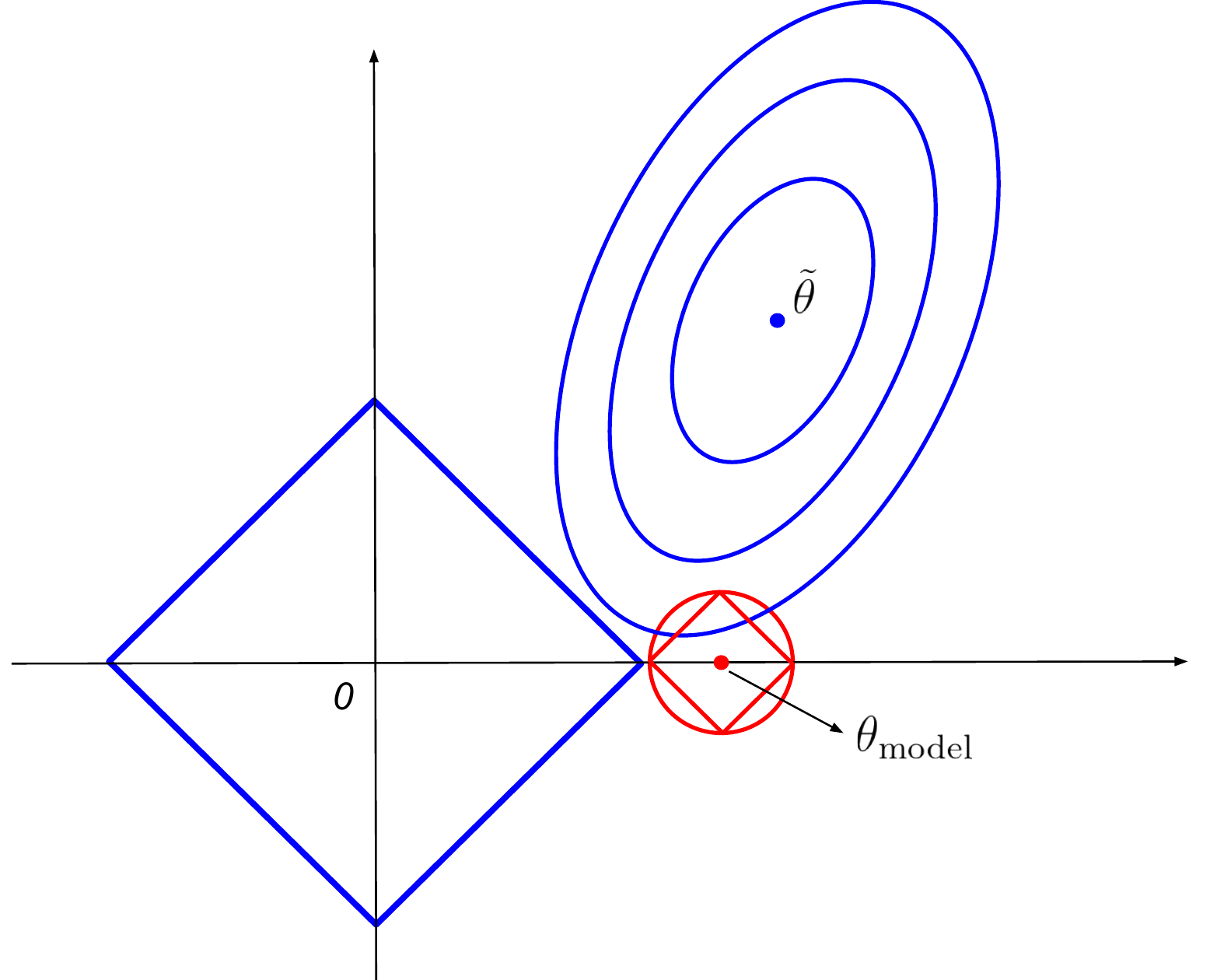}
\end{center}
   \caption{Illustration of the $\ell_2$-norm relaxation for the temporal consistency constraint: $\tilde{\bm{\theta}}$ is the optimal least-square solution and $\bm{\theta}_{\textrm{model}}$ is the template point. As $\bm{\theta}_{\textrm{model}}$ is sparse, its $\ell_2$-norm ball shares the same property, guiding the estimate to lie on a low-dimensional manifold. {\color{black}{Specifically, a larger $\lambda_2$ {\color{black}{restricts $\bm{\theta}$ to be close to $\bm{\theta}_{\textrm{model}}$.}}}}}\label{l1l2}
\end{figure}

Our feature selection process aims at selecting several specific elements in the filter $\bm{\theta}\in\mathbb{R}^{n^2}$ to preserve discriminative and {\color{black}{descriptive information}}. It is formulated as:
\begin{equation}
\bm{\theta}_{\bm{\phi}}=\textbf{diag}\left(\bm{\phi}\right)\bm{\theta},
\end{equation}
where $\textbf{diag}\left(\bm{\phi}\right)$ is the diagonal matrix generated from the indicator vector of selected features $\bm{\phi}$. Unlike traditional dimensionality reduction methods, such as the Principal Component Analysis (PCA) and Locally Linear Embedding (LLE), the indicator vector $\bm{\phi}$ enables dimensionality reduction as well as spatial structure preservation. The elements in $\bm{\phi}$ are either 0 or 1, disabling or enabling the corresponding element. Our feature selection enhanced filter design simultaneously selects spatial features and learns discriminative filters. It should be noted that the selected spatial features are implicitly shared by input $\bm{x}$, $\bm{\theta}_{\bm{\phi}}^\top\bm{x}=\bm{\theta}_{\bm{\phi}}^\top\bm{x}_{\bm{\phi}}$, which reveals that only the relevant features are activated for each training sample, forming a low-dimensional and compact feature representation. Thus, the spatial feature selection embedded learning stage can be formulated as:
\begin{equation}
\label{obj}
\begin{aligned}
\arg&\underset{\bm{\theta},\bm{\phi}}{\min}\left\|
\bm{\theta}\circledast {\bm{x}}-\bm{y}\right\|_2^2+\lambda_1\left\|\bm{\phi}\right\|_0\\
&s.t.~\bm{\theta}=\bm{\theta}_{\bm{\phi}}=\textbf{diag}\left(\bm{\phi}\right)\bm{\theta},
\end{aligned}
\end{equation}
where the indicator vector $\bm{\phi}$ can potentially be represented by $\bm{\theta}$ and $\left\|\bm{\phi}\right\|_0=\left\|\bm{\theta}\right\|_0$. As $\ell_0$-norm is non-convex, its convex envelope $\ell_1$-norm is widely used to approximate the sparsity~\cite{tibshirani1996regression}. On the other hand, in order to emphasise temporal consistency during tracking, the indicator vectors from successive video frames are assumed to be on a low-dimensional manifold. More specifically, we propose to restrict our estimate to lie in a $\ell_0$-norm ball around the current template, \ie $\left\|\bm{\theta}-\bm{\theta}_{\textrm{model}}\right\|_0<t$. 
{\color{black}{Such a temporal consistency constraint enables the selected spatial features to be  changed locally}}, preserving the discriminative layout. We formulate the temporal consistency preserving spatial feature selection model via $\ell_1$-norm relaxation as:
\begin{equation}
\arg\underset{\bm{\theta}}{\min}\left\|
\bm{\theta}\circledast{\bm{x}}-\bm{y}\right\|_2^2+\lambda_1\left\|\bm{\theta}\right\|_1+\lambda_2\left\|\bm{\theta}-\bm{\theta}_{\textrm{model}}\right\|_1,
\end{equation}
where $\lambda_1$ and $\lambda_2$ are tuning parameters and $\lambda_1<<\lambda_2$. As shown in Fig.~\ref{l1l2}, an intuitive explanation is that we impose a stronger constraint on temporal consistency than on spatial feature sparsity. In addition, the temporal consistency term promotes the sparsity of $\bm{\theta}$ by enhancing the similarity between the estimate $\bm{\theta}$ and the template $\bm{\theta}_{\textrm{model}}$. Note that as $\left\|\bm{\theta}-\bm{\theta}_{\textrm{model}}\right\|_2\leq\left\|\bm{\theta}-\bm{\theta}_{\textrm{model}}\right\|_1\leq n\left\|\bm{\theta}-\bm{\theta}_{\textrm{model}}\right\|_2$ and $\lambda_1<<\lambda_2$, we propose $\ell_2$-norm relaxation for the temporal consistency term to further simplify the objective as:
\begin{equation}
\label{obj1}
%\begin{aligned}
\arg\underset{\bm{\theta}}{\min}\left\|
\bm{\theta}\circledast{\bm{x}}-\bm{y}\right\|_2^2+\lambda_1\left\|\bm{\theta}\right\|_1+\lambda_2\left\|\bm{\theta}-\bm{\theta}_{\textrm{model}}\right\|_2^2,
%%\bm{\theta}=\textbf{diag}\left(\bm{\phi}\right)\bm{\theta}\\&
%\|\bm{\phi}\|_0 = M\end{aligned}
%\right.
%\end{aligned}
\end{equation}
where the spatial features $\bm{\phi}$ are selected by lasso regularisation controlled by $\lambda_1$. As lasso cannot control the number of non-zero entries in $\bm{\theta}$, 
we define this number by forcing $\|\bm{\phi}\|_0 = M$. The filter $\bm{\theta}$ optimised by this objective function can adaptively {\color{black}{highlight the spatial configuration so as to achieve sparsity and discrimination. }}
Specific spatial features corresponding to the target and background regions can simultaneously be activated to form a robust pattern. 
In addition, {\color{black}{since discriminative learning depends heavily on the reliability of supervision, the quality of the training samples is of paramount importance  to tracking performance.}}
Therefore, we promote temporal consistency by imposing smooth variation between consecutive frames with the help of the filter template $\bm{\theta}_{\bm{\textrm{model}}}$ in Eqn.~(\ref{obj1}). {\color{black}{In this way the diversity of dynamic and static appearance can be extracted and preserved by our temporal consistency preserving spatial feature selection.}}

\subsection{Generalising to Multi-channel Features}
\label{mul}
Multi-channel features, \eg HOG~\cite{Henriques2015High}, Colour-Names~\cite{Weijer2009Learning} and deep neural network features~\cite{Danelljan2016ECO}, have been widely used in visual object tracking. To this end, the traditional DCF paradigm \cite{Henriques2015High} explores each channel independently with equal weight. In contrast, CSRDCF~\cite{Lukezic2017Discriminative} proposes a weighting strategy for each channel at the decision stage. ECO~\cite{Danelljan2016ECO} applies a projection matrix to realise channel compression. 

In our approach, we aim to characterise our appearance model from single-channel signals to multi-channel feature representations. As multi-channel features share the same spatial layout, $\bm{\phi}$,  a group structure is considered in our multi-channel learning model~\cite{yuan2006model,gui2017feature}. 
We denote the multi-channel input as $\bm{\mathcal{X}}=\left\{\bm{x}_1,\bm{x}_2,\ldots,\bm{x}_L\right\}$ and the corresponding filters as $\bm{\mathit{\theta}}=\left\{\bm{\theta}_1,\bm{\theta}_2,\ldots,\bm{\theta}_L\right\}$, where $L$ is the number of channels.
For the $i$th channel, $\bm{x}_i=\left[x_i^1,x_i^2,\ldots,x_i^{D^2}\right]^\top\in\mathbb{R}^{D^2\times 1}$, the spatial size of the feature map is $D\times D$, $D^2$ is the number of spatial features, and $x_i^j$ is the element corresponding to the $j$th spatial feature in the $i$th channel. As the index vector $\bm{\phi}$ is applied to all the channels to configure a global spatial layout, the objective function in Eqn.~(\ref{obj1}) can be extended to multi-channel features with structured sparsity by minimising:
\begin{align}
\label{obj2}
%\arg\underset{\bm{\mathit{\theta}}}{\min}
h\left(\bm{\mathit{\theta}}\right)=
&
\sum\limits_{i=1}^{L}\left\|
\bm{\theta}_i\circledast{\bm{x}}_i-\bm{y}\right\|_2^2
+
\lambda_1 \sum_{j=1}^{D^2}{\sqrt{\sum_{i=1}^{L} \left(\theta_{i}^{j}\right)^2}}
\nonumber \\
&
+\lambda_2\sum\limits_{i=1}^{L}\left\|\bm{\theta}_i-\bm{\theta}_{\bm{\textrm{model}}~i}\right\|_2^2,
\end{align}
where $\theta_{i}^{j}$ is the $j$th element of the $i$th channel feature vector $\bm{\theta}_{i} \in \mathbb{R}^{D^2}$, the structured spatial feature selection term (the second term) calculates the $\ell_2$-norm of each spatial location and then performs $\ell_1$-norm to realise joint sparsity. Such group sparsity enables robust feature selection that reflects the joint contribution of feature maps at a single spatial location across all the feature channels. In addition, different channels are potentially weighted by structured sparsity, unifying the entire input during feature selection adaptively. 

\subsection{Optimisation}\label{opt}
In order to optimise Eqn.~(\ref{obj2}), we introduce slack variables to construct the following objective based on convex optimisation~\cite{bach2012structured}:
\begin{align}\label{obj3}
arg\underset{\bm{\mathit{\theta}},\bm{\mathit{\theta}}^\prime}{\min}
& 
\sum\limits_{i=1}^{L}\left\|
\bm{\theta}_i\circledast{\bm{x}}_i-\bm{y}\right\|_2^2+\lambda_1 \sum_{j=1}^{D^2}{\sqrt{\sum_{i=1}^{L} \left(\theta_{i}^{j}\right)^2}}
\nonumber \\
& +\lambda_2\sum\limits_{i=1}^{L}\left\|\bm{\theta}_i-\bm{\theta}_{\bm{\textrm{model}}~i}\right\|_2^2,
\nonumber \\
&s.t.~ \bm{\mathit{\theta}}=\bm{\mathit{\theta}}^\prime.
\end{align}
Exploiting augmented Lagrange multipliers to combine the equality constraint into the criterion function, our objective can be formulated to minimise the following Lagrange function:
\begin{align}
\label{obj4}
\mathcal{L}&=\sum\limits_{i=1}^{L}\left\|
\bm{\theta}_i\circledast{\bm{x}}_i-\bm{y}\right\|_2^2+\lambda_1 \sum_{j=1}^{D^2}{\sqrt{\sum_{i=1}^{L} \left(\theta_{i}^{j}\right)^2}}
\nonumber \\
&+\lambda_2\sum\limits_{i=1}^{L}\left\|\bm{\theta}_i-\bm{\theta}_{\bm{\textrm{model}}~i}\right\|_2^2 + \frac{\mu}{2}\sum\limits_{i=1}^{L}\left\|\bm{\theta}_i-\bm{\theta}^\prime_i+\frac{\bm{\eta}_i}{\mu}\right\|_2^2,
\end{align}
where $\bm{\mathcal{H}}=\left\{\bm{\eta}_1,\bm{\eta}_2,\ldots,\bm{\eta}_L\right\}$ are the Lagrange multipliers and $\mu>0$ {\color{black}{is the corresponding penalty parameter controlling the convergence rate~\cite{bertsekas1982constrained}. As $\mathcal{L}$ is convex, ADMM is exploited iteratively to optimise the following sub-problems with guaranteed convergence~\cite{Boyd2011Distributed}:
\begin{equation}
\label{obj5}
\left\{\begin{aligned}
&\bm{\mathit{\theta}}=\arg\underset{\bm{\mathit{\theta}}}{\min}~\mathcal{L}\left(\bm{\mathit{\theta}},\bm{\mathit{\theta}}^\prime,\bm{\mathcal{H}},\mu\right)\\
&\bm{\mathit{\theta}}^\prime=\arg\underset{\bm{\mathit{\theta}}^\prime}{\min}~\mathcal{L}\left(\bm{\mathit{\theta}},\bm{\mathit{\theta}}^\prime,\bm{\mathcal{H}},\mu\right) \\
&\bm{\mathcal{H}}=\arg\underset{\bm{\mathcal{H}}}{\min}~\mathcal{L}\left(\bm{\mathit{\theta}},\bm{\mathit{\theta}}^\prime,\bm{\mathcal{H}},\mu\right)
\end{aligned}\right. .
\end{equation}}}
\noindent \textbf{Updating} $\bm{\mathit{\theta}}$. Given $\bm{\mathit{\theta}}^\prime$, $\bm{\mathcal{H}}$ and $\mu$, optimising $\bm{\mathit{\theta}}$ is similar to the DCF learning scheme. We utilise the circulant structure and Parseval's formula to transform it into the frequency domain~\cite{Henriques2015High}, which requires solving the following convex optimisation problem for each channel independently:
\begin{align}
\label{sub1}
\hat{\bm{\theta}}_i =
&
\arg\underset{\hat{\bm{\theta}}_i}{\min}\left\|
\hat{\bm{\theta}}_i^\ast\odot {\hat{\bm{x}}}_i-\hat{\bm{y}}\right\|_2^2
\nonumber \\
& + \lambda_2\left\|\hat{\bm{\theta}}_i-\hat{\bm{\theta}}_{\bm{\textrm{model}}~i}\right\|_2^2+\frac{\mu}{2}\left\|\hat{\bm{\theta}}_i-\hat{\bm{\theta}}^\prime_i+\frac{\hat{\bm{\eta}}_i}{\mu}\right\|_2^2
\end{align}
that admits a closed-form optimal solution for $\hat{\bm{\theta}}_i$:
\begin{equation}
\label{sub1s}
\hat{\bm{\theta}}_i=\frac{\hat{\bm{x}}_i\odot\hat{\bm{y}}^\ast+\lambda_2\hat{\bm{\theta}}_{\bm{\textrm{model}}~i}+\frac{1}{2}\mu\hat{\bm{\theta}}^\prime_i-\frac{1}{2}\hat{\bm{\eta}}_i}
{\hat{\bm{x}}_i\odot\hat{\bm{x}}_i^\ast+\lambda_2+\frac{1}{2}\mu}.
\end{equation}
Note that the division operation in Eqn.~(\ref{sub1s}) is element-wise, as the element $\hat{\theta}_i^j$ is only determined by $\hat{x}_i^j$, $\hat{y}^j$, $\hat{\theta}^j_{\bm{\textrm{model}}~i}$, ${\hat{\theta}}_i^{\prime j}$ and $\hat{\eta}_i^j$ with parameters $\lambda_2$ and $\mu$. 
%Such calculation convenience is utilised for each channel to obtain the solution for sub-problem of updating $\bm{\mathit{\theta}}$.

\noindent \textbf{Updating} $\bm{\mathit{\theta}}^\prime$. Given $\bm{\mathit{\theta}}$, $\bm{\mathcal{H}}$ and $\mu$, optimising $\bm{\mathit{\theta}}^\prime$ involves solving the following optimisation problem:
\begin{equation}
\label{sub2}
\bm{\mathit{\theta}}^\prime =
\arg\underset{\bm{\mathit{\theta}}^\prime}{\min}\lambda_1 \sum_{j=1}^{D^2}{\sqrt{\sum_{i=1}^{L} \left(\theta_{i}^{j}\right)^2}}
 +\frac{\mu}{2}\sum\limits_{i=1}^{L}\left\|\bm{\theta}_i-\bm{\theta}^\prime_i+\frac{\bm{\eta}_i}{\mu}\right\|_2^2.
\end{equation}
We rewrite the objective by changing the summation index from channels to spatial features, 
\begin{equation}
\label{sub2'}
\bm{\mathit{\theta}}^\prime =
\arg\underset{\bm{\mathit{\theta}}^\prime}{\min}\lambda_1\sum\limits_{j=1}^{D^2}\left\|\bm{\theta}^{\prime j}\right\|_2
 +\frac{\mu}{2}\sum\limits_{j=1}^{D^2}\left\|\bm{\theta}^j-\bm{\theta}^{\prime j}+\frac{\bm{\eta}^j}{\mu}\right\|_2^2,
\end{equation}
{\color{black}{which can be separated for each spatial feature,
\begin{equation}
\label{sub2''}
\bm{\mathit{\theta}}^{\prime j} =
\arg\underset{\bm{\mathit{\theta}}^{\prime j}}{\min}\lambda_1\left\|\bm{\theta}^{\prime j}\right\|_2
 +\frac{\mu}{2}\left\|\bm{\theta}^j-\bm{\theta}^{\prime j}+\frac{\bm{\eta}^j}{\mu}\right\|_2^2.
\end{equation}
Setting the derivative of Eqn.~(\ref{sub2''}) to zero, we obtain $\frac{\lambda\bm{\mathit{\theta}}^{\prime j}}{\mu\|\bm{\mathit{\theta}}^{\prime j}\|_2}+\bm{\mathit{\theta}}^{\prime j}=\bm{g}^j$, where $\bm{g}^j=\bm{\theta}^j+\frac{\bm{\eta}^j}{\mu}$.
Therefore, $\bm{\mathit{\theta}}^{\prime j}$ and $\bm{g}^j$ share the same vector direction, indicating the exchangeability between the unit vectors $\frac{\bm{\mathit{\theta}}^{\prime j}}{\|\bm{\mathit{\theta}}^{\prime j}\|_2}$ and $\frac{\bm{g}^j}{\|\bm{g}^j\|_2}$.}}
The closed-form optimal solution can directly be calculated as~\cite{yang2011l2}:
\begin{equation}
\label{sub2s}
\bm{\theta}^{\prime j}=\max\left(0,1-\frac{\lambda_1}{\mu\left\|\bm{g}^j\right\|_2}\right)\bm{g}^j
\end{equation}
It is clear that  $\bm{\theta}^{\prime j}$ tends to shrink to zero by collaboratively integrating the constraints imposed by all the feature channels.

\noindent \textbf{Updating} $\bm{\mathcal{H}}$. Given $\bm{\mathit{\theta}}$, $\bm{\mathit{\theta}}^\prime$ and $\mu$, $\bm{\mathcal{H}}$ can be updated by:
\begin{equation}
\label{sub3s}
\bm{\mathcal{H}} = \bm{\mathcal{H}} + \mu\left(\bm{\mathit{\theta}}-\bm{\mathit{\theta}}^\prime\right).
\end{equation}
{\color{black}{We follow the standard ADMM~\cite{Boyd2010Distributed} with varying penalty parameter $\mu$, which is updated after each iteration as, $\mu = \min\left(\rho\mu,\mu_{\max}\right)$. $\rho>1$ enables the acceleration of $\mu$, improving the convergence, as well as making the performance less dependent on the initial choice of the penalty parameter. $\mu_{\max}$ avoids the choice of excessive values.
In addition, we restrict the maximum number of iterations in the practical implementation to $K$.}}

\noindent \textbf{Complexity}. The sub-problem of $\bm{\mathit{\theta}}$ requires FFT and inverse FFT in each iteration, which can be solved in $\mathcal{O}\left(LD^2\log\left(D\right)\right)$. The remaining element-wise operations can be solved in $\mathcal{O}\left(1\right)$ each. The total complexity of our optimisation framework is $\mathcal{O}\left(KLD^2\log\left(D\right)\right)$. 

\begin{algorithm}[t]
\begin{algorithmic}
\vspace{0.03in}
\STATE
\textbf{Input}
Image frame $\bm{I}_t$, filter model $\bm{\theta}_{\bm{\textrm{model}}}$, target center coordinate $p_{t-1}$ and scale size $w\times h$ from frame $t-1$;

\textbf{Tracking:}

\STATE Extract search windows with $S$ scales from $\bm{I}_t$ at $p_{t-1}$;

\STATE Obtain corresponding feature representations $\left[\bm{\mathcal{X}}\left(s\right)\right]_{s=1}^S$;

\STATE Calculate response scores $\bm{f}$ using Eqn.~(\ref{res});

\STATE Set $p_t$ and scale size $w\times h$ using Eqn.~(\ref{posscale});

\textbf{Learning:}

\STATE Obtain multi-channel feature representations $\bm{\mathcal{X}}$ based on current target bounding box;

\STATE Optimise $\bm{\theta}$ using Eqn.~(\ref{obj5}) for $K$ iterations;

\textbf{Updating:}

\STATE Update filter model $\bm{\theta}_{\bm{\textrm{model}}}$ using Eqn.~(\ref{updatetheta});

\STATE
\textbf{Output} Target bounding box (centre coordinate $p_t$ and current scale size $w\times h$). Updated filter model $\bm{\theta}_{\bm{\textrm{model}}}$ for frame $t$.
\vspace{0.03in}
\end{algorithmic}
\caption{LADCF tracking algorithm.}
\label{alg}
\end{algorithm}

\subsection{Tracking Framework}
We propose the LADCF tracker based on learning adaptive  discriminative correlation filters incorporating temporal consistency-preserving spatial feature selection. The tracking framework is summarised in Algorithm~\ref{alg}. 

\noindent \textbf{Position and scale detection}. We follow fDSST~\cite{danelljan2017discriminative} to achieve target position and scale detection simultaneously. When the new frame $\bm{I}_{t}$ becomes available, we extract a search window set $\left[\bm{I}_t^{\textrm{patch}}\left\{s\right\}\right]$ with multiple scales, $s=1,2,\ldots,S$, with  $S$ denoting the number of scales. For each scale $s$, the search window patch is centred around the target centre position $p_{t-1}$ with a size of $a^N n\times a^N n$ pixels, where $a$ is the scale factor and $N=\lfloor\frac{2s-S-1}{2}\rfloor$. We resize each patch to $n\times n$ pixels with bilinear interpolation, where $n\times n$ is the basic search window size which is determined by the target size $w\times h$ and padding parameter $\varrho$ as: $n=\left(1+\varrho\right)\sqrt{w\times h}$. Then we extract multi-channel features for each scale search window as $\bm{\mathcal{X}}\left(s\right)\in\mathbb{R}^{D^2\times L}$. Given the filter template $\bm{\mathit{\theta}}_{\bm{\textrm{model}}}$, the response scores can efficiently be calculated in the frequency domain as:
\begin{equation}\label{res}
\hat{\bm{f}}\left(s\right)=\hat{\bm{x}}\left(s\right)\odot\hat{\bm{\theta}}^\ast_{\bm{\textrm{model}}}.
\end{equation}
Having performed IDFT of $\hat{\bm{f}}\left(s\right)$ for each scale, the relative position and scale are obtained by using the maximum value of $\bm{f}\in\mathbb{R}^{D^2\times S}$ as $\bm{f}\left(\Delta p,s^\ast\right)$. Then the resulting target bounding box parameters (centre $p_t$ and scale size $w\times h$) are set as:
\begin{equation}\label{posscale}
\left\{
\begin{aligned}
&p_{t}=p_{t-1}+\frac{n}{D}\Delta p\\
&w=a^{\lfloor\frac{2s^\ast-S-1}{2}\rfloor}w\\
&h=a^{\lfloor\frac{2s^\ast-S-1}{2}\rfloor}h
\end{aligned}\right. .
\end{equation}

\noindent \textbf{Learning, updating and initialisation}. It should be noted that in the learning stage, the multi-channel input $\bm{\mathcal{X}}$ in Eqn.~(\ref{obj2}) forms the feature representation of the padded image patch centred at $p_{t}$ with size $n\times n$. We make use of our multi-channel temporal consistency preserving spatial feature selection embedded  appearance learning and optimisation framework as depicted in subsection~\ref{mul} and subsection~\ref{opt} to obtain the filter $\bm{\mathit{\theta}}$. {\color{black}{In order to control the number of selected spatial features, we sort the $\ell_2$-norm of each spatial feature vector $\left\|\bm{\theta}^j\right\|_2$ in the descending order and only preserve the largest $M$ vectors.}} We adopt the same updating strategy as the traditional DCF method~\cite{Henriques2012Exploiting}:
\begin{equation}\label{updatetheta}
\bm{\theta}_{\bm{\textrm{model}}} =(1-\alpha)\bm{\theta}_{\bm{\textrm{model}}}+\alpha\bm{\theta},
\end{equation}
where $\alpha$ is the updating rate. More specifically, as $\bm{\theta}_{\bm{\textrm{model}}}$ is not available in the learning stage for the first frame, we use a pre-defined mask with only the target region activated to optimise $\bm{\theta}$ as in BACF~\cite{Galoogahi2017Learning}, and then initialise $\bm{\theta}_{\bm{\textrm{model}}}=\bm{\theta}$ after the learning stage of the first frame.

\section{Performance Evaluation}\label{experiment}
We perform qualitative and quantitative experimental evaluations to validate our method. In this section, we first describe the implementation details, including the feature extraction methods and the parameter settings used in our LADCF method. We then introduce the benchmarking datasets and evaluation metrics used in our experiments, as well as the state-of-the-art trackers used for comparison. We analyse the experimental results on different datasets and discuss the advantages of our method. In addition, an analysis of different components and parameters of our LADCF method is carried out to explore the sensitivity and effect of specific parameters. 

\subsection{Implementation Details}
\noindent \textbf{Feature representations:} We use both hand-crafted and deep features in our method. 
It has been demonstrated that robust feature representation plays the most essential role in high-performance visual object tracking~\cite{wang2015understanding,gundogdu2018good}. 
{\color{black}{We equip the proposed LADCF method with only hand-crafted features to compare with the trackers not using deep features and construct LADCF$^\ast$ with both hand-crafted and deep features to facilitate a fair comparison with the trackers using deep features. Table~\ref{fr} shows the detailed setting of the features used for the evaluation. }}
For CNN, we use the middle (Conv-3) convolutional layer of the VGG network powered by the MatConvNet toolbox\footnote{\url{http://www.vlfeat.org/matconvnet/}}~\cite{vedaldi2015matconvnet}.

\begin{table}[!t]
\footnotesize
\renewcommand{\arraystretch}{1.1}
\caption{Feature representations for the proposed method.}
\label{fr}
\centering
\begin{tabular}{lccc}
\hline
Feature Type    &  \multicolumn{2}{c}{Hand-crafted}  & Deep\\
\hline
Feature & HOG & Colour-Names & CNN\\
Channels & 31 & 10 & 512\\
Cell Size & $4\times 4$ & $4\times 4$ & $16\times 16$\\
\hline 
LADCF & \ding{51}  &\ding{51}  & \ding{55}\\
 LADCF$^\ast$ &\ding{51}&\ding{51}  & {\ding{51}}\\
\hline
\end{tabular}
\end{table}

\begin{figure*}[!t]
\centering
\subfloat[Hand-crafted features]{
\label{otb100hc}
\includegraphics[width=0.24\linewidth]{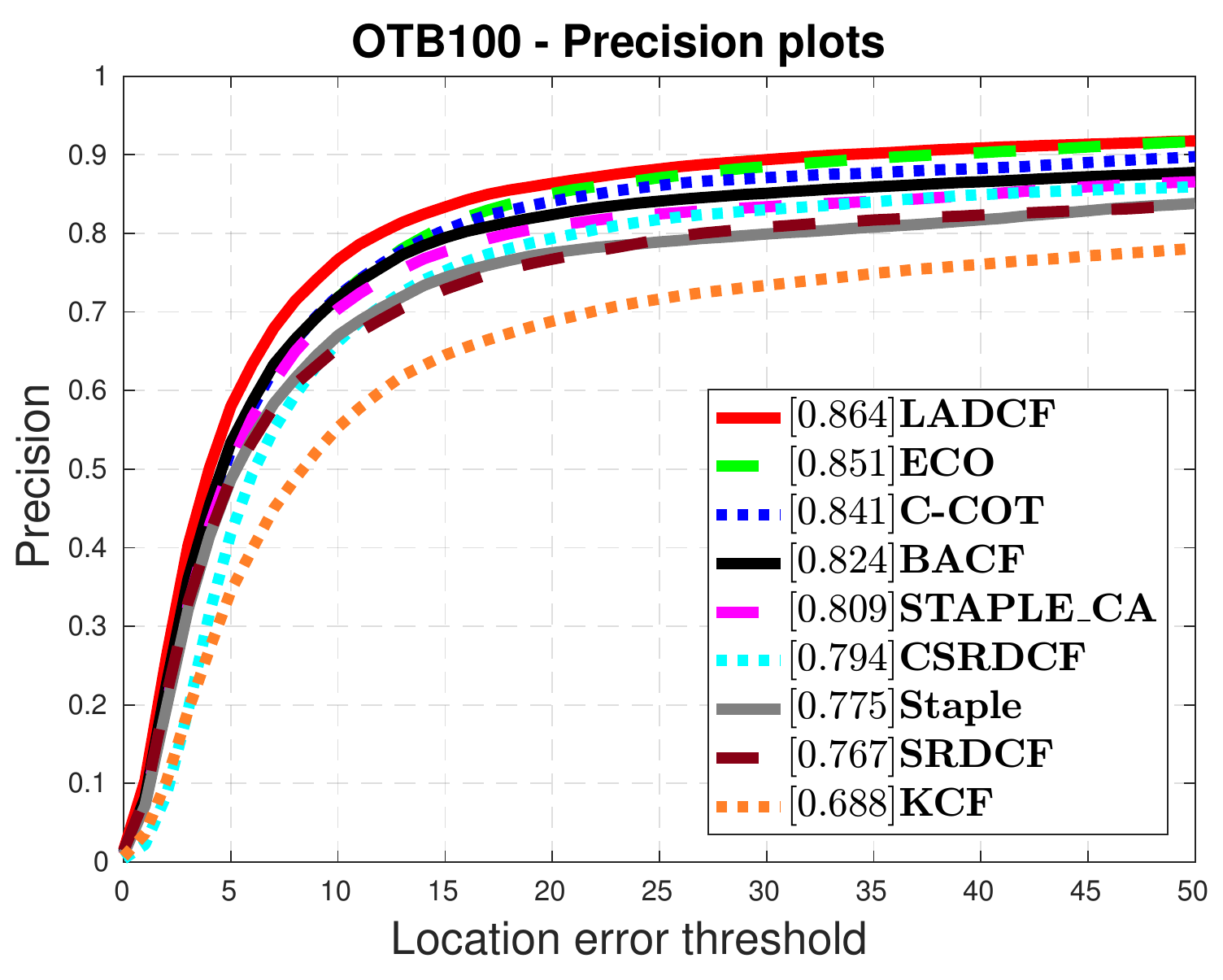}
\includegraphics[width=0.24\linewidth]{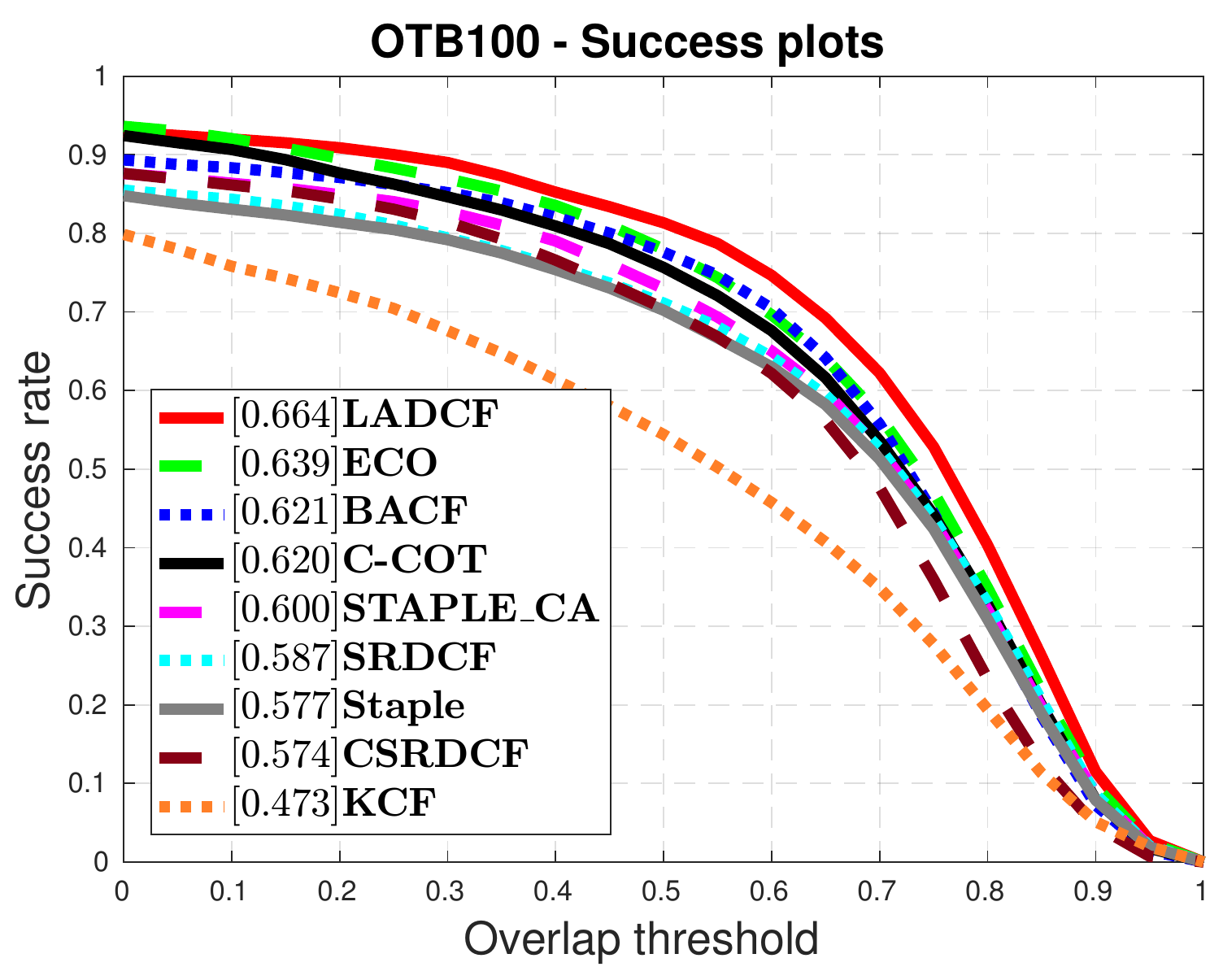}
}
\subfloat[Deep features]{
\label{otb100deep}
\includegraphics[width=0.24\linewidth]{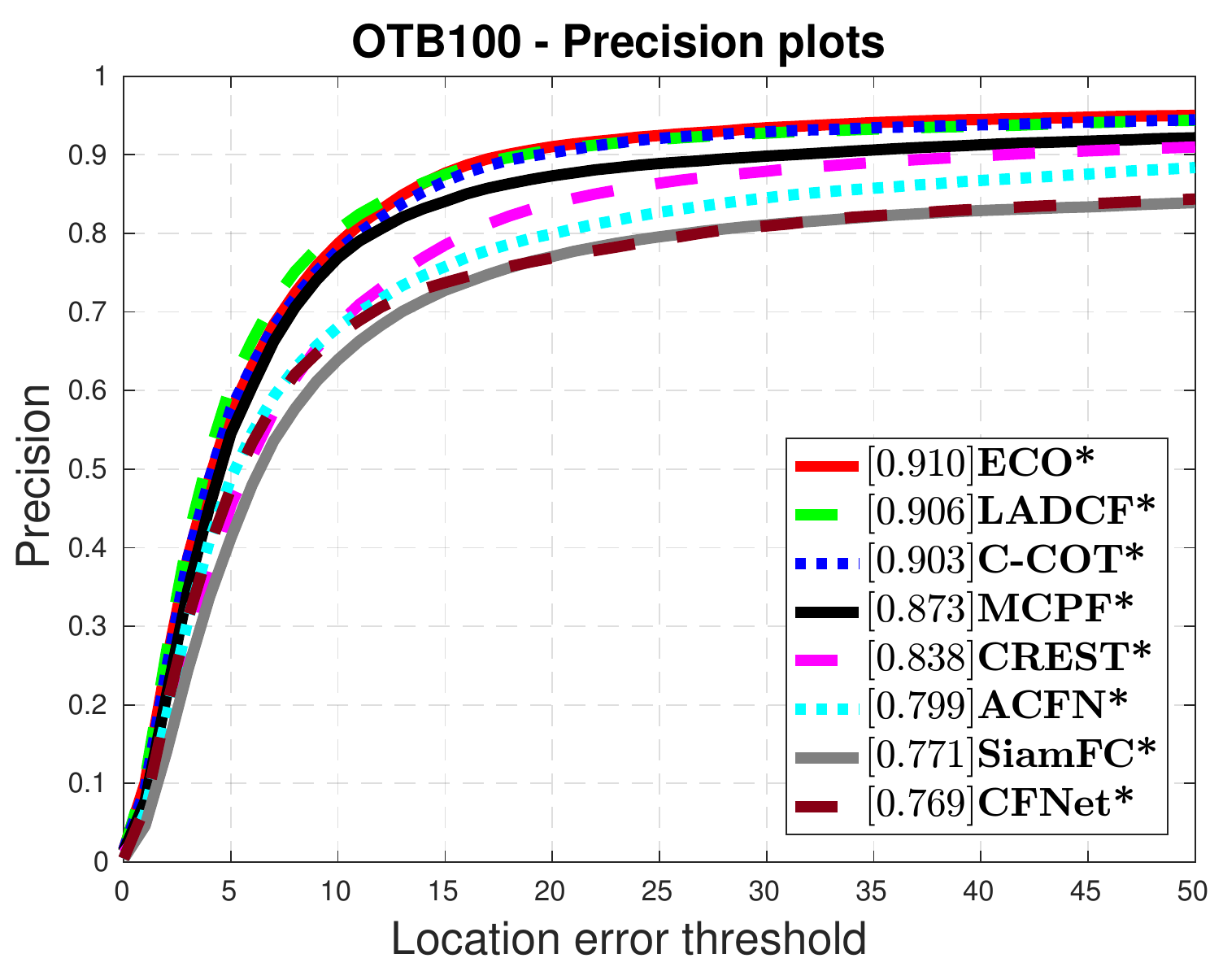}
\includegraphics[width=0.24\linewidth]{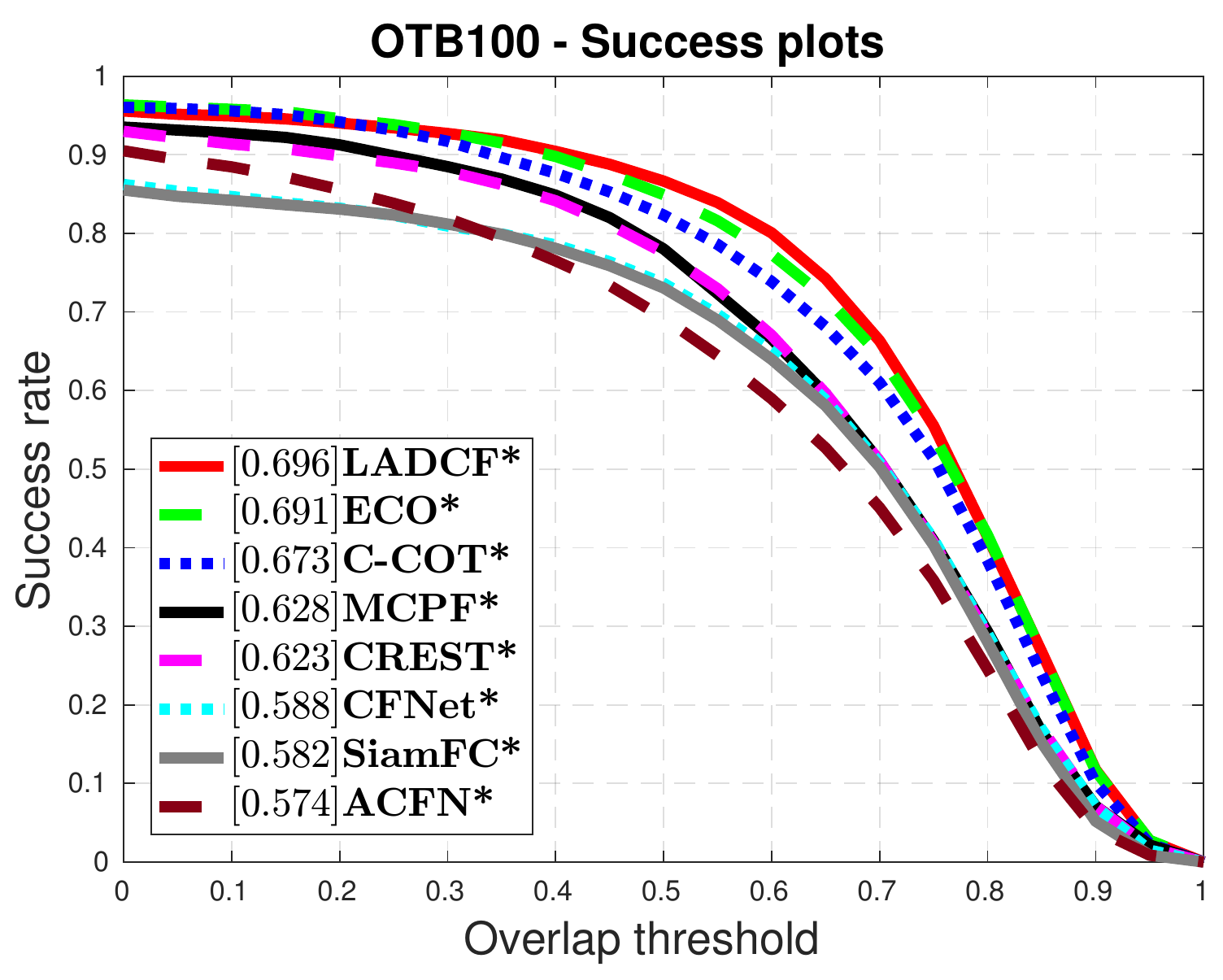}
}
\caption{A comparison of the proposed LADCF method with the state-of-the-art trackers on OTB100 using (a) hand-crafted features and (b) deep features. The results are evaluated using the precision and success plots.}
\label{otb_100_ladcf}
\end{figure*}

\begin{table*}[!t]
\footnotesize
\renewcommand{\arraystretch}{1.1}
\caption{A comparison of the proposed LADCF method with the state-of-the-art trackers using hand-crafted features, evaluated on OTB2013, OTB50 and OTB100 in terms of OP. The best three results are highlighted in {\color{red}{red}}, {\color{blue}{blue}} and {\color{brown}{brown}}.}
\label{otbophc}
\centering
\begin{tabular}{clccccccccc}
\hline
 & & KCF & CSRDCF & Staple & STAPLE\_CA & SRDCF & BACF & ECO & C-COT & LADCF\\
\hline
\multirow{ 3}{*}{Mean OP ($\%$)}  &OTB2013 &  60.8 & 74.4 & 73.8 & 77.6 & 76.0 & {\color{blue}{\textbf{84.0}}} & {\color{brown}{\textbf{82.4}}} & 78.9 & {\color{red}{\textbf{85.0}}}\\
&OTB50    & 47.7 & 66.4 & 66.5 & 68.1 & 66.1 & 70.9 & {\color{blue}{\textbf{73.4}}} & {\color{brown}{\textbf{72.3}}} & {\color{red}{\textbf{77.5}}}\\
 &OTB100   & 54.4 & 70.5 & 70.2 & 73.0 & 71.1 & {\color{brown}{\textbf{77.6}}} & {\color{blue}{\textbf{78.0}}} & 75.7 & {\color{red}{\textbf{81.3}}}\\
\hline
 Mean FPS ($\%$) on CPU & OTB100& 92.7 & 4.6 & 23.8 & 20.1 & 2.7 & 16.3 & 15.1 & 1.8 & 18.2\\
\hline
\end{tabular}
\end{table*}

\begin{table*}[!t]
\footnotesize
\renewcommand{\arraystretch}{1.1}
%\setcaptionwidth{0.86\linewidth}
\caption{A comparison of the proposed LADCF method with the state-of-the-art trackers using deep features, evaluated on OTB2013, OTB50 and OTB100 in terms of OP. The best three results are highlighted in {\color{red}{red}}, {\color{blue}{blue}} and {\color{brown}{brown}}.}
\label{otbopdeep}
\centering
\begin{tabular}{clcccccccc}
\hline
&  & CFNet$^\ast$ & SiamFC$^\ast$ & ACFN$^\ast$ & CREST$^\ast$ & MCPF$^\ast$ & ECO$^\ast$ & C-COT$^\ast$ & LADCF$^\ast$\\
\hline
\multirow{ 3}{*}{Mean OP ($\%$)}  &OTB2013 &  78.3 & 77.9 & 75.0 & {\color{brown}{\textbf{86.0}}} & 85.8 & {\color{blue}{\textbf{88.7}}} & 83.7 & {\color{red}{\textbf{90.7}}}\\
 &OTB50   & 68.8 & 68.0 & 63.2 & 68.8 & 69.0 & {\color{blue}{\textbf{81.0}}} & {\color{brown}{\textbf{80.9}}} & {\color{red}{\textbf{82.5}}}\\
& OTB100  & 73.6 & 73.0 & 69.2 & 77.6 & 78.0 & {\color{blue}{\textbf{84.9}}} & {\color{brown}{\textbf{82.3}}} & {\color{red}{\textbf{86.7}}}\\
\hline
 Mean FPS ($\%$) on CPU & \multirow{ 2}{*}{OTB100}& 1.4 & - & - & - & 0.5 & 1.1 & 0.2 & 1.3 \\
 Mean FPS ($\%$) on GPU & & 8.7 & 12.6 & 13.8 & 10.1 & 3.2 & 8.5 & 2.2 & 10.8 \\
\hline
\end{tabular}
\end{table*}

\begin{figure*}[!t]
\begin{center}
   \includegraphics[width=0.245\linewidth]{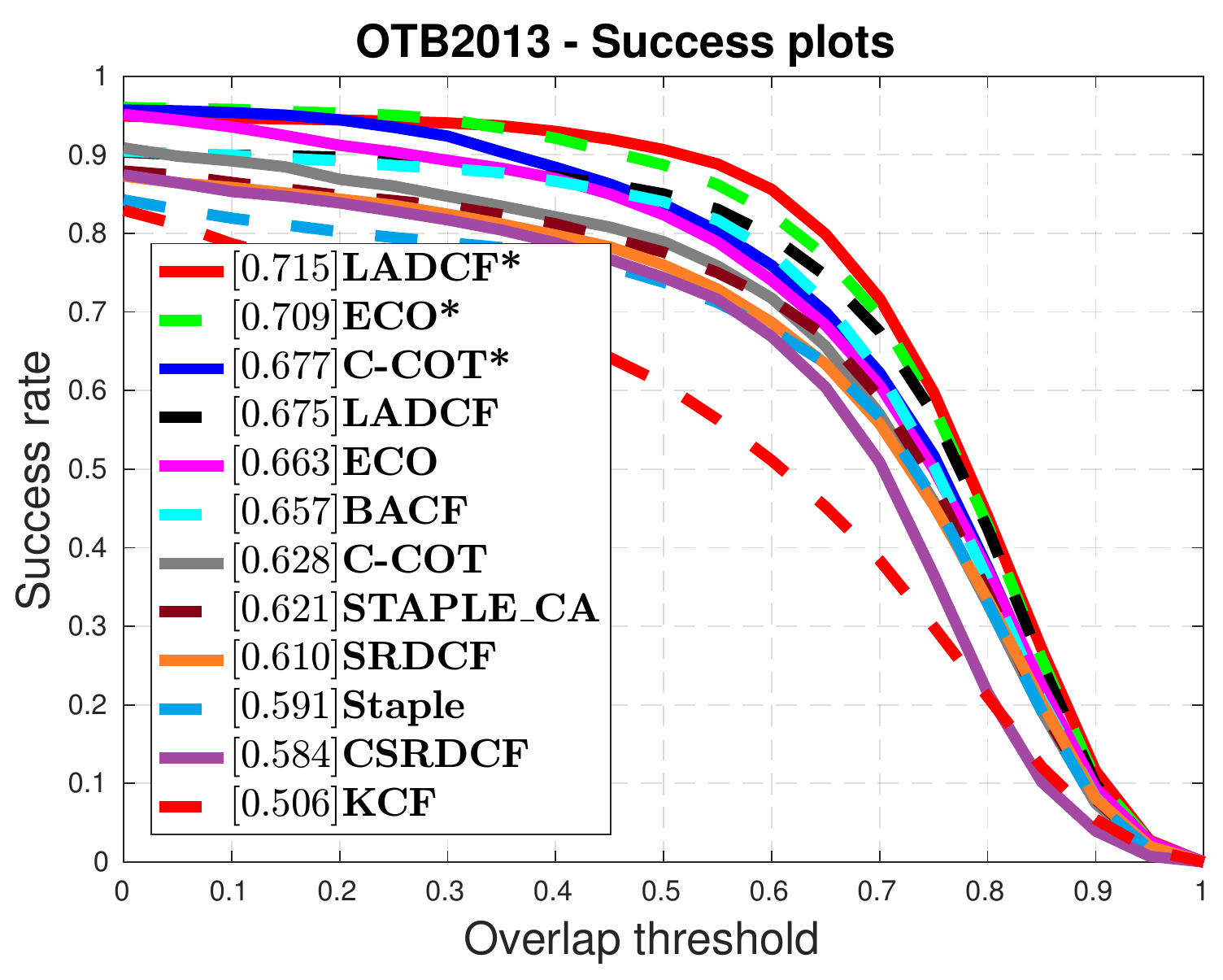}
   \includegraphics[width=0.245\linewidth]{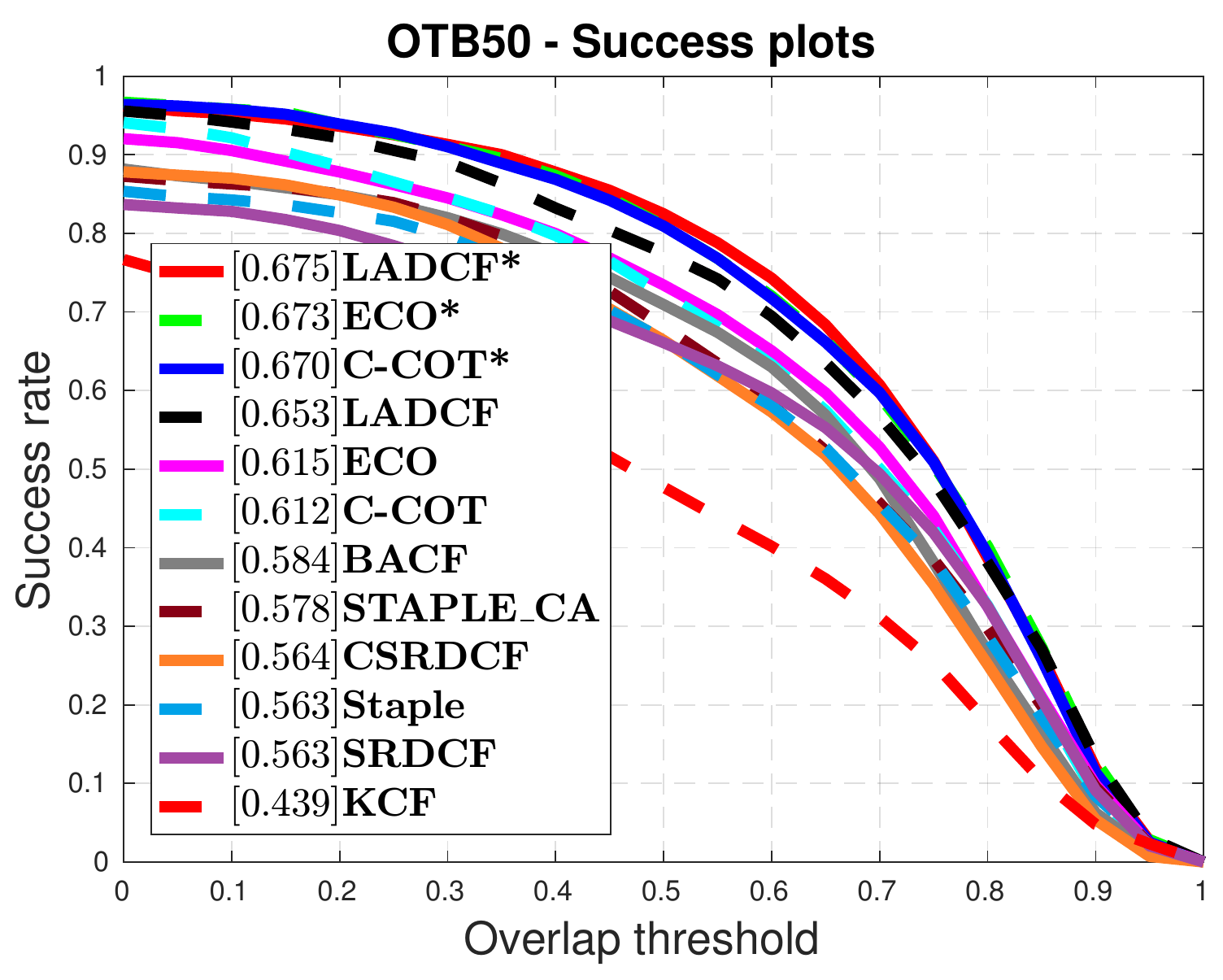}
   \includegraphics[width=0.245\linewidth]{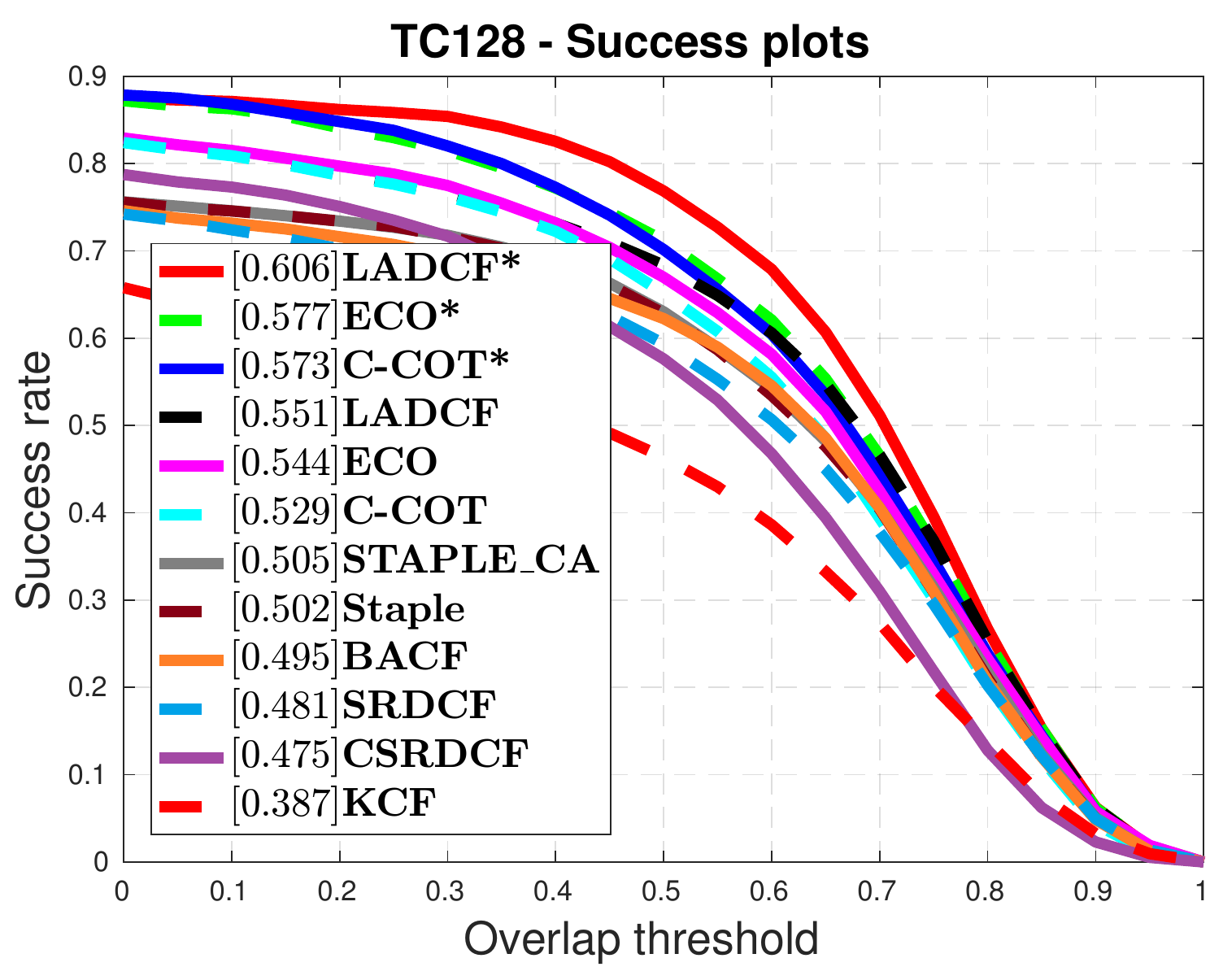}
   \includegraphics[width=0.245\linewidth]{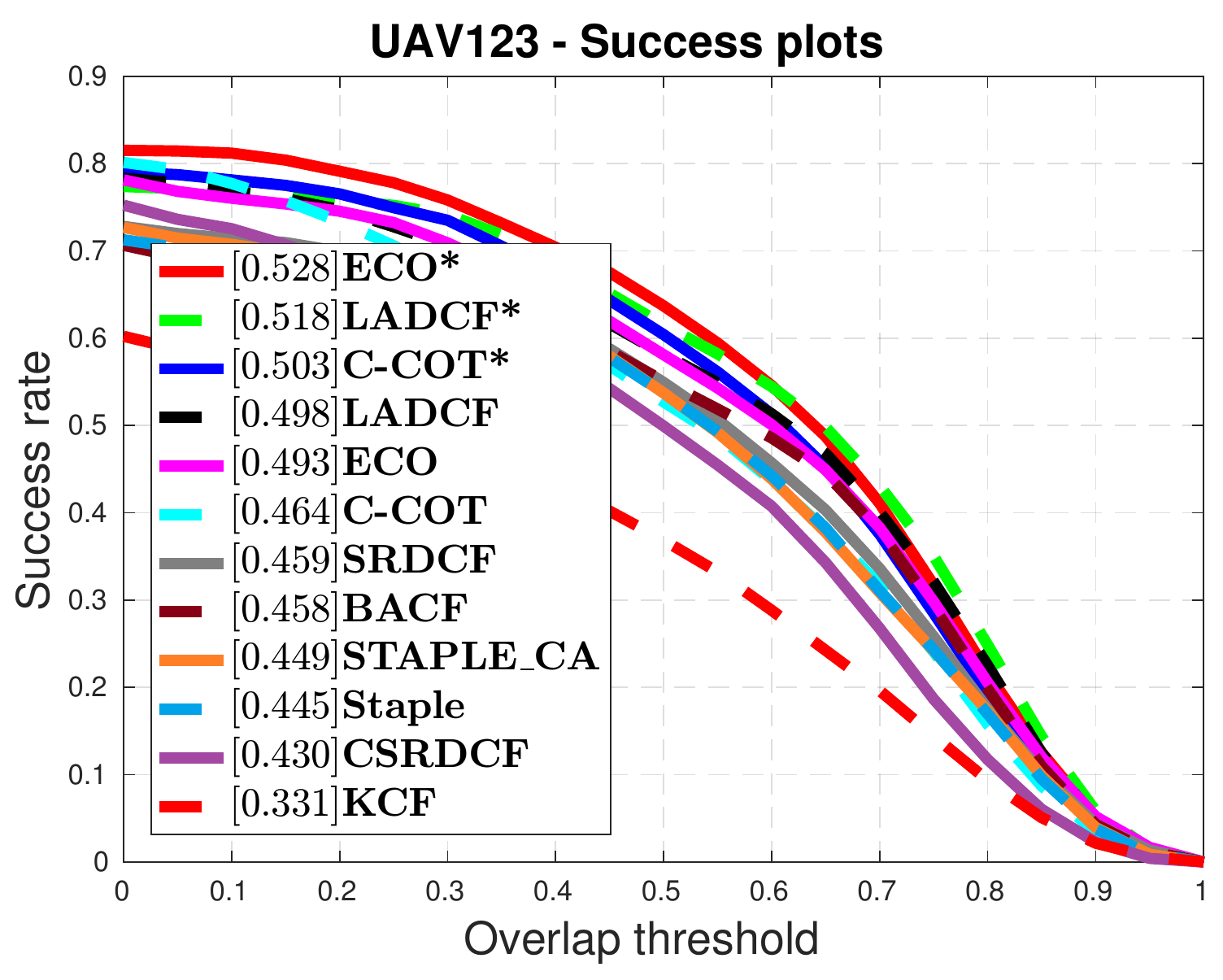}
\end{center}
   \caption{Success plots of tracking performance on OTB2013, OTB50, Temple-Colour and UAV123, with AUC score reported in the figure legend.}
\label{successplot4sets}
\end{figure*}

\begin{figure*}[!t]
   \includegraphics[width=0.245\linewidth]{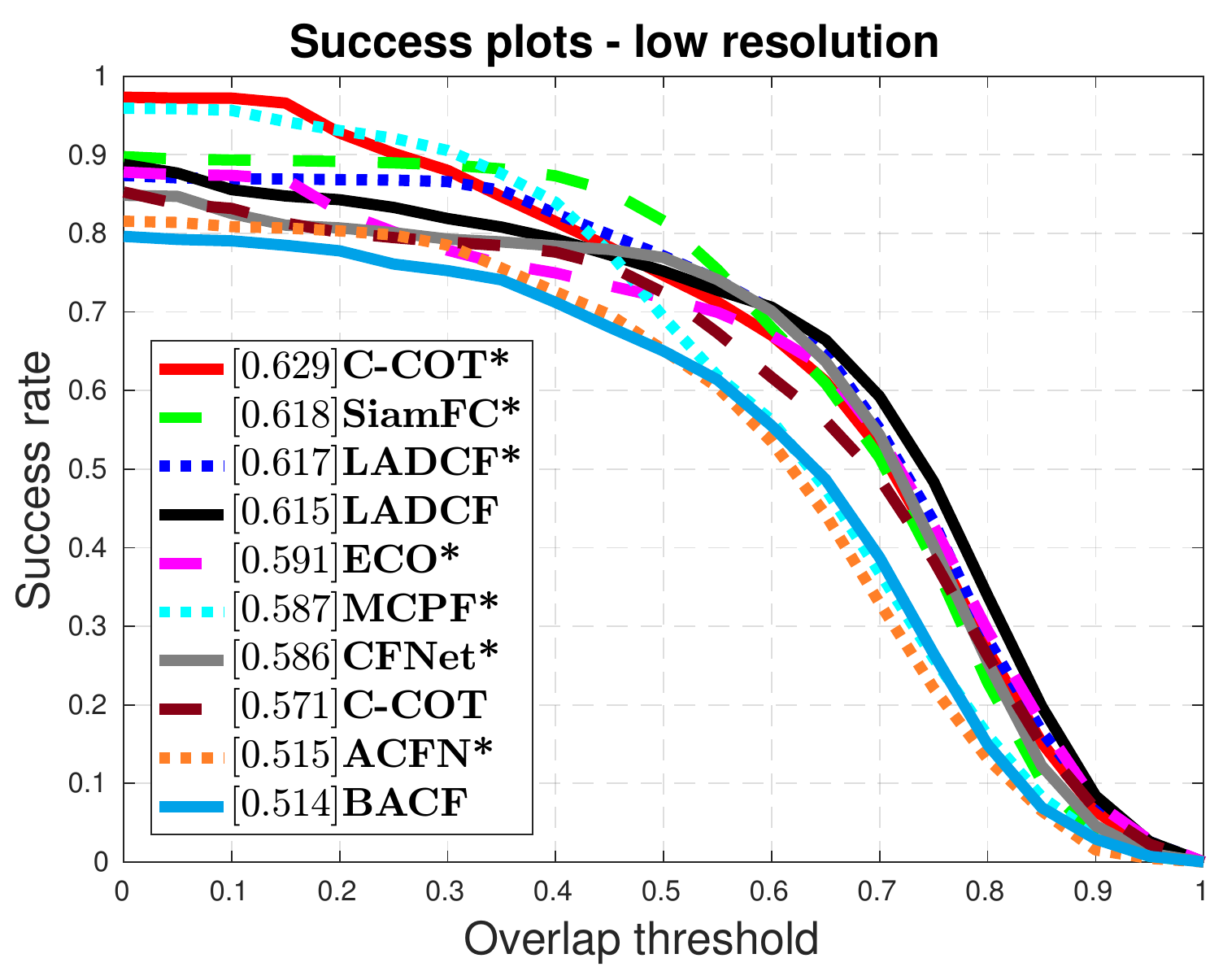}
   \includegraphics[width=0.245\linewidth]{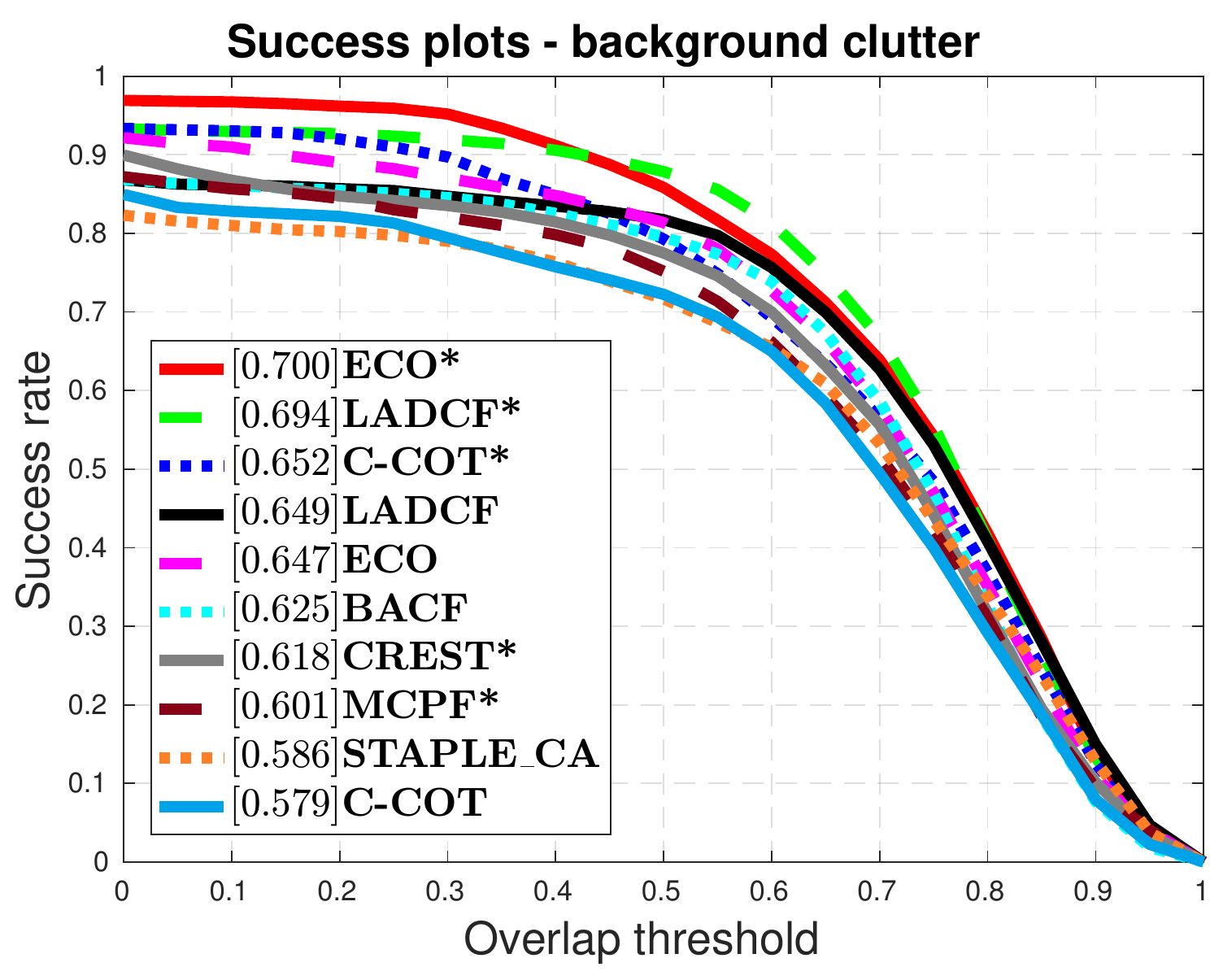}
   \includegraphics[width=0.245\linewidth]{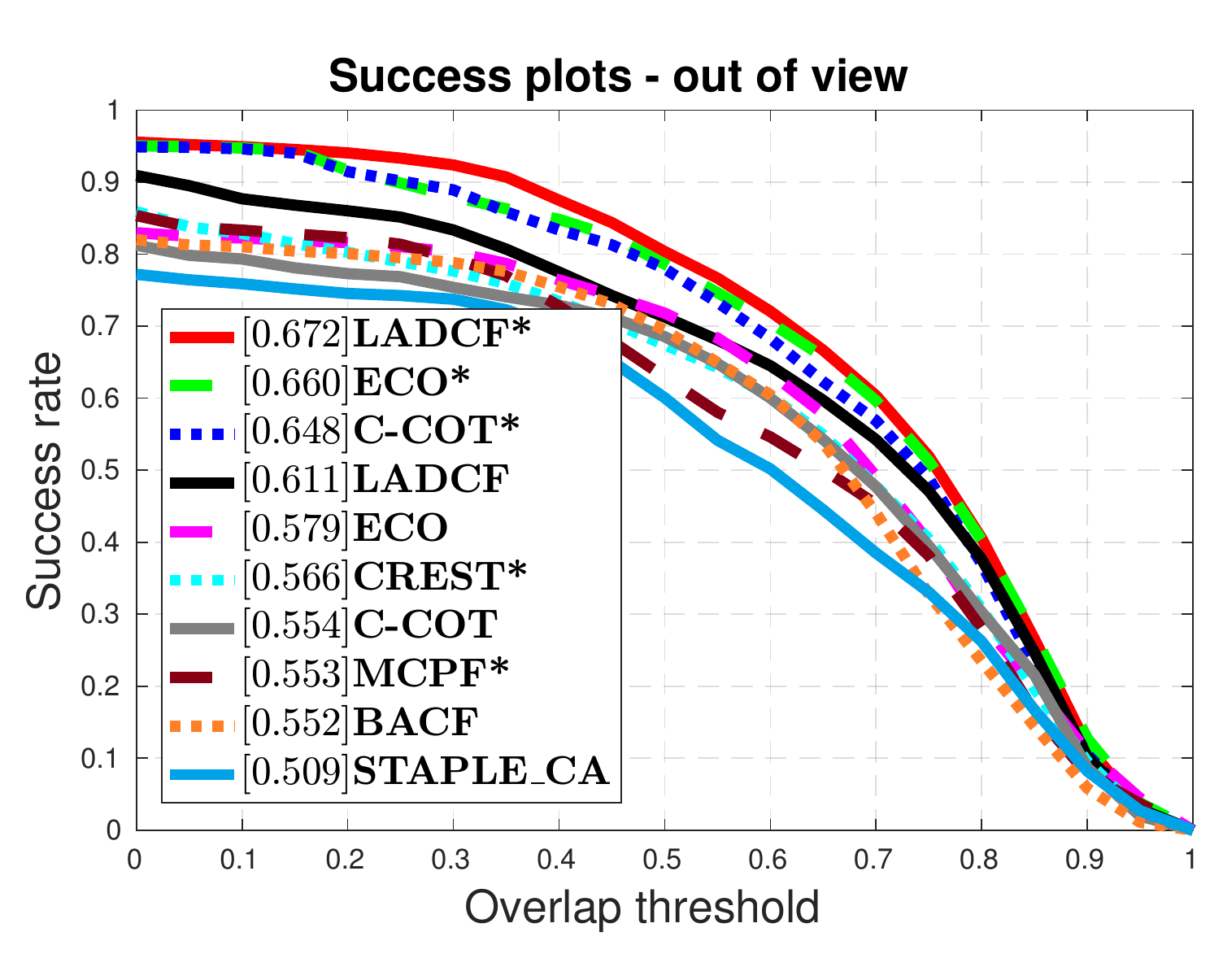}
   \includegraphics[width=0.245\linewidth]{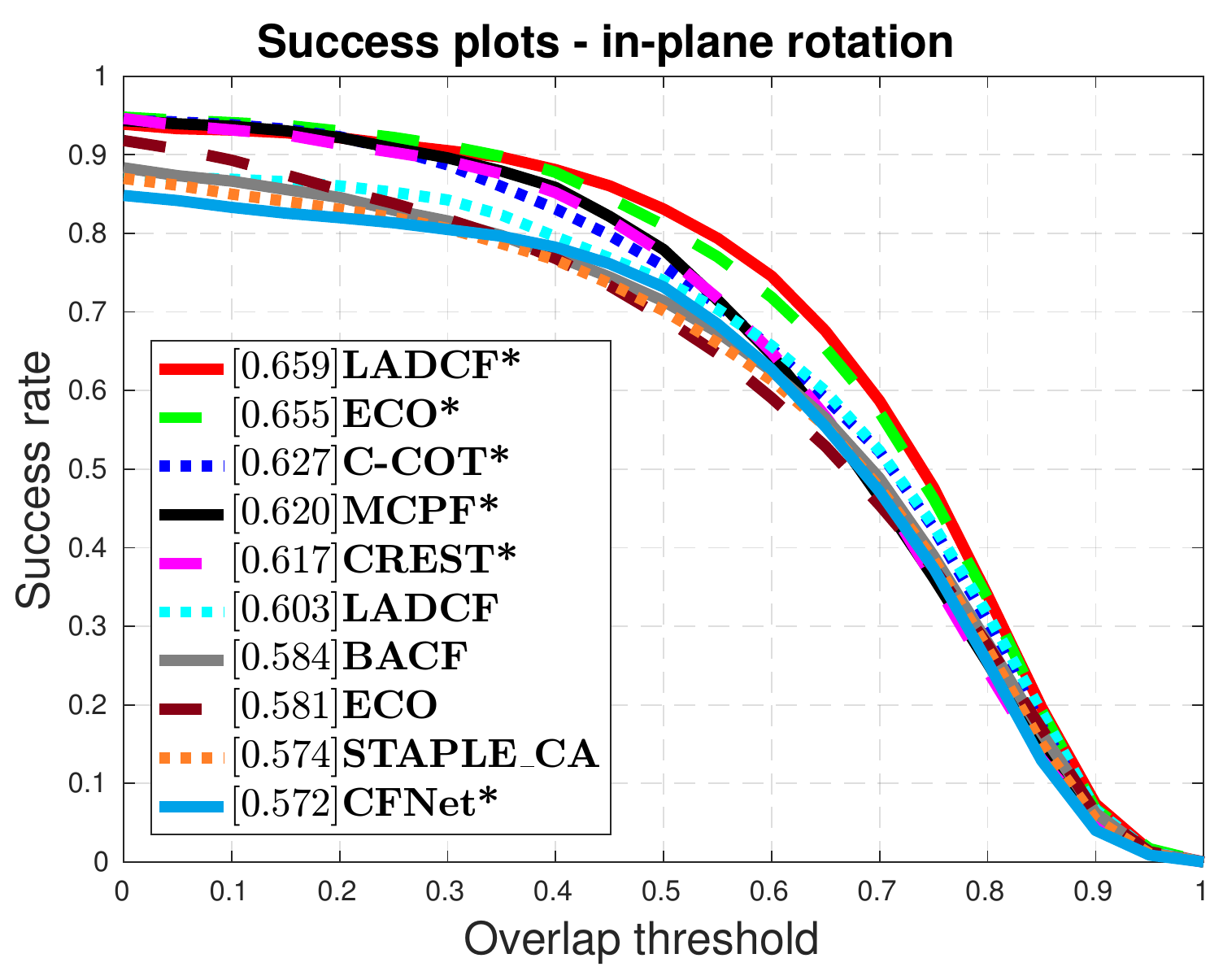}\\
   \includegraphics[width=0.245\linewidth]{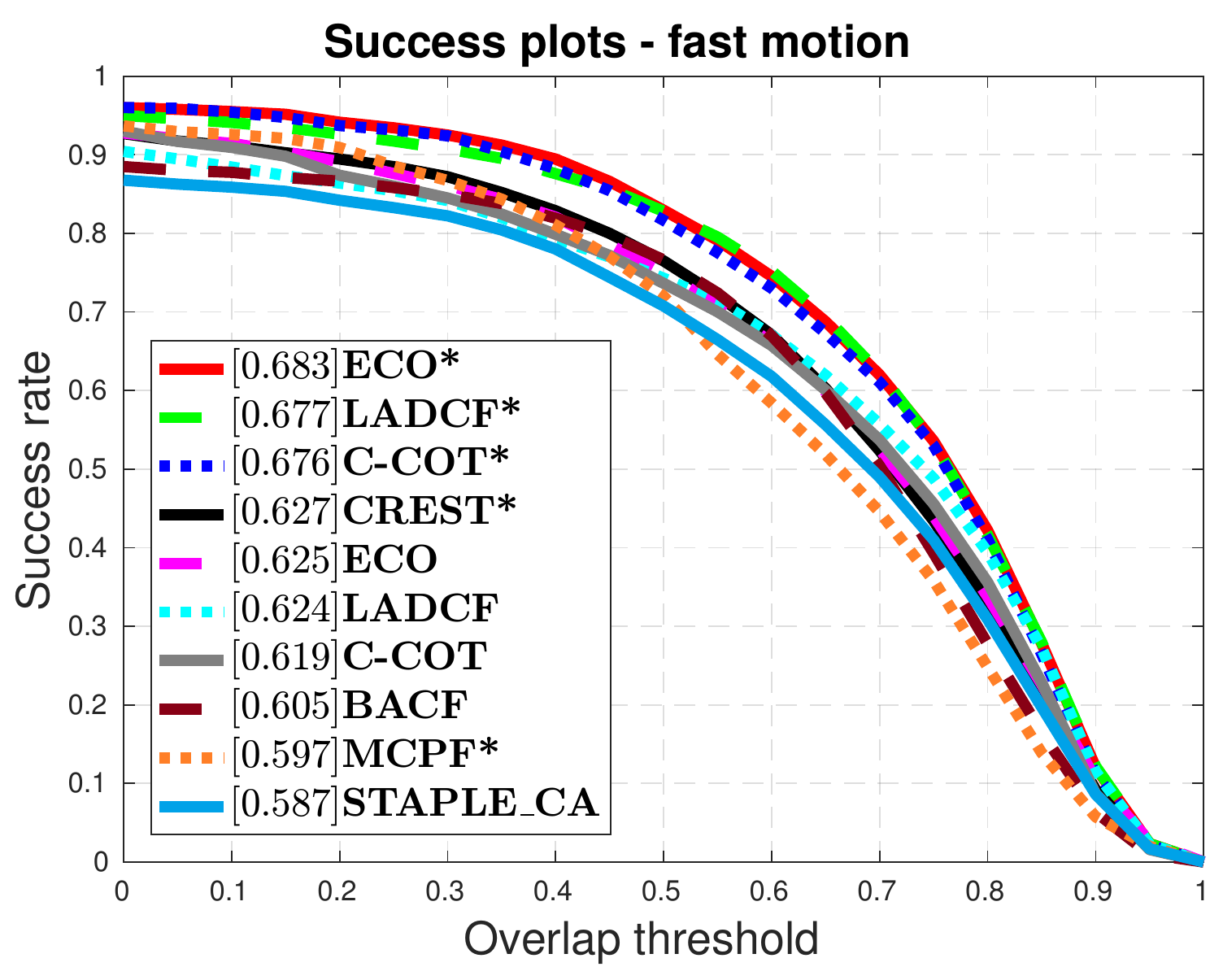}
   \includegraphics[width=0.245\linewidth]{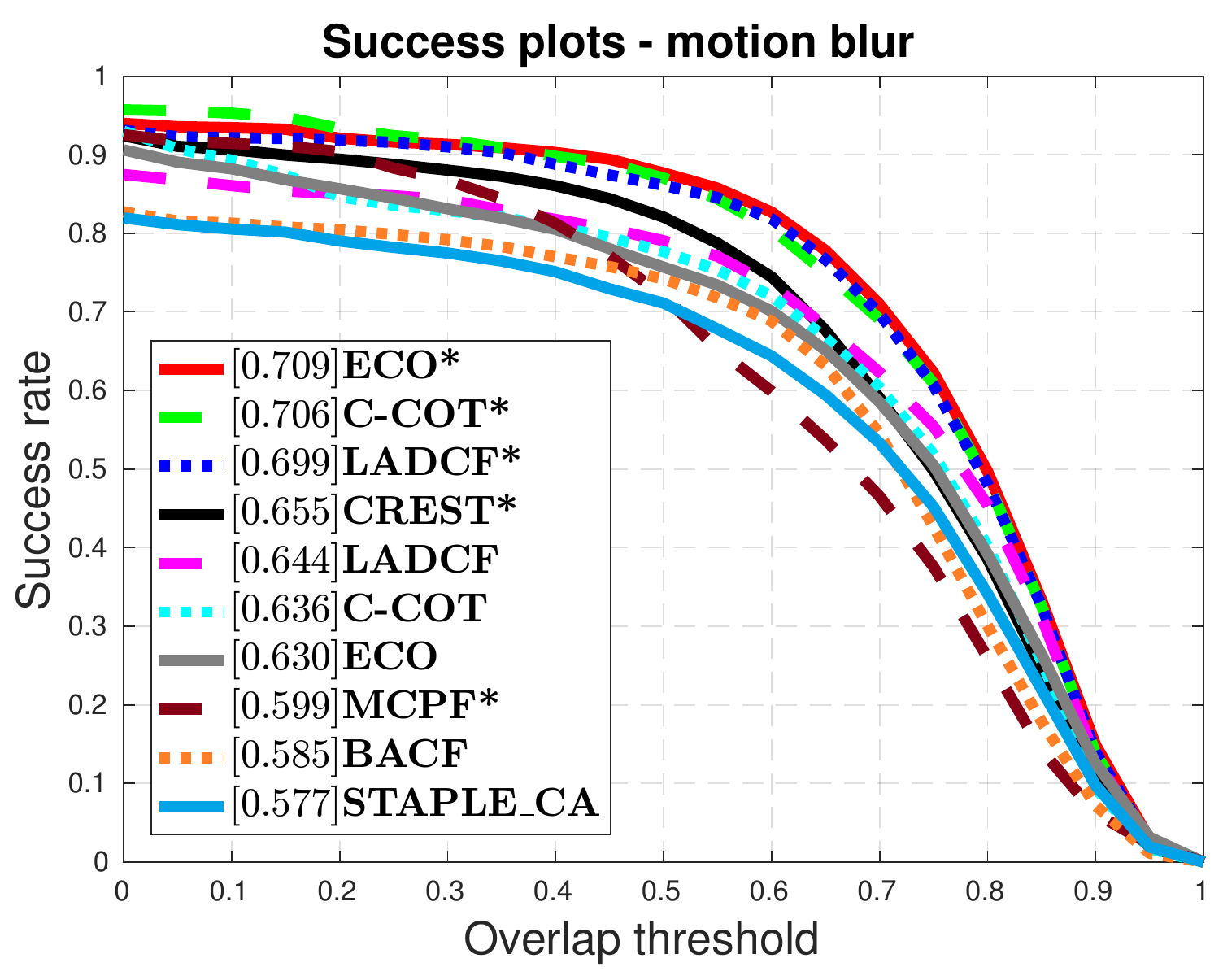}
   \includegraphics[width=0.245\linewidth]{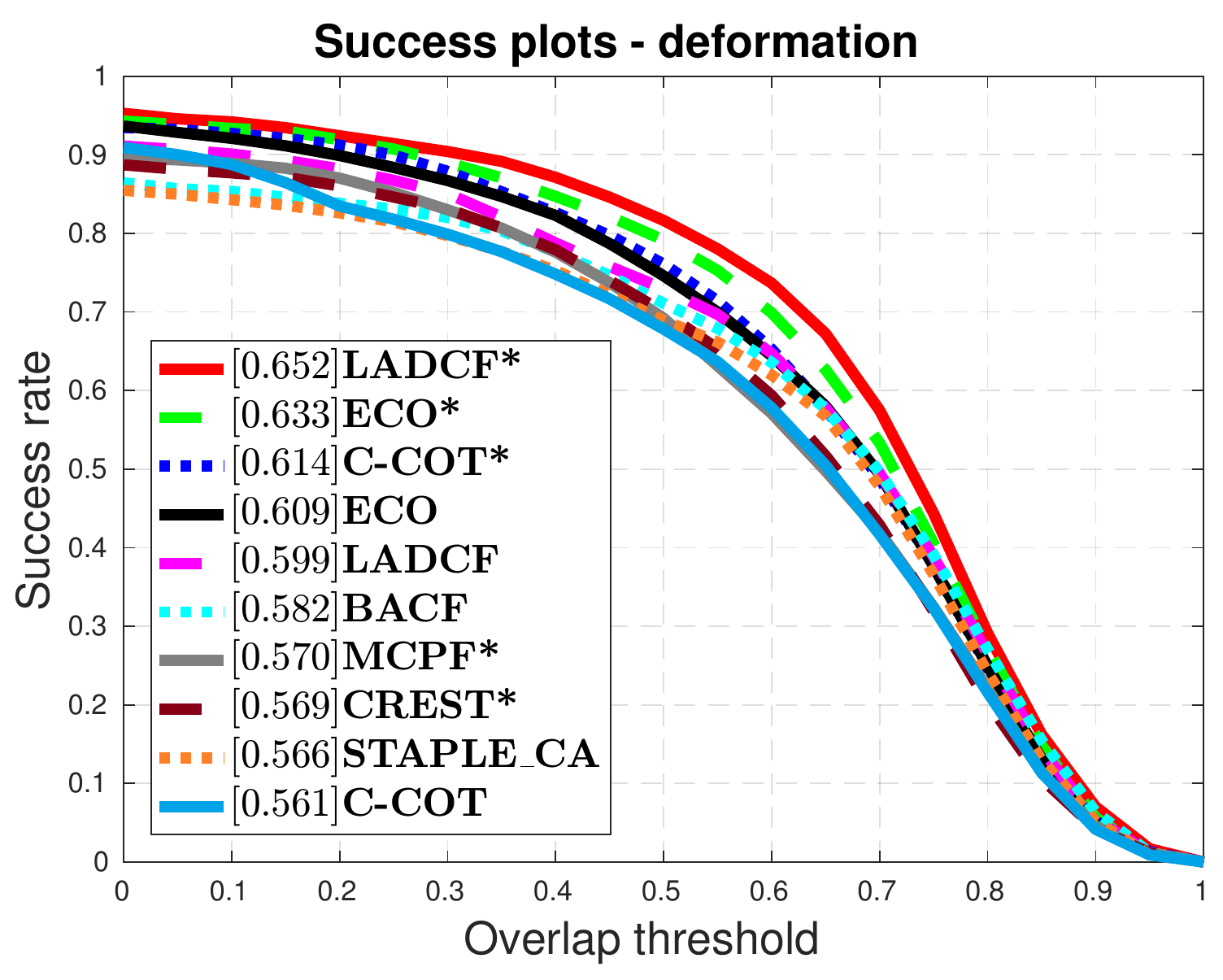}
   \includegraphics[width=0.245\linewidth]{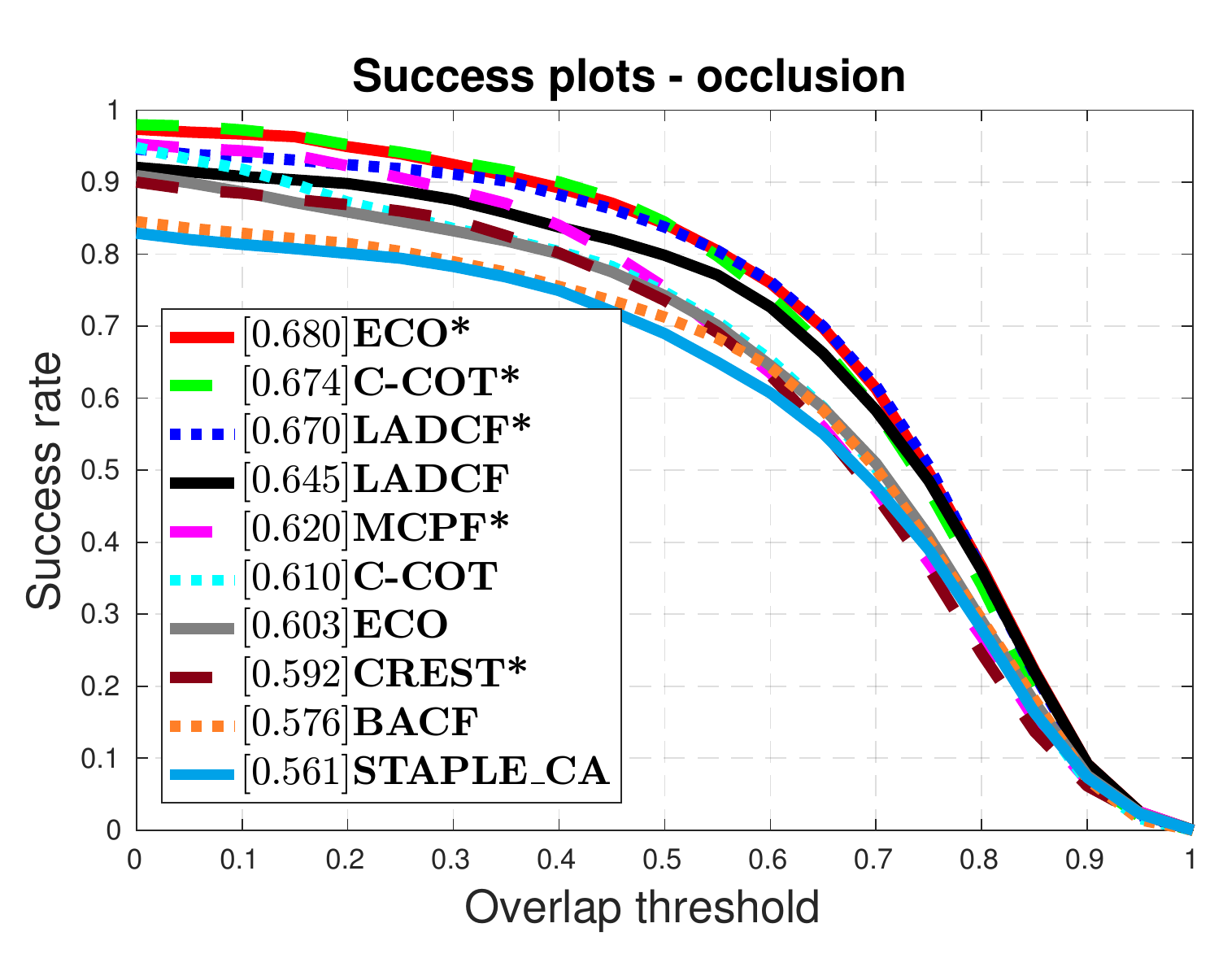}\\
    \includegraphics[width=0.245\linewidth]{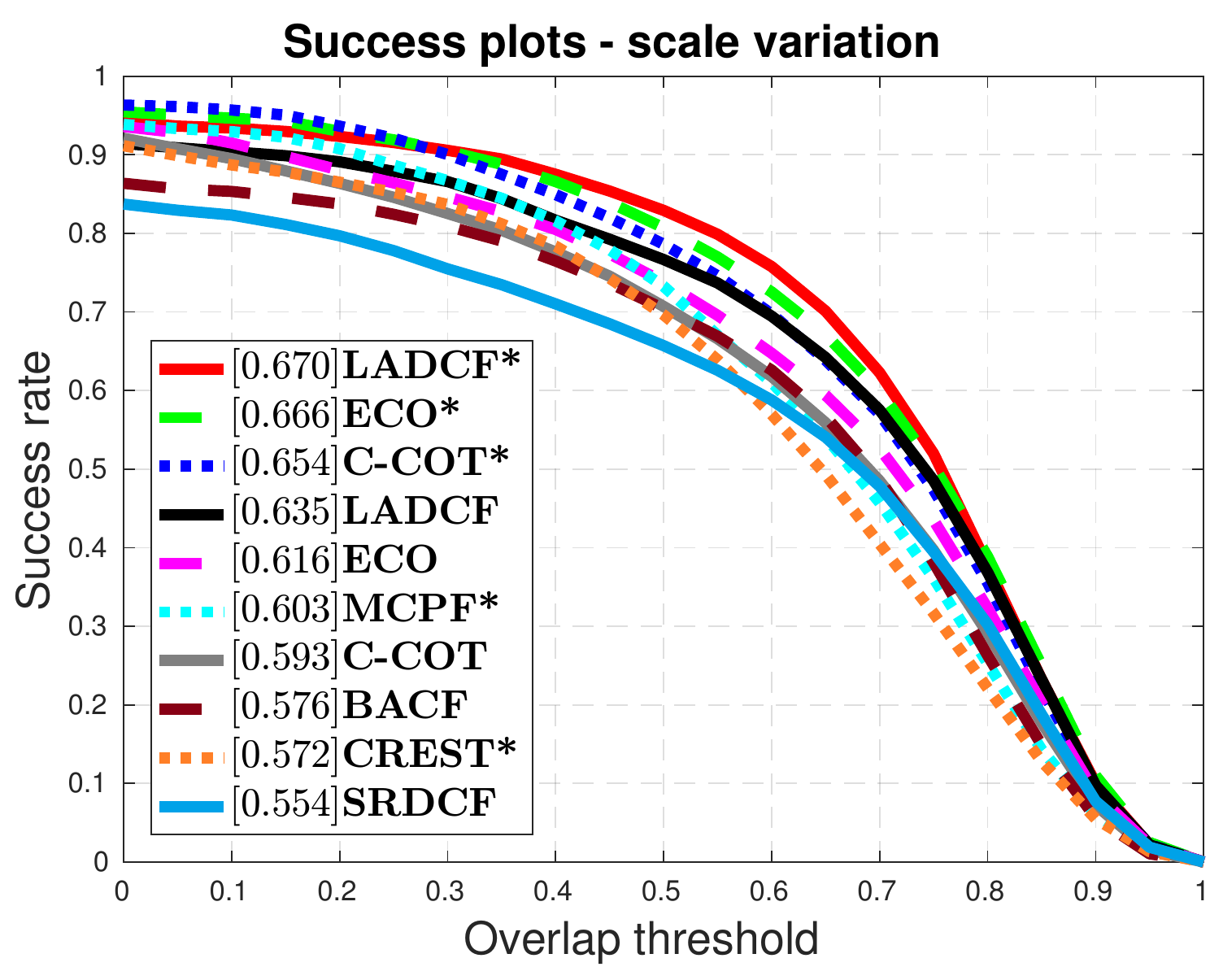}
   \includegraphics[width=0.245\linewidth]{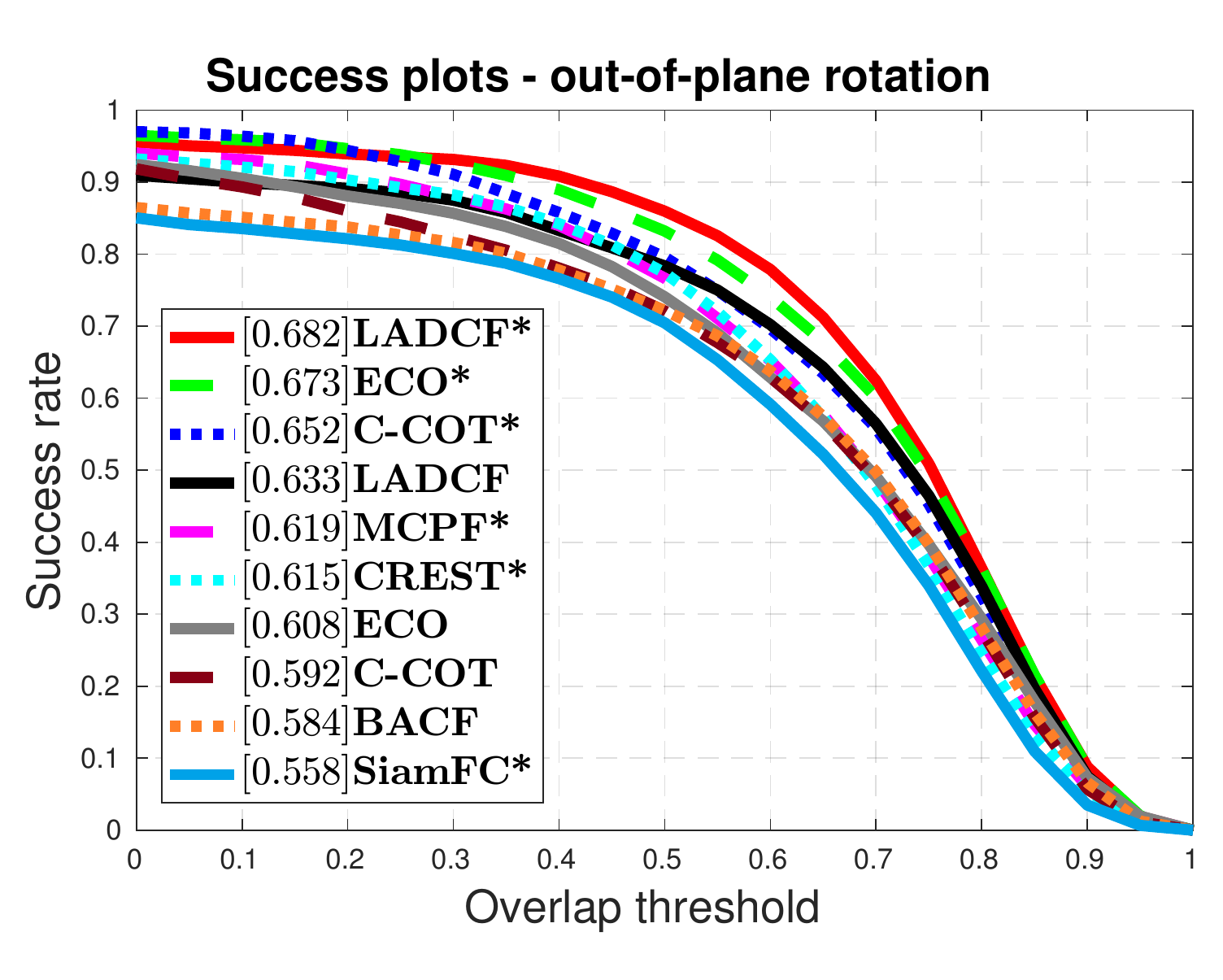}
   \includegraphics[width=0.245\linewidth]{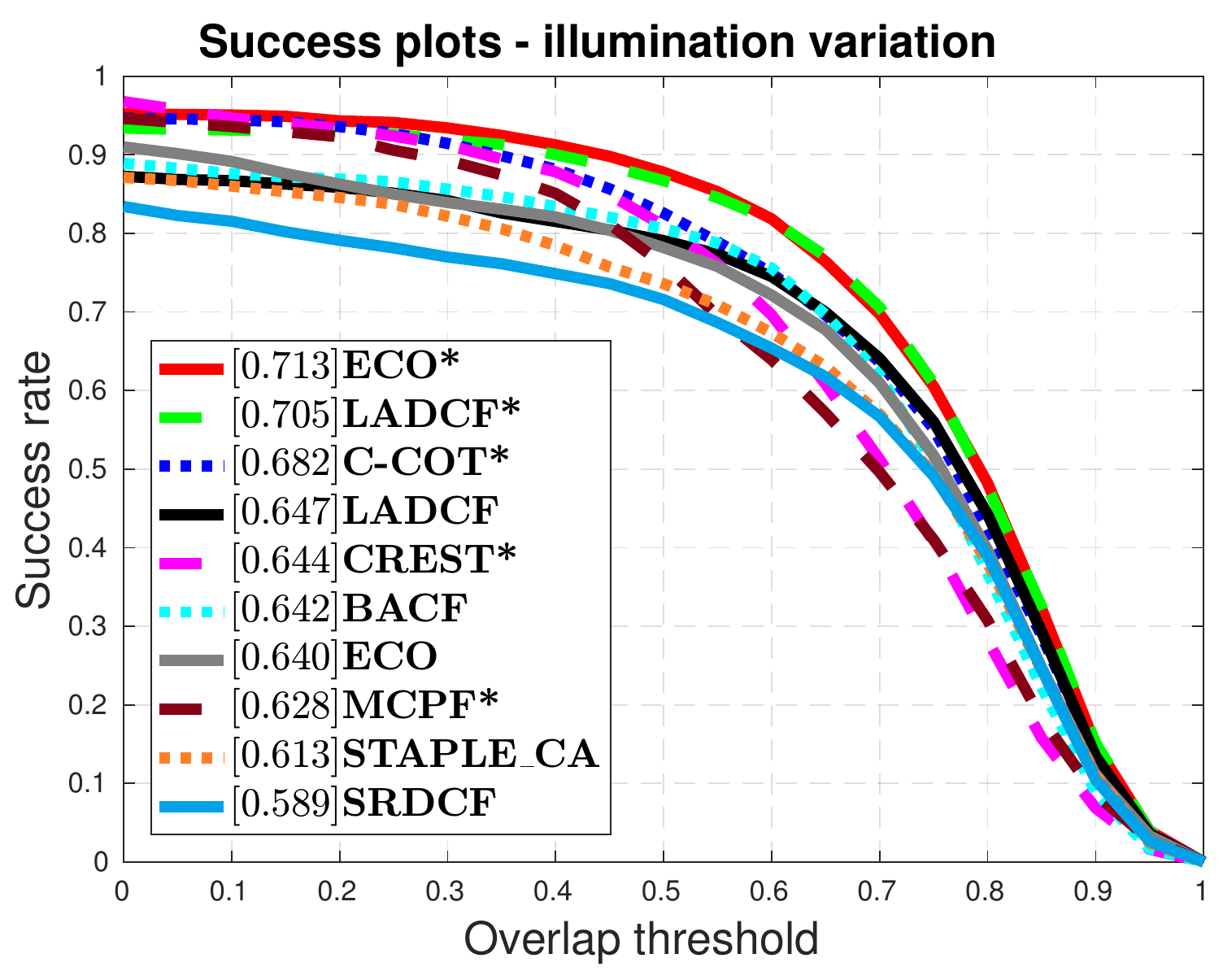}
   \caption{Success plots based on tracking results on OTB100 in 11 sequence attributes, \ie low resolution, background clutter, out of view, in-plane rotation, fast motion, motion blur, deformation, occlusion, scale variation, out-of-plane rotation and illumination variation. For clarity, only the top 10 trackers are presented for each attributes.}\label{attributes}
\end{figure*}

\noindent \textbf{Parameters setup:} We set the regularisation parameters $\lambda_1$ and $\lambda_2$ in Eqn.~(\ref{obj4}) to $1$ and $15$, respectively. The initial penalty parameter $\mu=1$ and maximum penalty $\mu_{\max}=20$ reached with the amplification  rate $\rho=5$. We set the maximum number of iterations as $K=2$, the padding parameter as $\varrho=4$, the scale factor as $a=1.01$ and the number of scales as $S=5$. 
The number of selected spatial features is set as $M = D^2\times r$, where $D^2$ is the number of spatial features, determined by the target size and feature cell size in each sequence, and $r$ is the selection ratio. 
{\color{black}{In practice, to perform feature selection, we use the measures in Eqn.~(\ref{sub2s}). Specifically, we first calculate the group attributes in channel domain ($\theta^j$) and then eliminate the features across all the spatial locations corresponding to a predefined proportion with the lowest grouping attributes. 
This selection strategy has been commonly used in many previous studies~\cite{Peng2015Robust,xu2011two}.}}
For the LADCF using only hand-crafted features, we set $r = 5\%$. For deep feature based LADCF$^\ast$, we set $r = 20\%$. The learning rate $\alpha$ in Eqn.~(\ref{updatetheta}) is set to $0.95$ and $0.13$ for LADCF and LADCF$^{\ast}$, respectively. 

\noindent \textbf{Experimental platform:} We implement our LADCF and LADCF$^{\ast}$ in MATLAB 2016a on an Intel i5 2.50 GHz CPU and NVIDIA GeForce GTX 960M GPU. The source code is available at \url{https://github.com/XU-TIANYANG/LADCF.git}.

\subsection{Experimental Setup}
\noindent \textbf{Datasets:} We evaluate the performance of our tracking method on five well-known benchmarks: OTB2013~\cite{Wu2013Online}, OTB50~\cite{Wu2015Object}, OTB100~\cite{Wu2015Object}, Temple-Colour~\cite{Liang2015Encoding},  UAV123~\cite{mueller2016benchmark} and VOT2018~\cite{kristan2018sixth}. 
OTB2013, OTB50 and OTB100 are widely used datasets that respectively contain 51, 50 and 100 annotated video sequences with 11 sequence attributes. 
Temple-Colour is composed of 128 colour video sequences and UAV123 consists of 123 challenging sequences.
The VOT2018 benchmark has 60 short-term video sequences.

%\begin{figure}[!t]
%\begin{center}
%   \includegraphics[width=0.49\linewidth]{./img/DEEP_ERROR_OTB100.pdf}
%   \includegraphics[width=0.49\linewidth]{./img/DEEP_AUC_OTB100.pdf}
%\end{center}
%   \caption{The experimental results of trackers using deep features/structures on OTB100. The precision plots (left) and the success plots (right) are presented. }
%   \label{otb100deep}
%\end{figure}

\begin{table*}[!t]
\footnotesize
\renewcommand{\arraystretch}{1.1}
%\setcaptionwidth{0.86\linewidth}
\caption{Tracking results on VOT2018. (The best three results are highlighted by {\color{red}{red}}, {\color{blue}{blue}} and {\color{brown}{brown}}.)}
\label{vot18}
\centering
\begin{tabular}{l|ccccccc|cc}
\hline
& ECO$^\ast$ & CFCF$^\ast$~\cite{gundogdu2018good} & CFWCR$^\ast$~\cite{he2017correlation} & LSART$^\ast$~\cite{sun2018learning} & UPDT$^\ast$~\cite{bhat2018unveiling} & SiamRPN$^\ast$~\cite{zhu2018distractor} & MFT$^\ast$~\cite{kristan2018sixth} & \makecell{LADCF$^\ast$\\VGG}& \makecell{LADCF$^\ast$\\ResNet50}\\
\hline
\textbf{EAO} & 0.280 & 0.286 & 0.303 & 0.323  & 0.378 & {\color{brown}{\textbf{0.383}}} & {\color{blue}{\textbf{0.385}}} & 0.338 & {\color{red}{\textbf{0.389}}} \\
\textbf{Accuracy} & 0.483 & 0.509 & 0.484 & 0.493 & {\color{blue}{\textbf{0.536}}} & {\color{red}{\textbf{0.586}}} & 0.505 & {\color{brown}{\textbf{0.512}}} & 0.503 \\
\textbf{Robustness} & 0.276 & 0.281 & 0.267 & 0.218 & {\color{brown}{\textbf{0.184}}} & 0.276 & {\color{red}{\textbf{0.140}}} & 0.197 & {\color{blue}{\textbf{0.159}}} \\
\hline 
\end{tabular}
\end{table*}

\noindent \textbf{Evaluation metrics:} We follow the One Pass Evaluation (OPE) protocol~\cite{Wu2013Online} to evaluate the performance of different trackers. 
The precision and success plots are reported based on centre location error and bounding box overlap. The Area Under Curve (AUC),  Overlap Precision (OP, percentage of overlap ratios exceeding 0.5) and Distance Precision (DP, percentage of location errors within 20 pixels) are the criteria used in the evaluation.
The speed of a tracking algorithm is measured in Frames Per Second (FPS).
For VOT2018, we employ the expected average overlap (EAO), accuracy value (A) and robustness value (R) to evaluate the performance~\cite{Kristan2016The}.
\noindent \textbf{State-of-the-art competitors:} 
We compare our proposed tracking method with 13 state-of-the-art trackers, including  Staple~\cite{Bertinetto2016Staple}, SRDCF~\cite{Danelljan2015Learning}, KCF~\cite{Henriques2015High}, CSRDCF~\cite{Lukezic2017Discriminative}, STAPLE\_CA~\cite{mueller2017context}, BACF~\cite{Galoogahi2017Learning}, C-COT~\cite{Danelljan2016Beyond}, ECO~\cite{Danelljan2016ECO}, ACFN$^\ast$~\cite{choi2017attentional}, CREST$^\ast$~\cite{song-iccv17-CREST}, SiamFC$^\ast$~\cite{bertinetto2016fully}, CFNet$^\ast$~\cite{valmadre2017end},  MCPF$^\ast$~\cite{zhang2017multi}, C-COT$^\ast$ and ECO$^\ast$. 
Specifically, trackers followed by $^\ast$ are equipped with deep features/structures. 
For a fair comparison, we use the original publicly available codes from the authors for the evaluation.

\subsection{Results and Analysis}
\noindent \textbf{Quantitative results of the overall tracking performance:}
We report the experimental results using the precision and success plots obtained on OTB100 in Fig.~\ref{otb_100_ladcf}, with DP and AUC scores in the figure legend.
Compared with the other trackers using hand-crafted features in Fig.~\ref{otb100hc}, our LADCF performs best by achieving $86.4\%$ in DP and $66.4\%$ in AUC.
Again, compared with deep features/structures based trackers in Fig.~\ref{otb100deep}, our LADCF$^\ast$ outperforms all the other trackers in terms of AUC with a score of $69.6\%$. Our LADCF$^\ast$ achieves $90.6\%$ in DP, which ranks the second best, with only $0.4\%$ fall behind ECO$^\ast$. The main difference between our LADCF$^\ast$ and ECO$^\ast$ is that ECO$^\ast$ employs continuous convolution operators and uses a mixture of different deep network layers (Conv-1 and Conv-5), which benefit its accuracy. 

\begin{figure*}[htbp]
\begin{center}
%% cancel the following %s after edit
\includegraphics[width=0.16\linewidth]{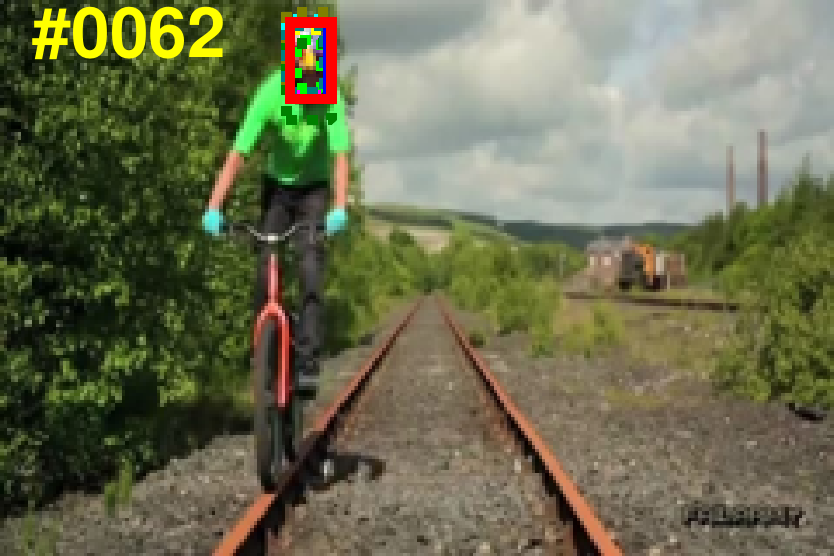}
\includegraphics[width=0.16\linewidth]{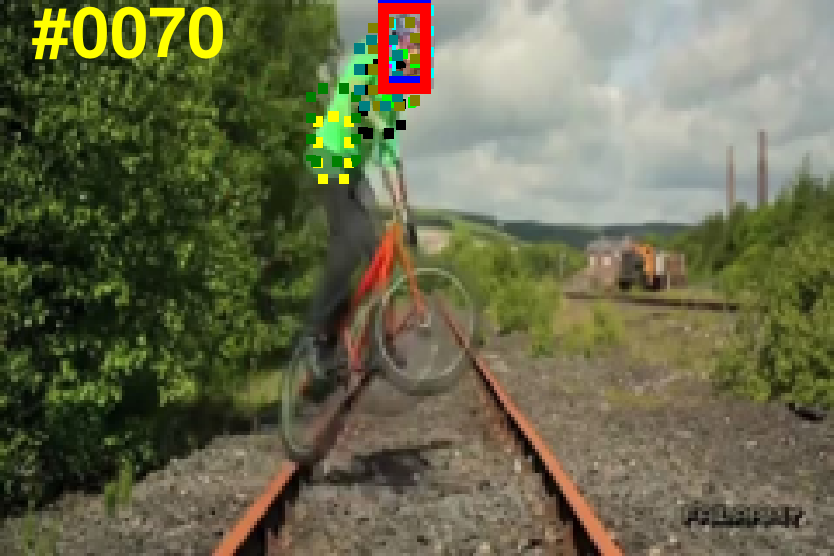}
\includegraphics[width=0.16\linewidth]{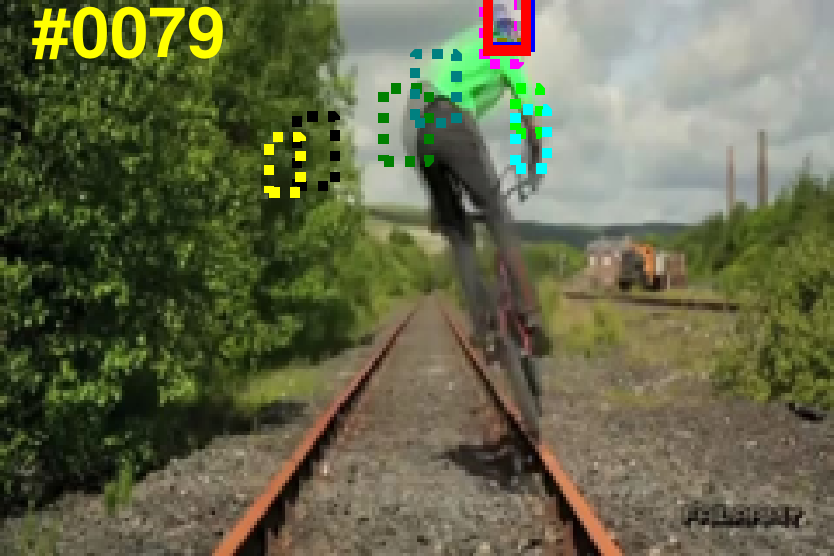}
\includegraphics[width=0.16\linewidth]{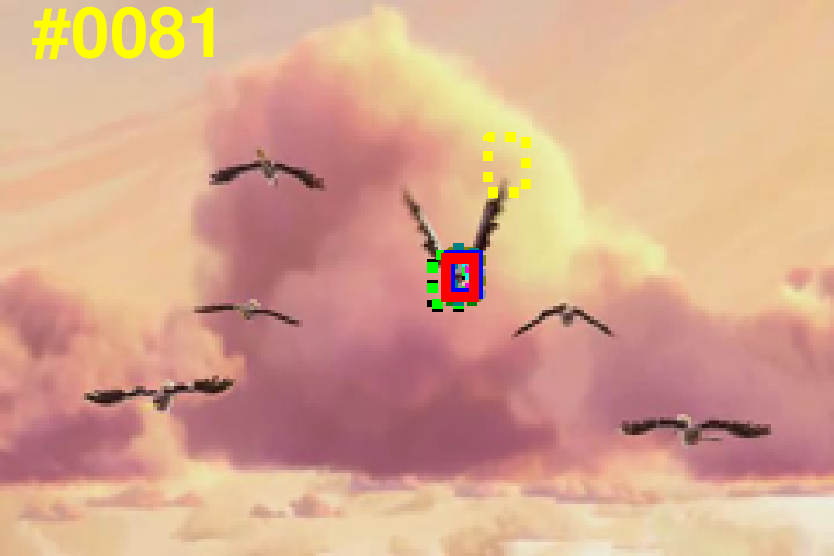}
\includegraphics[width=0.16\linewidth]{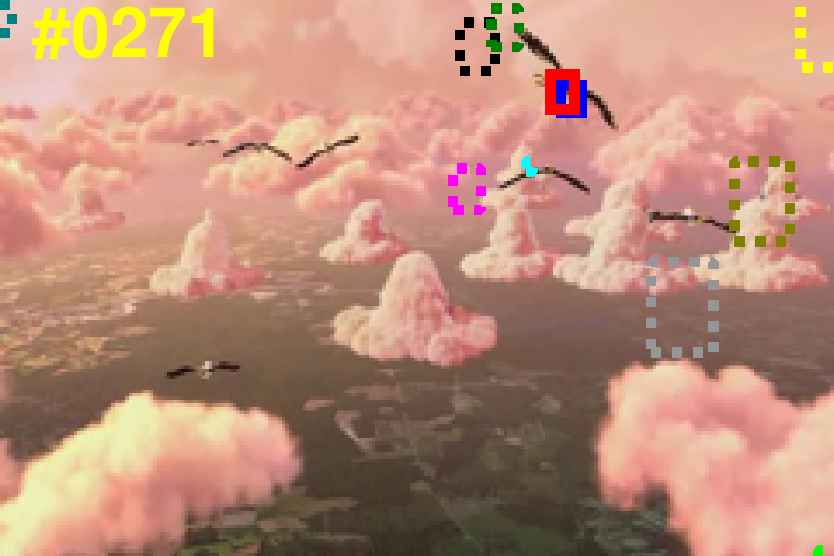}
\includegraphics[width=0.16\linewidth]{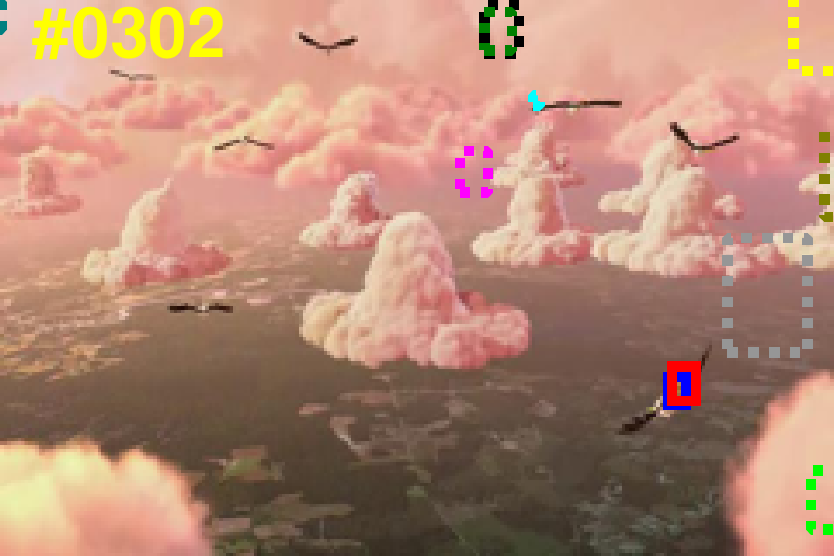}\\
\includegraphics[width=0.16\linewidth]{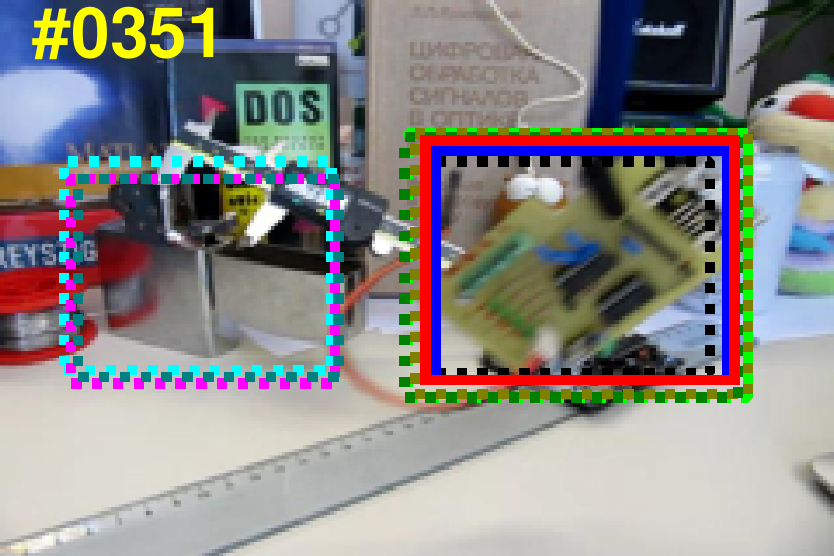}
\includegraphics[width=0.16\linewidth]{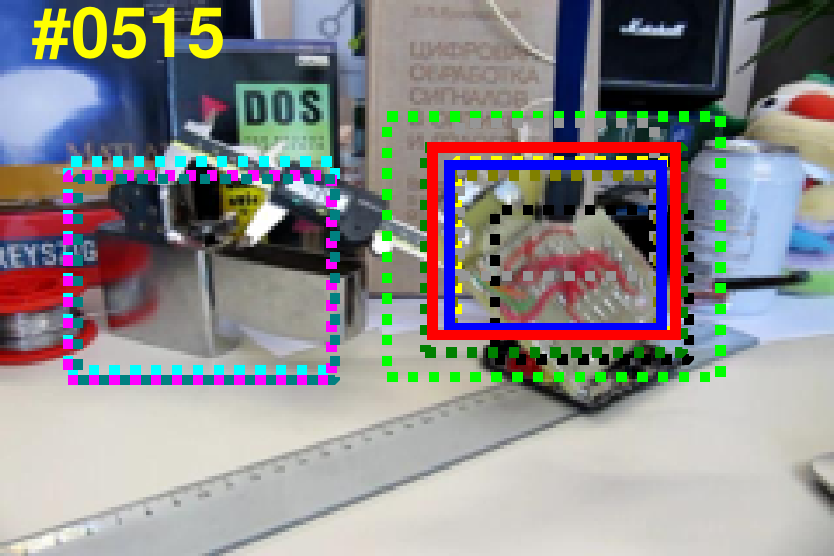}
\includegraphics[width=0.16\linewidth]{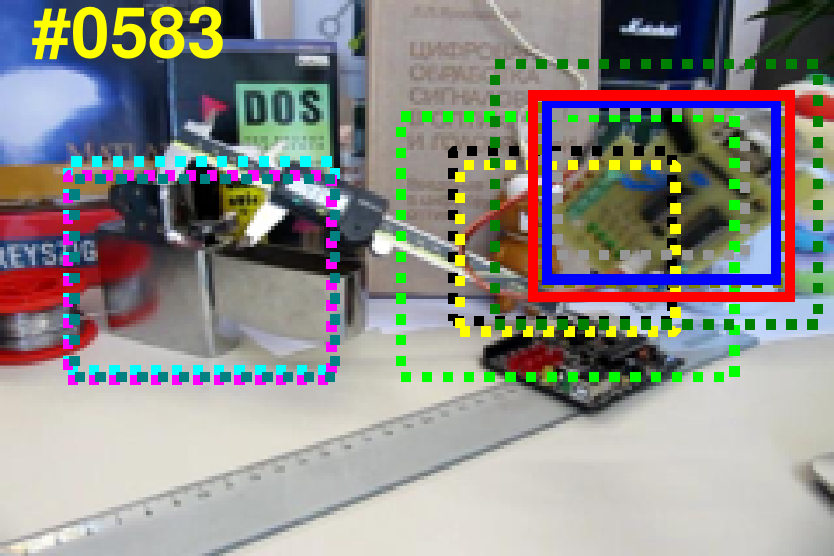}
\includegraphics[width=0.16\linewidth]{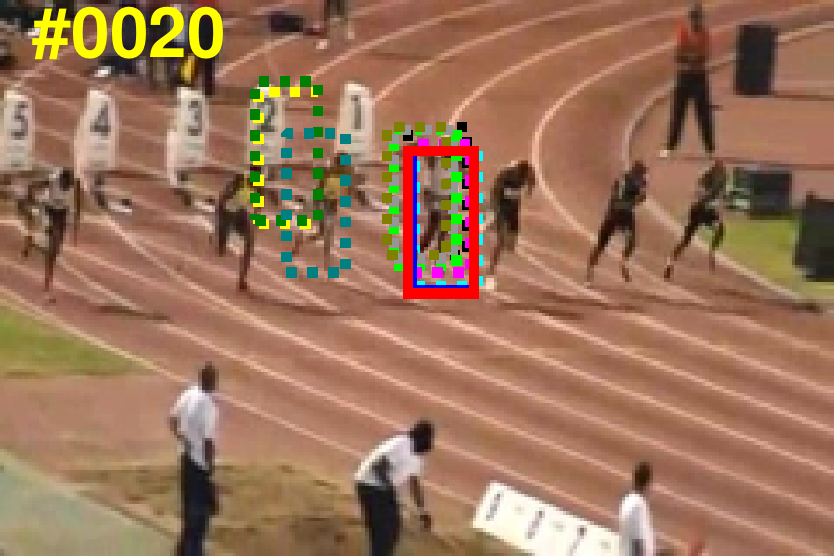}
\includegraphics[width=0.16\linewidth]{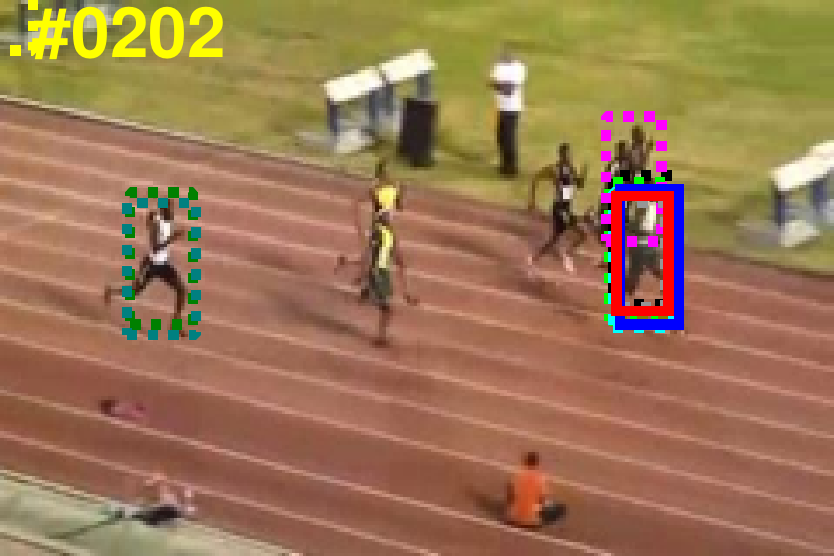}
\includegraphics[width=0.16\linewidth]{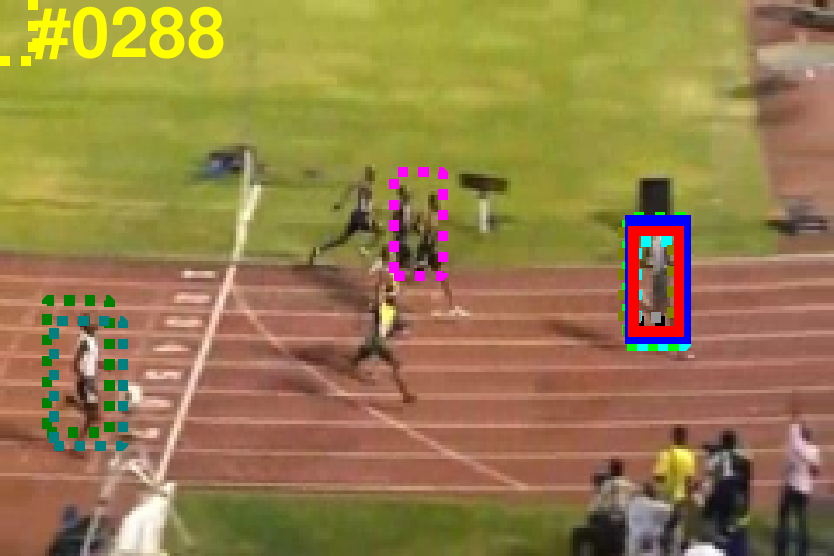}\\
\includegraphics[width=0.16\linewidth]{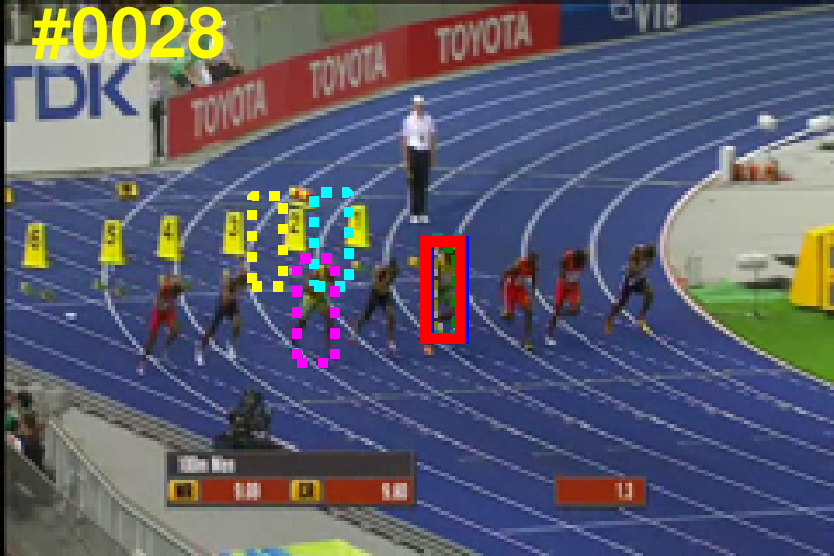}
\includegraphics[width=0.16\linewidth]{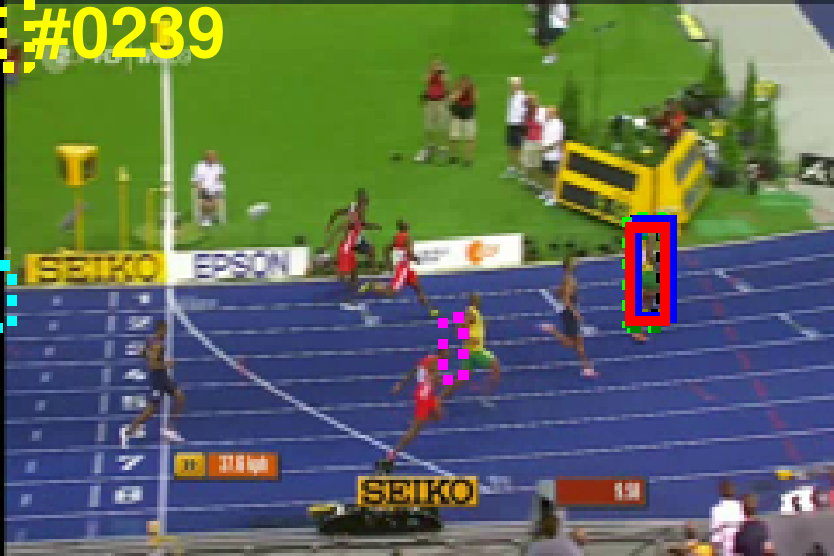}
\includegraphics[width=0.16\linewidth]{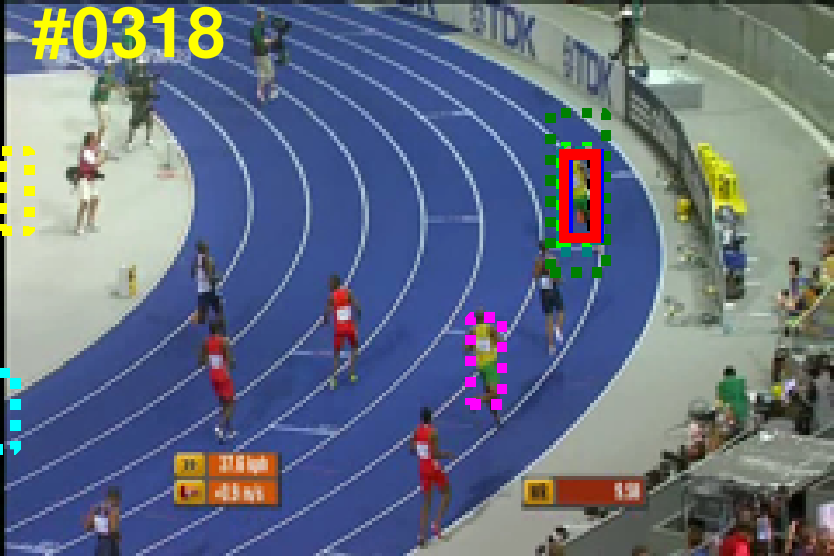}
\includegraphics[width=0.16\linewidth]{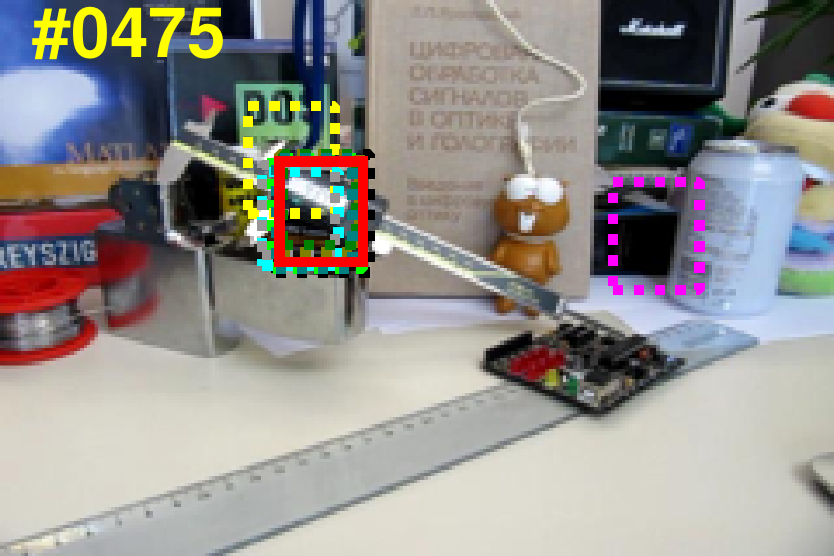}
\includegraphics[width=0.16\linewidth]{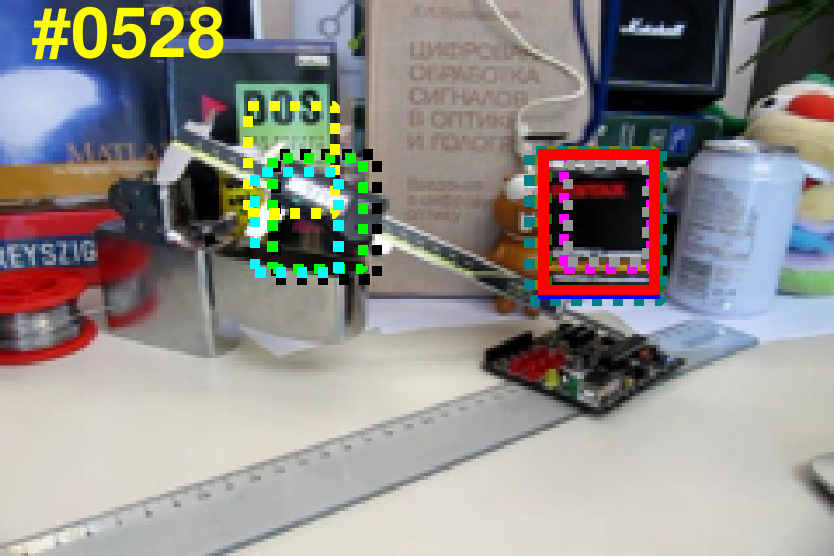}
\includegraphics[width=0.16\linewidth]{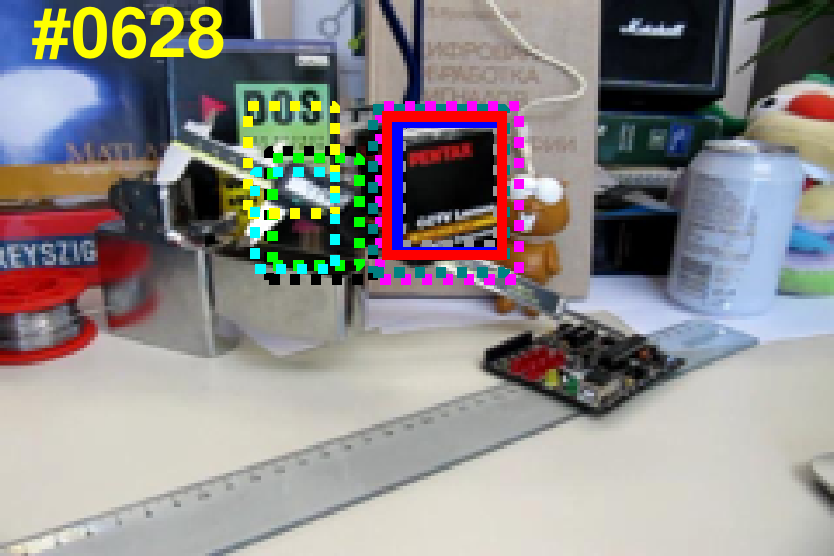}\\
\includegraphics[width=0.16\linewidth]{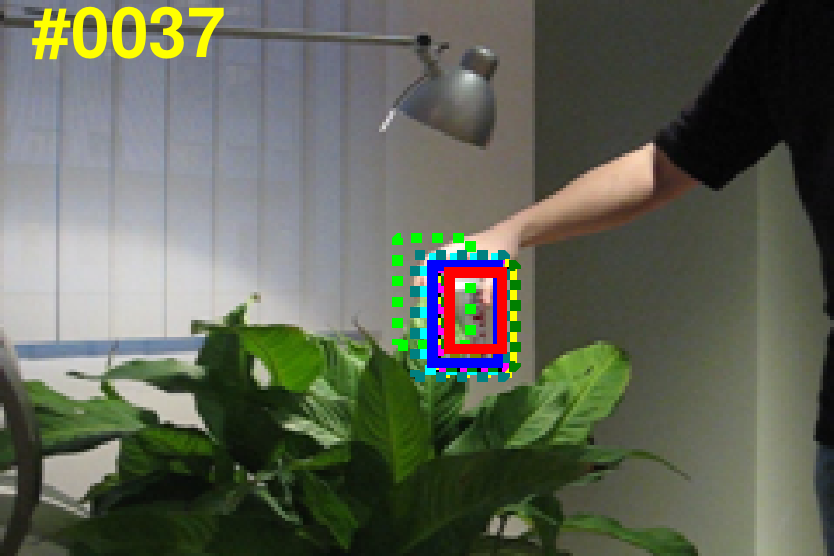}
\includegraphics[width=0.16\linewidth]{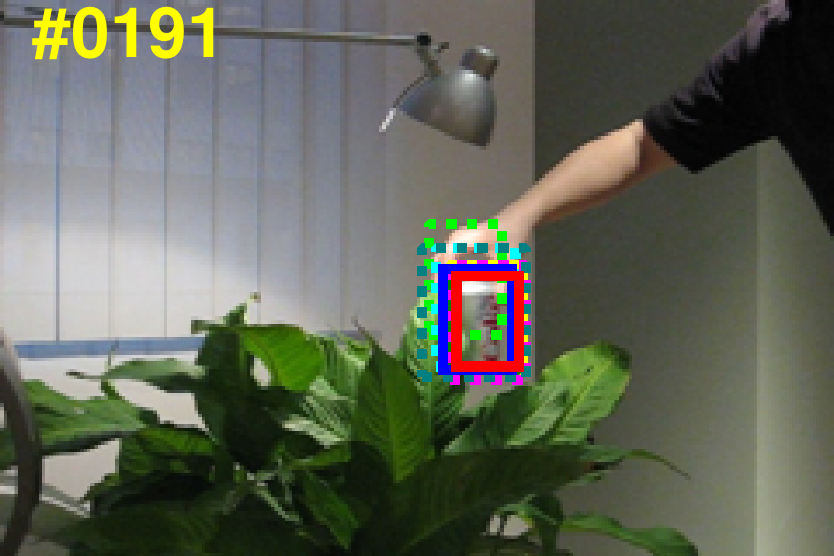}
\includegraphics[width=0.16\linewidth]{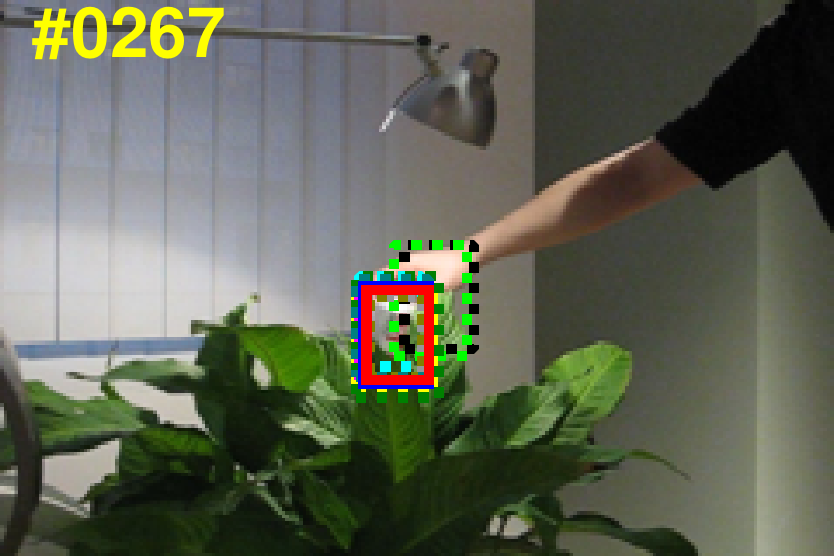}
\includegraphics[width=0.16\linewidth]{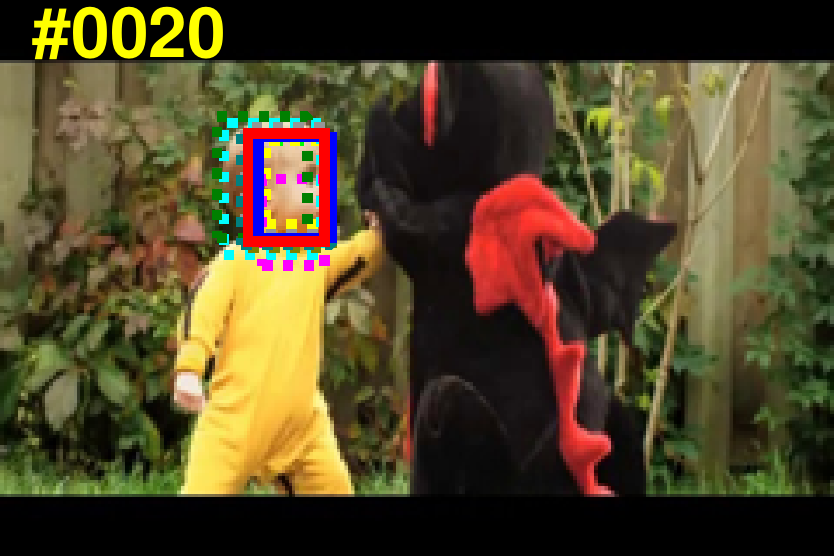}
\includegraphics[width=0.16\linewidth]{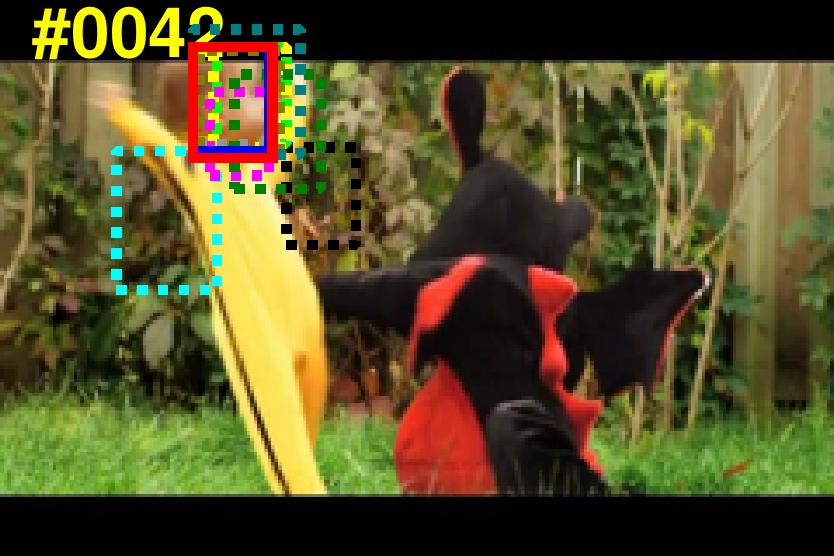}
\includegraphics[width=0.16\linewidth]{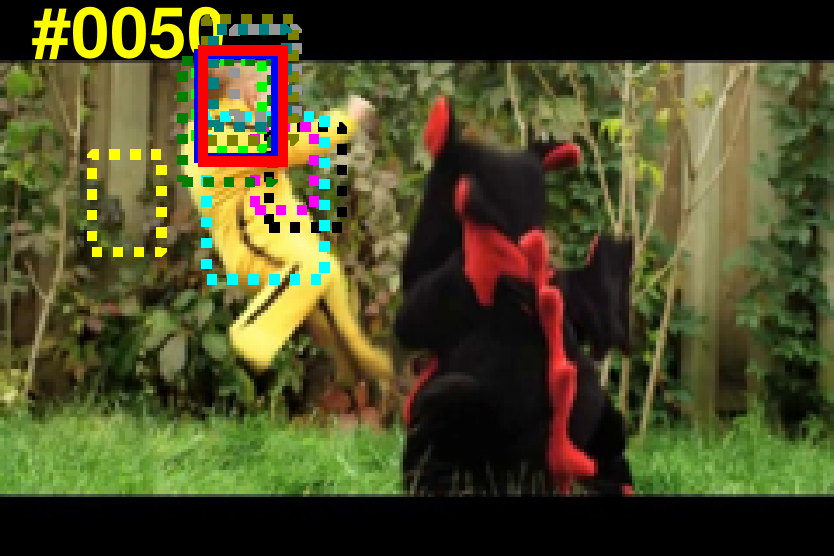}\\
\includegraphics[width=0.16\linewidth]{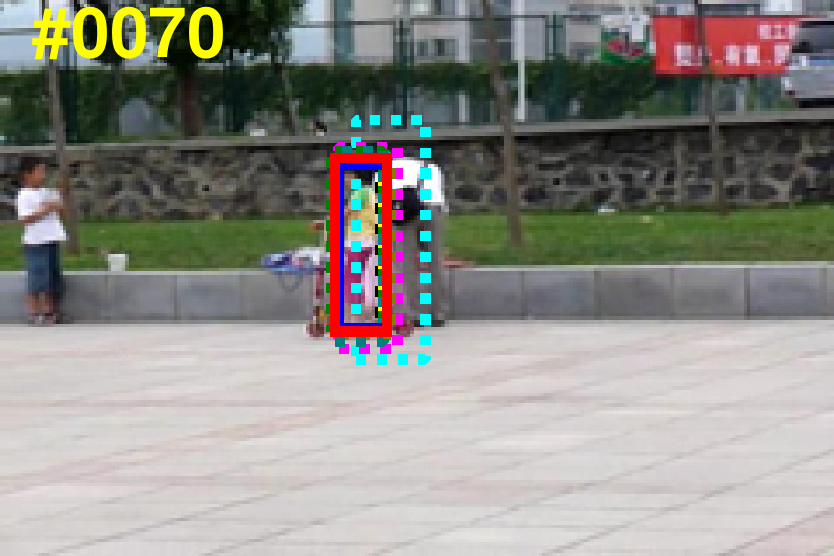}
\includegraphics[width=0.16\linewidth]{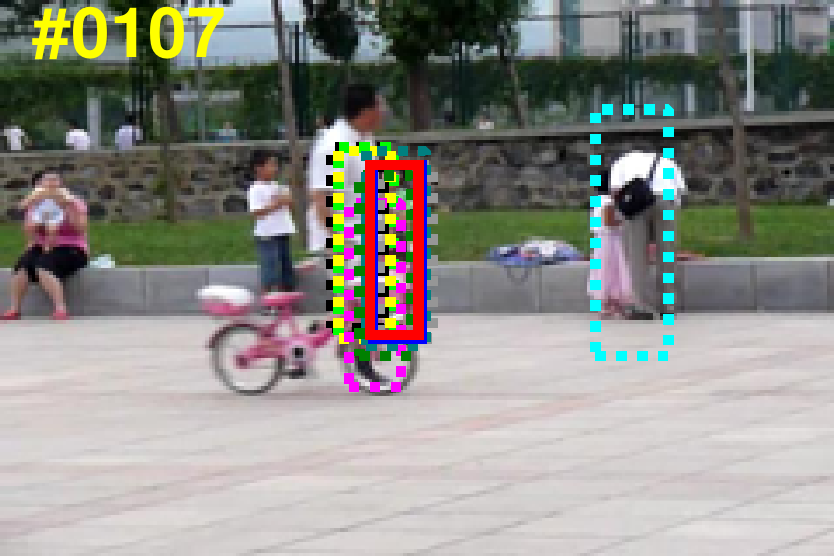}
\includegraphics[width=0.16\linewidth]{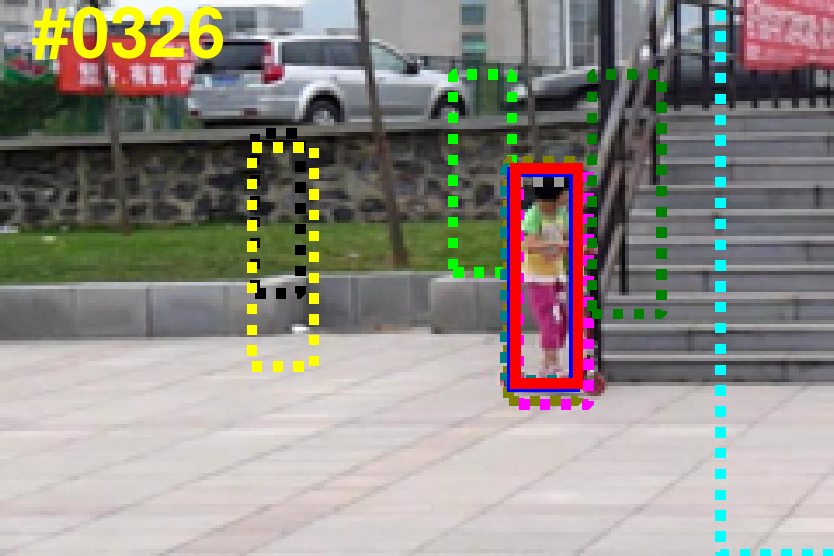}
\includegraphics[width=0.16\linewidth]{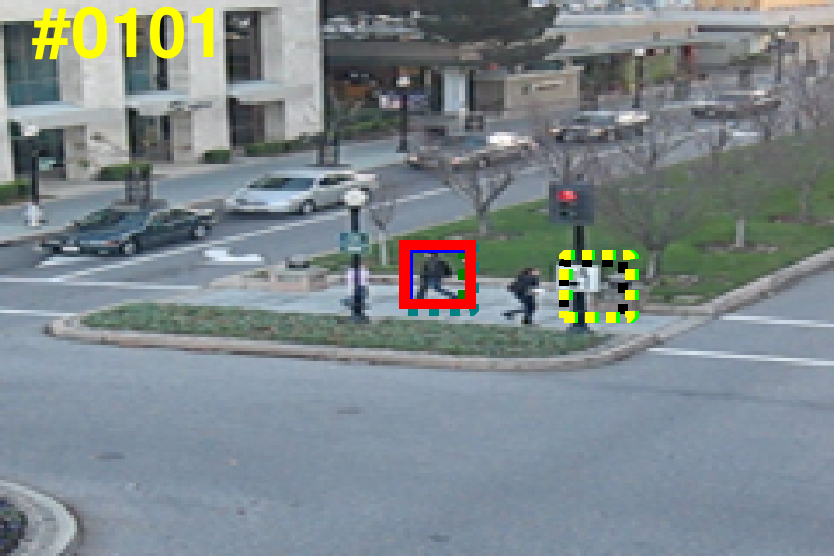}
\includegraphics[width=0.16\linewidth]{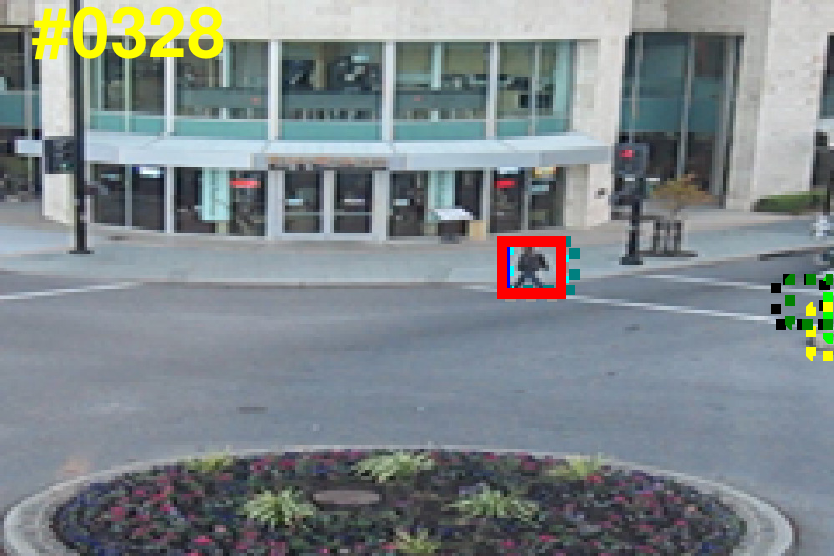}
\includegraphics[width=0.16\linewidth]{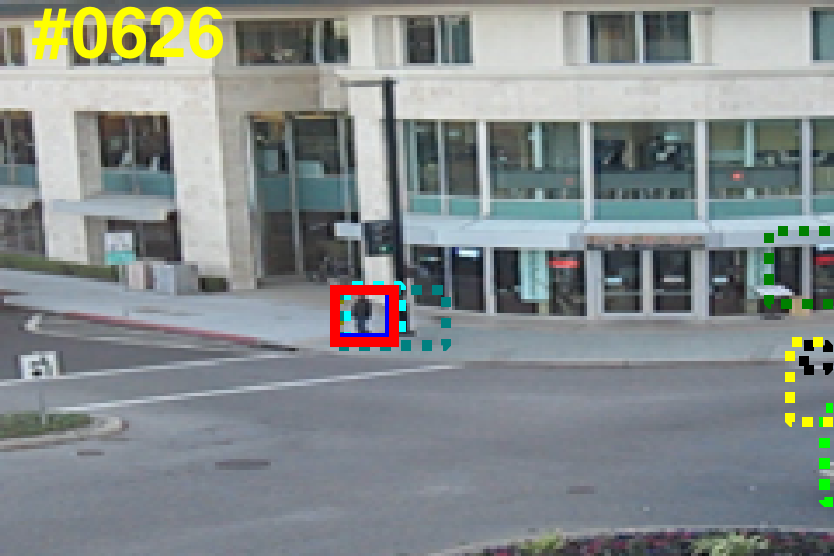}\\
\includegraphics[width=0.16\linewidth]{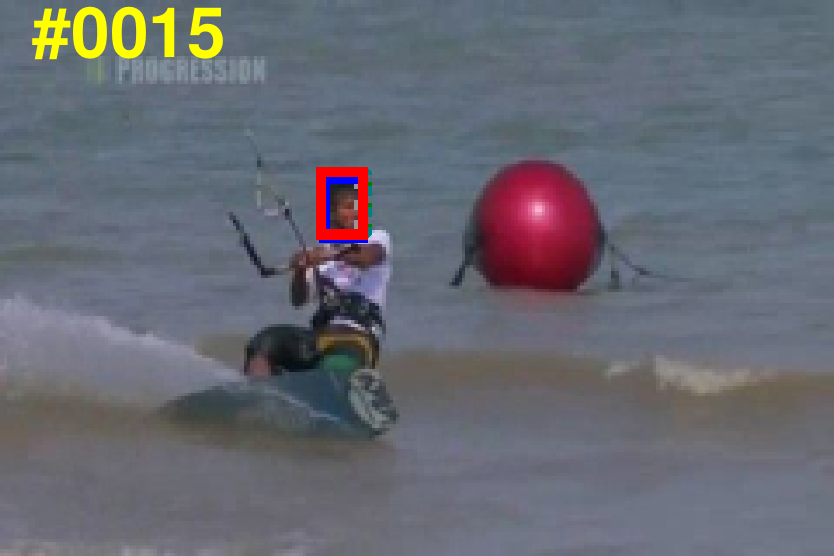}
\includegraphics[width=0.16\linewidth]{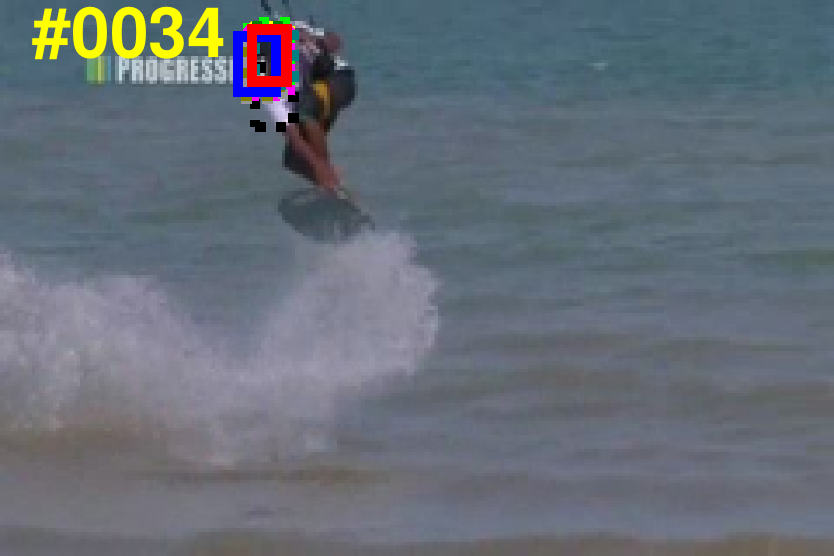}
\includegraphics[width=0.16\linewidth]{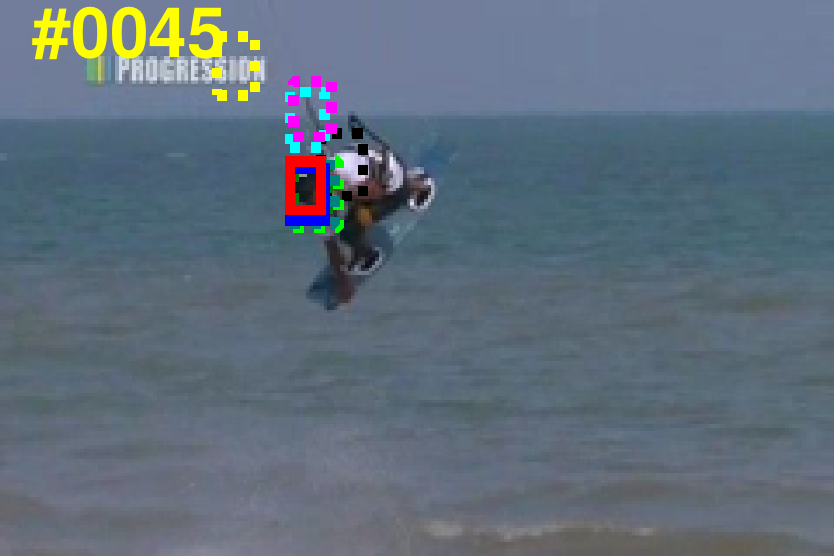}
\includegraphics[width=0.16\linewidth]{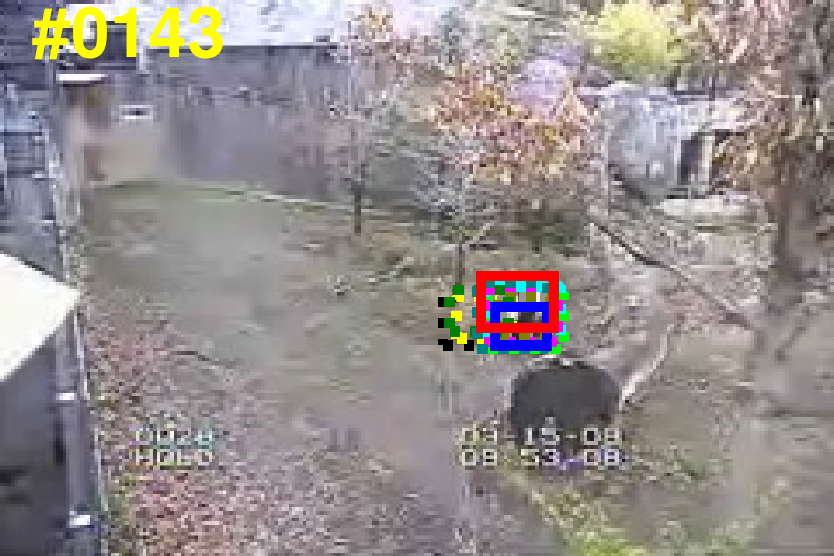}
\includegraphics[width=0.16\linewidth]{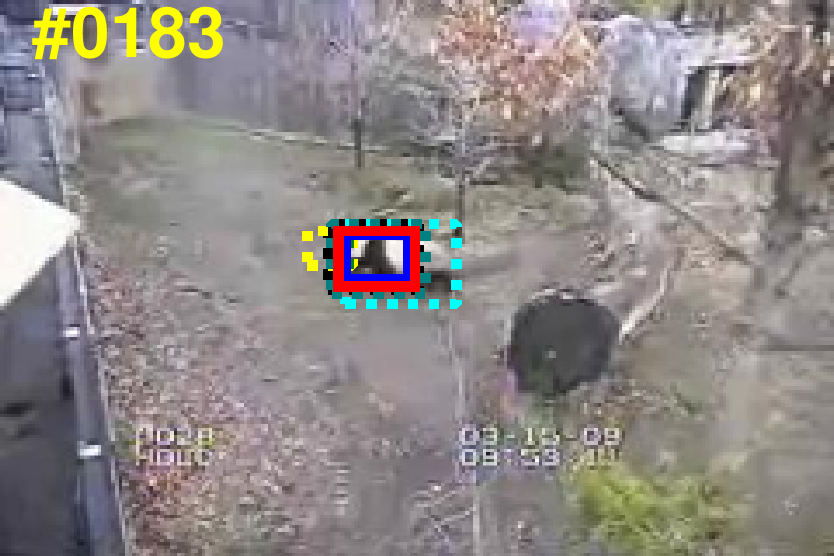}
\includegraphics[width=0.16\linewidth]{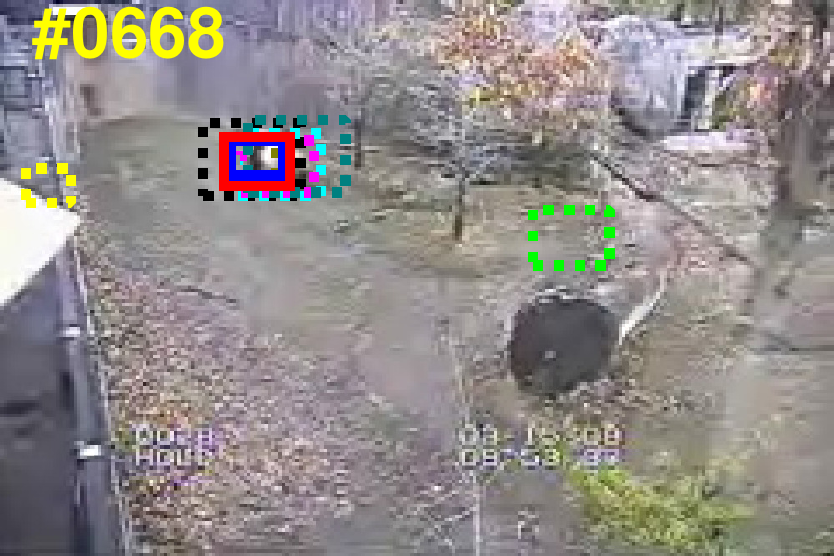}\\
\includegraphics[width=0.16\linewidth]{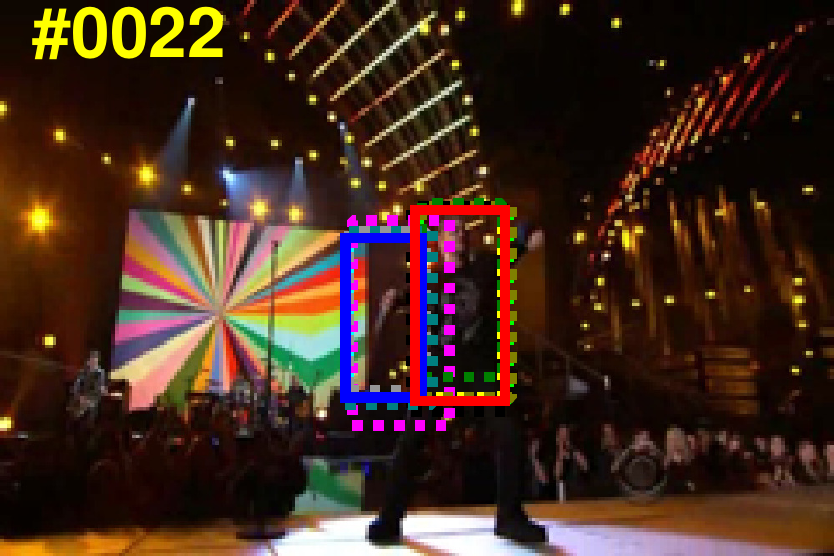}
\includegraphics[width=0.16\linewidth]{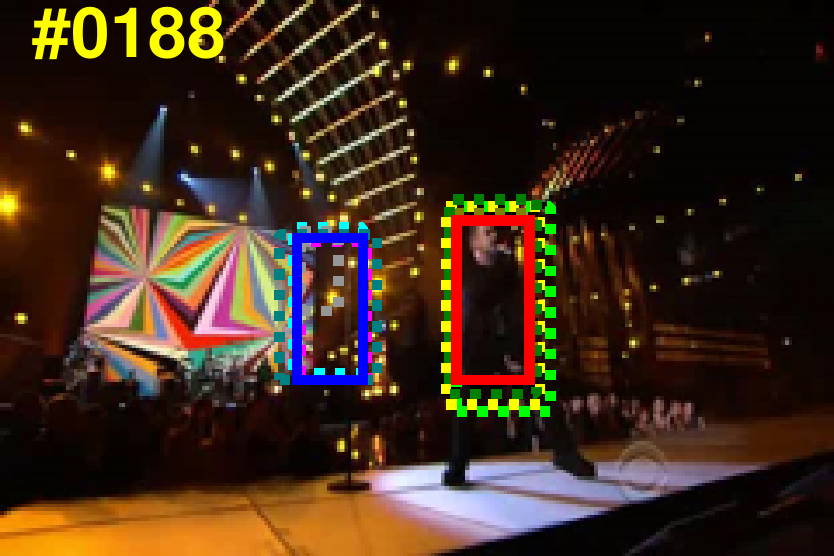}
\includegraphics[width=0.16\linewidth]{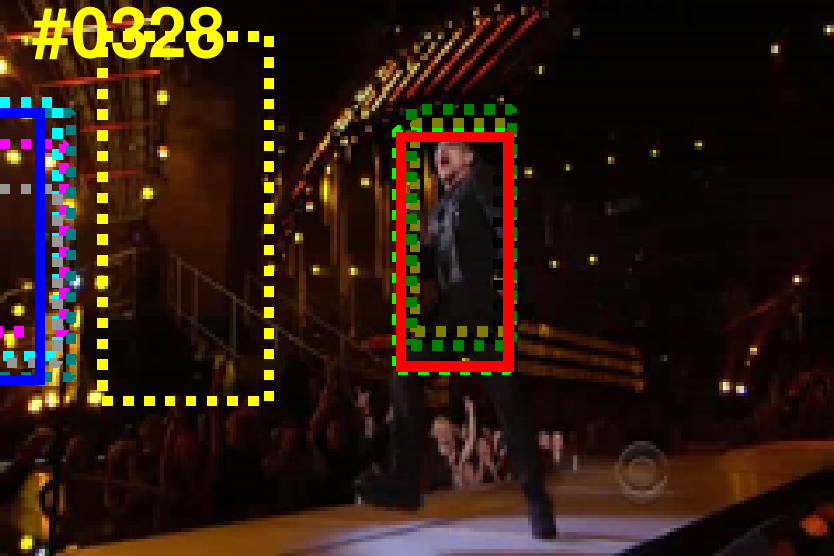}
\includegraphics[width=0.16\linewidth]{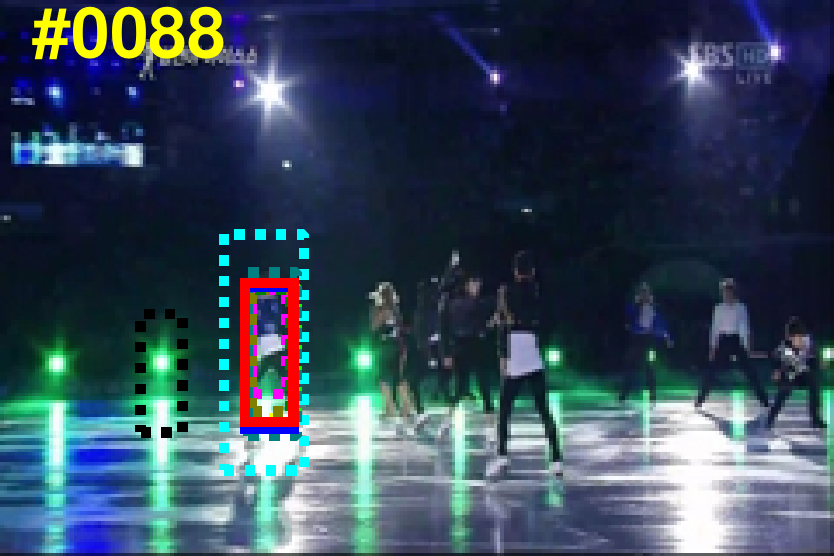}
\includegraphics[width=0.16\linewidth]{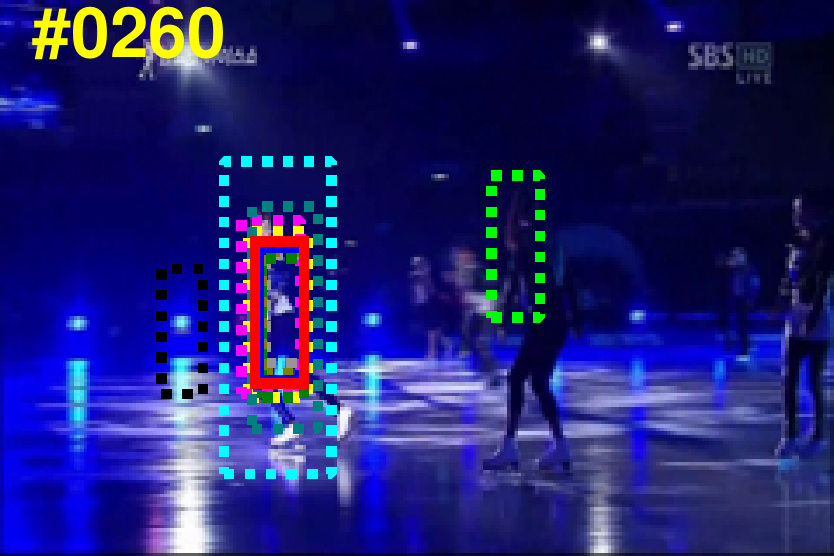}
\includegraphics[width=0.16\linewidth]{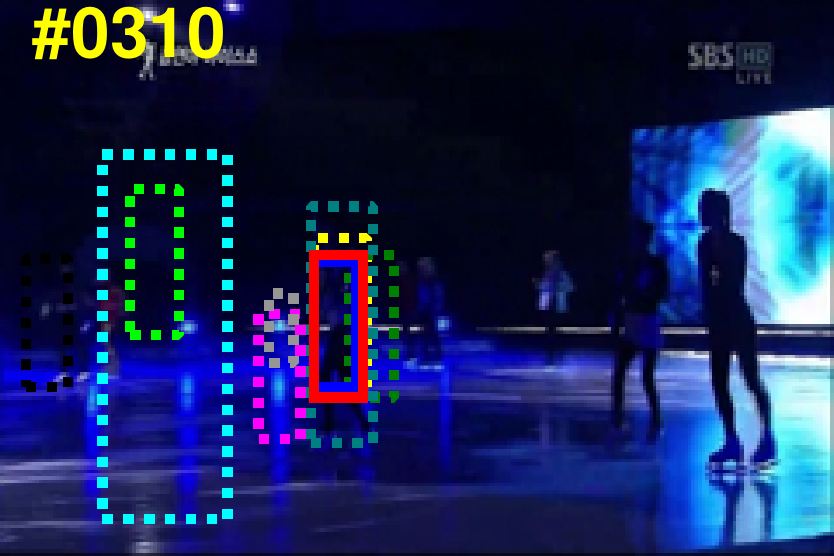}\\
\includegraphics[width=0.16\linewidth]{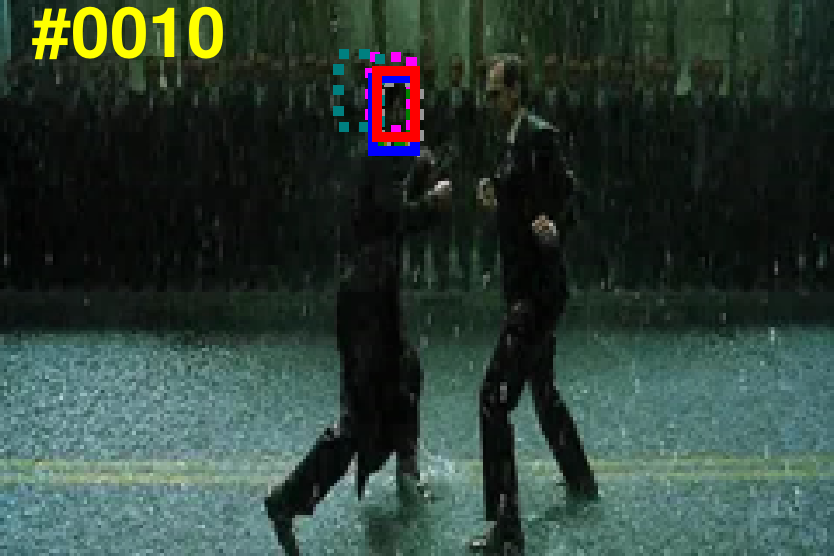}
\includegraphics[width=0.16\linewidth]{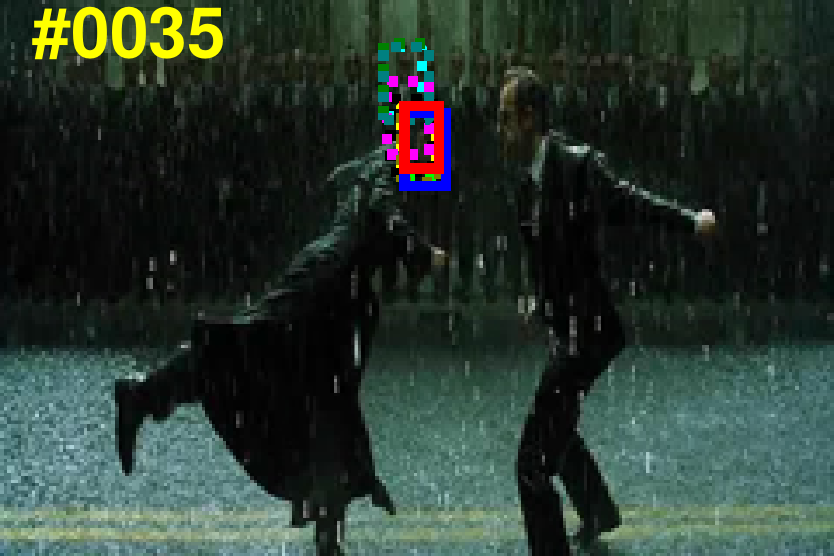}
\includegraphics[width=0.16\linewidth]{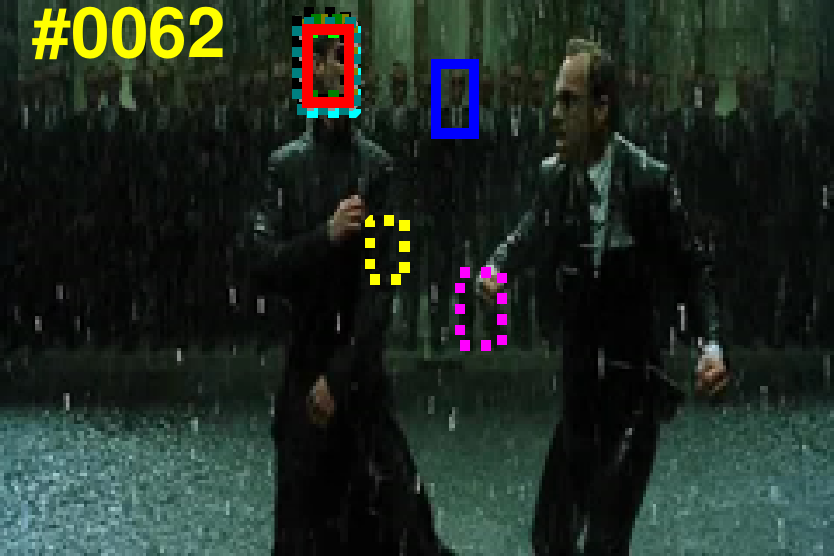}
\includegraphics[width=0.16\linewidth]{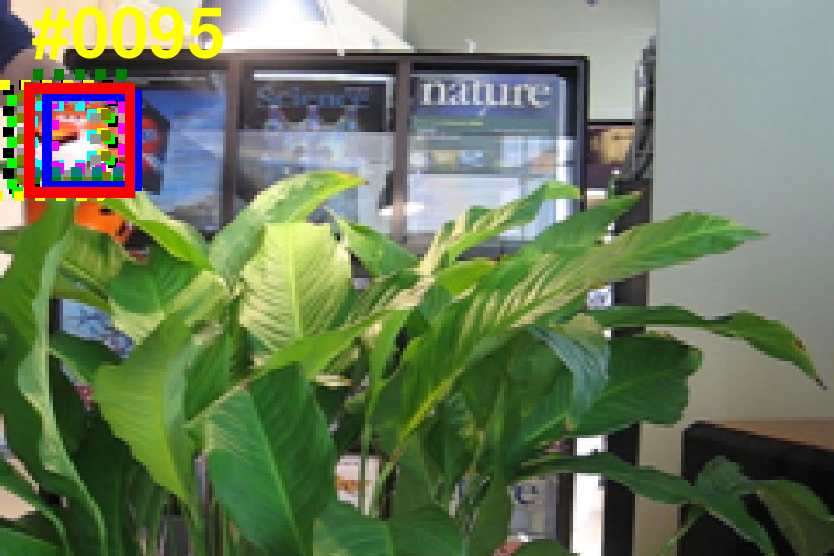}
\includegraphics[width=0.16\linewidth]{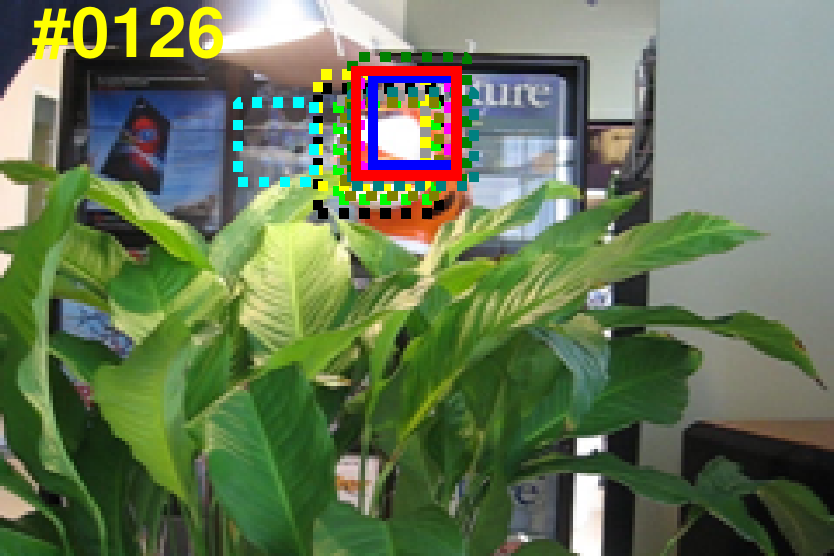}
\includegraphics[width=0.16\linewidth]{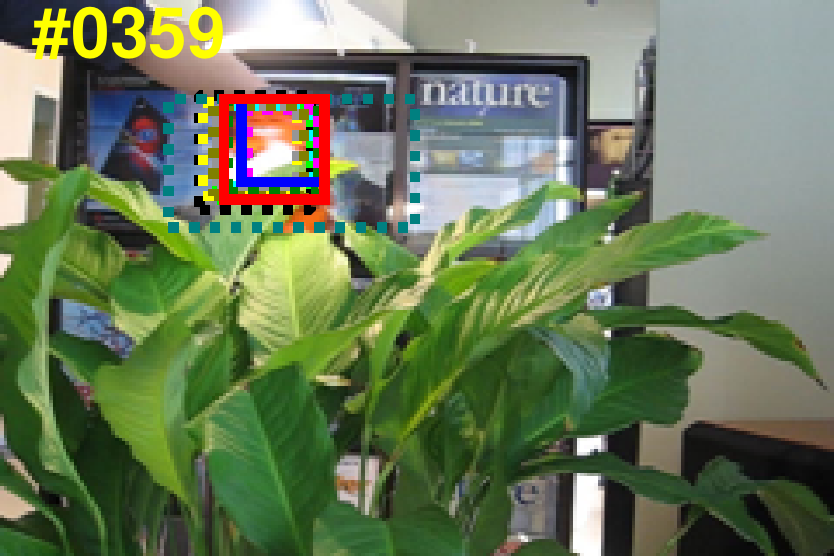}\\
\includegraphics[trim={0mm 123mm 0mm 0mm},clip,width=0.9\linewidth]{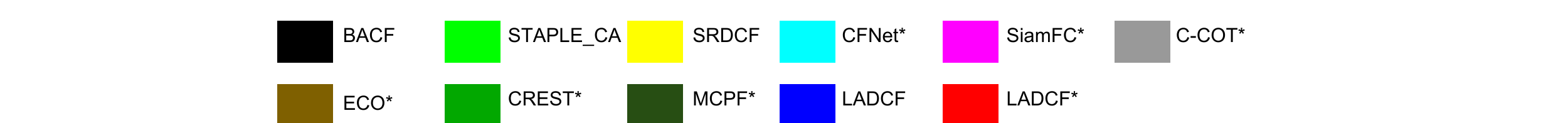}
\end{center}
\caption{Illustration of qualitative tracking results on challenging sequences (Left column top to down: \textit{Biker}, \textit{Board}, \textit{Bolt}, \textit{Coke}, \textit{Girl2}, \textit{Kitesurf}, \textit{Singer2} and \textit{Matrix}. Right column top to down: \textit{Bird1}, \textit{Bolt2}, \textit{Box}, \textit{Dragonbaby}, \textit{Human3}, \textit{Panda}, \textit{Skating1} and \textit{Tiger2}). The colour bounding boxes are the corresponding results of BACF, STAPLE\_CA, SRDCF, CFNet$^\ast$, SiamFC$^\ast$, C-COT$^\ast$, ECO$^\ast$, CREST$^\ast$, MCPF$^\ast$, LADCF and LADCF$^\ast$, respectively.}\label{qualitative}
\end{figure*}

We further evaluate the proposed LADCF method in terms of the OP metric on the OTB2013, OTB50 and OTB100 datasets. 
The results comparing with a number of state-of-the-art trackers are reported in Table~\ref{otbophc} and Table~\ref{otbopdeep}, using hand-crafted and deep features, respectively. 
Among the trackers equipped with hand-crafted features, LADCF achieves the best results with absolute gains of $1\%$, $4.1\%$ and $3.3\%$ respectively over the second best one on the three datasets. 
The comparison with deep features/structures based trackers also supports the superiority of LADCF$^\ast$ over the state-of-the-art approaches.
LADCF$^\ast$ performs better than ECO$^\ast$ by $2\%$, $1.5\%$ and $1.8\%$ on the OTB2013, OTB50 and OTB100 datasets.

We also present the success plots on OTB2013, OTB50, Temple-Colour and UAV123 in Fig.~\ref{successplot4sets}.
We focus on the success plots here as overlaps are more important for evaluating the tracking performance. 
As shown in the figure, our LADCF beats all the other methods using hand-crafted trackers on these four datasets.
LADCF outperforms recent trackers, \ie CSRDCF (by $6.8\%\sim 9.1\%$), STAPLE\_CA (by $4.6\%\sim 7.5\%$), C-COT (by $2.2\%\sim 4.7\%$), BACF (by $1.8\%\sim 6.9\%$) and ECO (by $0.5\%\sim 3.8\%$). 
In addition, LADCF$^\ast$ is better than ECO$^\ast$ and C-COT$^\ast$ except on UAV123. 
Specifically, LADCF$^\ast$ achieves an AUC score of $60.6\%$ on Temple-Colour, which is better than ECO$^\ast$ and C-COT$^\ast$ by $2.9\%$ and $3.3\%$, respectively. 
We find our consistent feature selection model is particularly useful on Temple-Colour, because all the sequences in Temple-Colour are of colour format, that is more suitable for Colour-Names and CNN features.
However, ECO$^\ast$ leads with $1\%$ above LADCF$^\ast$ on UAV123. 
A plausible reason is that the average number of frames per sequence in UAV123 is 915. 
This is much higher than OTB2013 (578), OTB50 (591) and Temple-Colour (429). 
Our consistent feature selection model only stores the filter model $\bm{\theta}_{\bm{\textrm{model}}}$ from the previous frame, while ECO$^\ast$ is more sophisticated to deal with long-term tracking by collecting clusters of historical training samples during tracking.  

{\color{black}{We report the evaluation results of the proposed LADCF method on VOT2018 in Table~\ref{vot18}.
LADCF$^\ast$ equipped with VGG features achieves a better EAO score, 0.338, compared to ECO$^\ast$, CFCF$^\ast$ and CFWCR$^\ast$.
In addition, the EAO score of the proposed LADCF method equipped with ResNet-50~\cite{he2015deep}, which is also used in UPDT$^\ast$ and MFT$^\ast$, outperforms all the other trackers, demonstrating the effectiveness of the proposed feature selection framework.}}

\noindent \textbf{Quantitative tracking performance results  on sequence attributes:}
We provide the tracking results parameterised by 11 attributes on OTB100 in Fig.~\ref{attributes}. 
Our LADCF$^\ast$ outperforms other trackers in 5 attributes, \ie out of view, in-plane rotation, deformation, scale variation and out-of-plane rotation. 
Our consistent feature selection embedded appearance model enables adaptive spatial layout recognition, focusing on the relevant target and background regions with shared motion properties to create a robust complementary tracking pattern. 
The results of LADCF$^\ast$ in the other 6 attributes are among the top 3, demonstrating the effectiveness and robustness of our method. 
In addition, the superiority of LADCF is more obvious as it achieves the best performance in 9 attributes, compared to the other hand-crafted feature trackers. 
In particular, the performance of LADCF exhibits significant gains  ($4.4\%$, $3.2\%$ and $3.5\%$ as compared with the second best one in the attributes of low resolution, out of view and occlusion.

\noindent \textbf{Qualitative tracking performance results:}
Fig.~\ref{qualitative} shows the qualitative results of the state-of-the-art methods, \ie BACF, STAPLE\_CA, SRDCF, CFNet$^\ast$, SiamFC$^\ast$, C-COT$^\ast$, ECO$^\ast$, CREST$^\ast$, MCPF$^\ast$ as well as our LADCF and LADCF$^\ast$, on some challenging video sequences.
The difficulties are posed by rapid changes in the appearance of targets. 
Our LADCF and LADCF$^\ast$ perform well on these challenges as we employ consistent embedded feature selection to identify the pertinent spatial layout. 
Sequences with deformations (\textit{Bolt}, \textit{Dragonbaby}) and out of view (\textit{Biker}, \textit{Bird1}) can be successfully tracked by our methods without any failures. Videos with occlusions (\textit{Board}, \textit{Girl2}, \textit{Human3}, \textit{Tiger2}) also benefit from our strategy of employing temporal consistency. Specifically, LADCF and LADCF$^\ast$ are expert in solving in-plane and out-of-plane rotations (\textit{Coke}, \textit{Dragonbaby}, \textit{Skating1}), because the proposed adaptive spatial regularisation approach provides a novel solution to fusing the appearance information from the central region and surroundings.

\subsection{Self Analysis}
In this part, we provide a deep analysis to each component of our proposed LADCF method, \ie feature configurations, temporal consistency and feature selection.
\begin{table}[!t]
\footnotesize
\renewcommand{\arraystretch}{1.1}
\caption{Tracking performance on OTB100 with different feature configurations.}
\label{feature_config}
\centering
\begin{tabular}{clc}
\hline
 \multicolumn{2}{c}{Features}  & AUC score \\
\hline
 \multirow{2}{*}{Hand-crafted}   & HOG &  64.30\%\\
& HOG+CN & 66.44\% \\
\hline 
 \multirow{5}{*}{Hand-crafted+CNN} &HOG+CN+Conv-1 & 66.72\%\\
 &HOG+CN+Conv-2 & 66.99\%\\
  &HOG+CN+Conv-3 & 69.65\%\\
 &HOG+CN+Conv-4 & 69.72\%\\
  &HOG+CN+Conv-5 & 68.15\%\\
\hline
\end{tabular}
\end{table}

First, we employ 7 feature configurations to test our model using AUC metric on OTB100. As shown in Table~\ref{feature_config}, LADCF achieves $2.1\%$ improvement by combining Colour-Names with HOG. The middle convolutional layers (Conv-3 and Conv-4) significantly improve the performance, compared with low (Conv-1 and Conv-2) and high layers (Conv-5).
\begin{figure}[!t]
\label{impact_alpha_r}
\centering
\subfloat[learning rate]{
\label{Impact_of_alpha}
\includegraphics[width=0.9\linewidth]{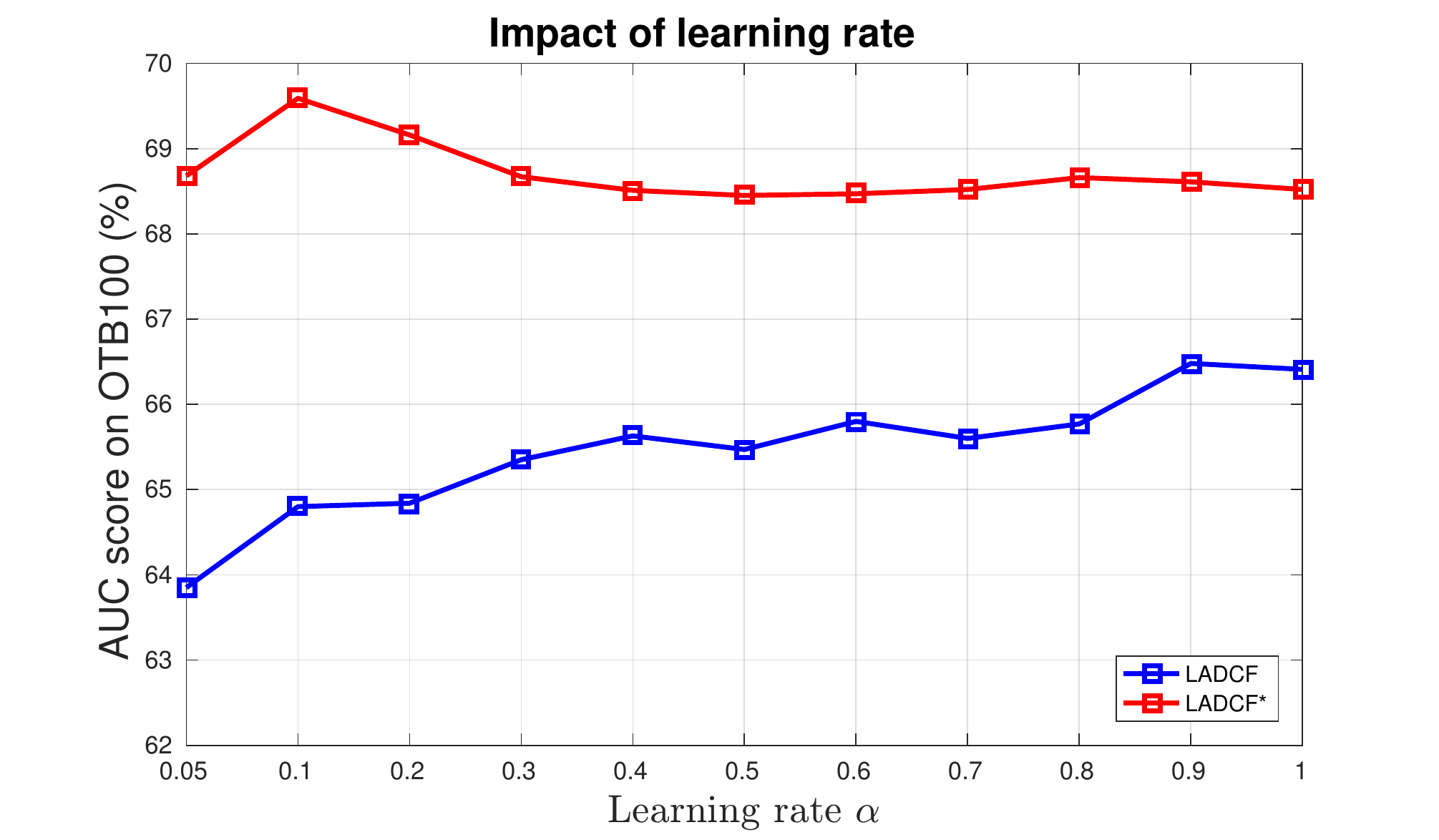}
}
\\
\subfloat[feature selection ratio]{
\label{Impact_of_r}
\includegraphics[width=0.9\linewidth]{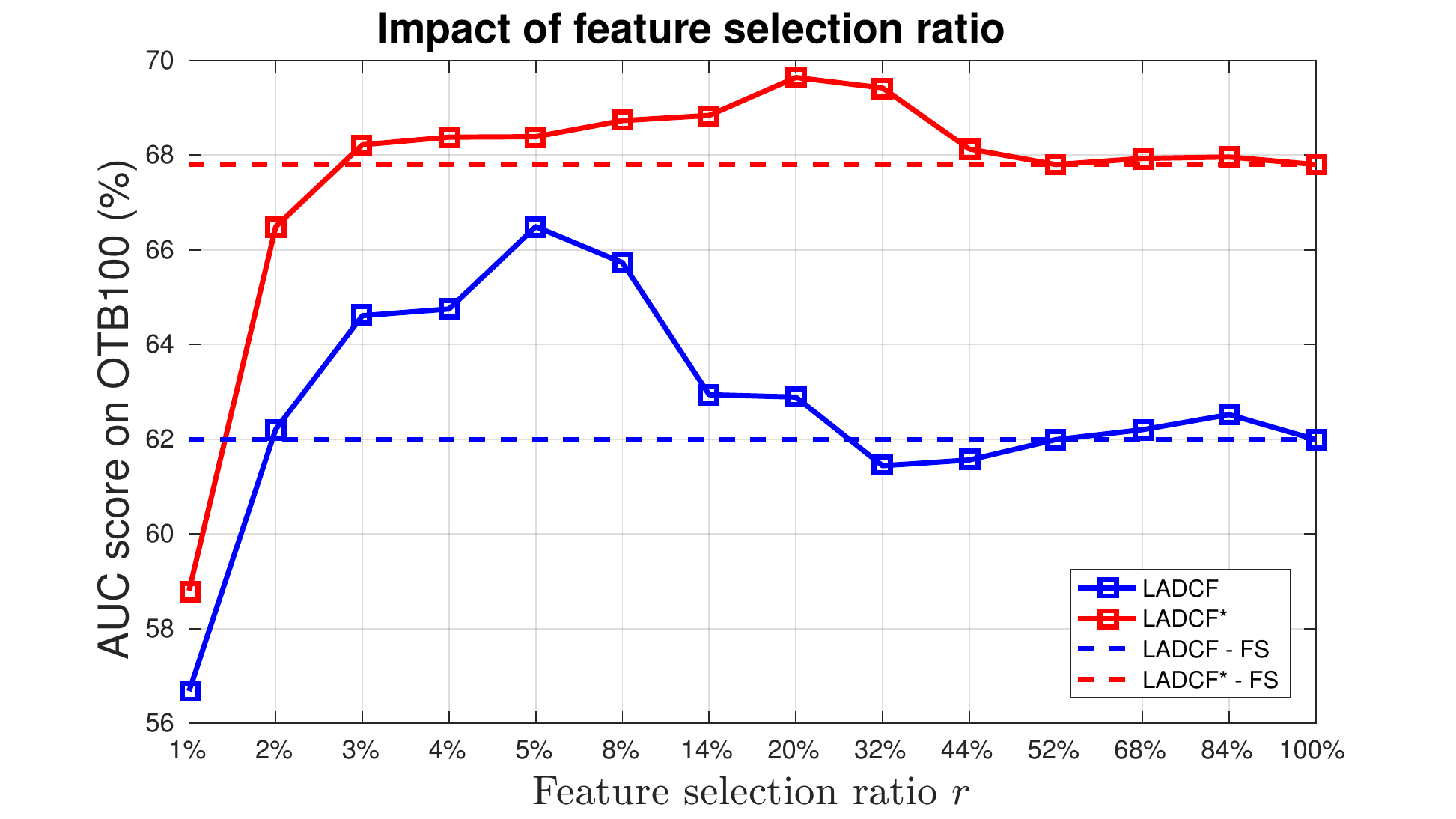}
}
\caption{The experimental results obtained by our temporal consistency preserving spatial feature selection enhanced appearance model on OTB100 for (a) different learning rates and (b) different feature selection ratios.}
\end{figure}

Second, we analyse the sensitivity of the algorithm to the learning rate $\alpha$ and feature selection ratio $r$. As shown in Fig.~\ref{Impact_of_alpha}, the tracking results vary smoothly with respect to the learning rate $\alpha$,  demonstrating that our method achieves stable performance with the proposed temporal consistency by forcing the learned filters to be instantiated in a low-dimensional manifold space to reflect diversity and preserve the filter generalisation capacity. 
{\color{black}{In addition, we analyse the impact of our temporal consistency component on OTB100 in Table~\ref{temp_config}.
LADCF / LADCF$^\ast$ achieve AUC improvement of $3.1\%/1.8\%$ and DP improvement of $2.9\%/2.4\%$ respectively, demonstrating its merit in the proposed feature selection method.}}

%\begin{figure}[!t]
%\begin{center}
%   \includegraphics[width=0.9\linewidth]{./img/Impact_of_r.pdf}
%\end{center}
%   \caption{The experimental results obtained by our temporal consistency preserving spatial feature selection enhanced appearance model on OTB100 for different feature selection ratios.}
%\end{figure}

\begin{table}[!t]
\footnotesize
\renewcommand{\arraystretch}{1.1}
\caption{Tracking performance on OTB100 in terms of temporal consistency.}
\label{temp_config}
\centering
\begin{tabular}{c|cc|cc}
\hline
 &   \multicolumn{2}{c|}{AUC score} &  \multicolumn{2}{c}{DP score} \\
\hline
 temporal consistency & \ding{51} & \ding{55} & \ding{51} & \ding{55} \\
\hline 
LADCF & 66.4\% & 63.3\% & 86.4\% & 83.5\% \\
LADCF$^\ast$ & 69.6\% & 67.8\% & 90.6\% & 88.2\% \\
\hline
\end{tabular}
\end{table}

{\color{black}{Third}}, we analyse the impact of the feature selection ratio $r$ in Fig.~\ref{Impact_of_r}. 
The dash-dotted lines (LADCF-FS, LADCF$^\ast$-FS) denote the corresponding results without feature selection. It cannot be overemphasised that hand-crafted and deep features achieve impressive improvements with the selection ratios ranging from $2\%\sim 20\%$ and $3\%\sim 40\%$ respectively. 
The results support the conclusion that the tracking performance can be improved by using the proposed feature selection embedded filter learning scheme.

\section{Conclusion}\label{conclusion}
We proposed an effective temporal consistency preserving spatial feature selection embedded approach to realise real-time visual object tracking with outstanding performance. By reformulating the appearance learning model with embedded feature selection and imposing temporal consistency, we achieve adaptive discriminative filter learning on a low dimensional manifold with enhanced interpretability. Both hand-crafted and deep features are considered in our multi-channel feature representations. 
The extensive experimental results on tracking benchmark datasets demonstrate the effectiveness and robustness of our method, compared with state-of-the-art trackers.

%% References


% or


%\appendices

%\section{}

\section*{Acknowledgment}
This work was supported in part by the EPSRC Programme Grant (FACER2VM) EP/N007743/1, EPSRC/dstl/MURI project EP/R018456/1, the National Natural Science Foundation of China (61373055, 61672265, 61876072, 61602390) and the NVIDIA GPU Grant Program.

\ifCLASSOPTIONcaptionsoff
  \newpage
\fi



\bibliographystyle{IEEEtran}
\bibliography{sample}
%\begin{thebibliography}{1}

\newpage
\begin{IEEEbiography}[{\includegraphics[width=1in,height=1.25in,clip,keepaspectratio]{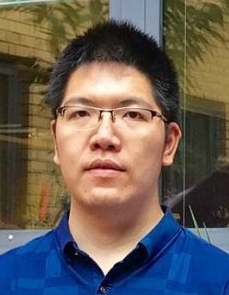}}]{Tianyang Xu} received the B.Sc. degree in electronic engineering from the Nanjing University, Nanjing, China, in 2011. He is a PhD student at the School of Internet of Things Engineering, Jiangnan University, Wuxi, China. He is currently a visiting PhD student at the Centre for Vision, Speech and Signal Processing (CVSSP), University of Surrey, United Kingdom. His research interests include computer vision and machine learning.
\end{IEEEbiography}

\begin{IEEEbiography}[{\includegraphics[width=1in,height=1.25in,clip,keepaspectratio]{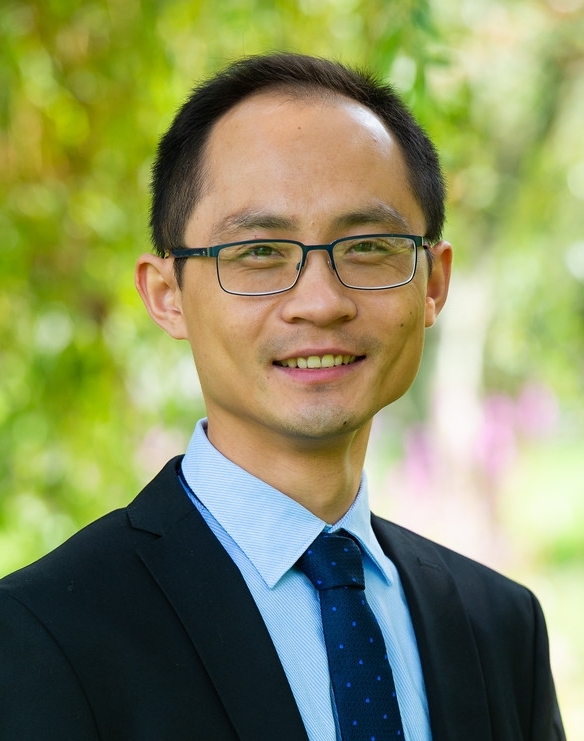}}]{Zhen-Hua Feng} (S'13-M'16) received the Ph.D. degree from the Centre for Vision, Speech and Signal Processing, University of Surrey, U.K. in 2016. He is currently a research fellow at the University of Surrey. His research interests include pattern recognition, machine learning and computer vision.

He has published more than 30 scientific papers in top-ranked conferences and journals, including IEEE Conference on Computer Vision and Pattern Recognition, IEEE Transactions on Image Processing, IEEE Transactions on Cybernetics, IEEE Transactions on Information Forensics and Security, Pattern Recognition, Information Sciences etc. He has received the 2017 European Biometrics Industry Award from the European Association for Biometrics (EAB) and the AMDO 2018 Best Paper Award for Commercial Applications.
\end{IEEEbiography}

\begin{IEEEbiography}[{\includegraphics[width=1in,height=1.25in,clip,keepaspectratio]{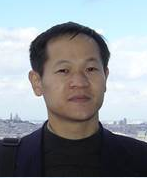}}]{Xiao-Jun Wu} received the B.Sc. degree in mathematics from Nanjing Normal University, Nanjing, China, in 1991. He received the M.S. degree and the Ph.D. degree in pattern recognition and intelligent systems from Nanjing University of Science and Technology, Nanjing, China, in 1996 and 2002, respectively.

He is currently a Professor in artificial intelligent and pattern recognition at the Jiangnan University, Wuxi, China. His research interests include pattern recognition, computer vision, fuzzy systems, neural networks and intelligent systems.
\end{IEEEbiography}

\begin{IEEEbiography}[{\includegraphics[width=1in,height=1.25in,clip,keepaspectratio]{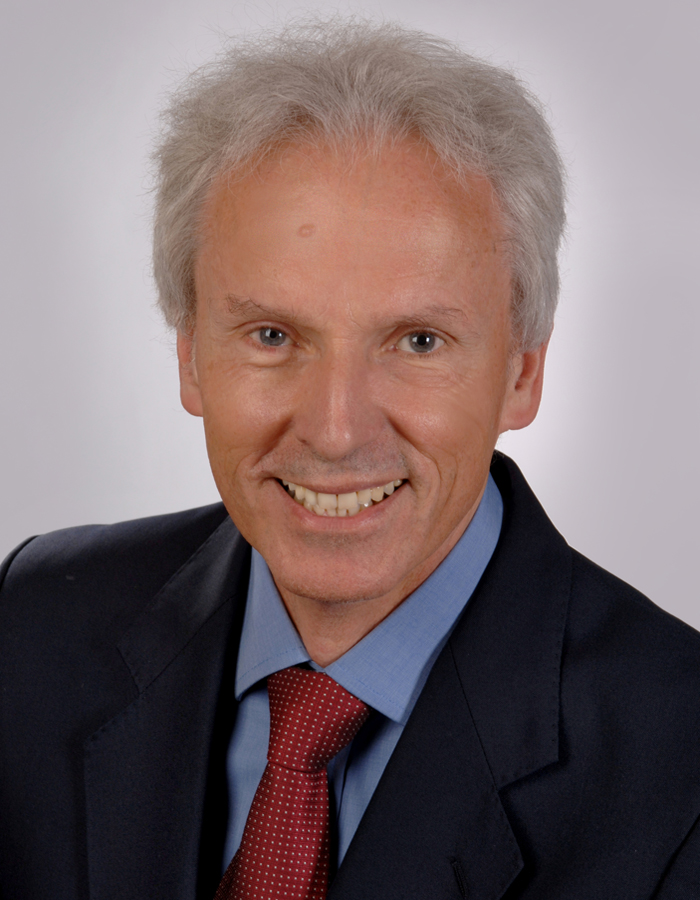}}]{Josef Kittler}(M'74-LM'12) received the B.A., Ph.D., and D.Sc. degrees from the University of Cambridge, in 1971, 1974, and 1991, respectively. He is a distinguished Professor of Machine Intelligence at the Centre for Vision, Speech and Signal Processing, University of Surrey, Guildford, U.K. He conducts research in biometrics, video and image database retrieval, medical image analysis, and cognitive vision. He published the textbook Pattern Recognition: A Statistical Approach and over 700 scientific papers. His publications have been cited more than 60,000 times (Google Scholar).

He is series editor of Springer Lecture Notes on Computer Science. He currently serves on the Editorial Boards of Pattern Recognition Letters, Pattern Recognition and Artificial Intelligence, Pattern Analysis and Applications. He also served as a member of the Editorial Board of IEEE Transactions on Pattern Analysis and Machine Intelligence during 1982-1985. He served on the Governing Board of the International Association for Pattern Recognition (IAPR) as one of the two British representatives during the period 1982-2005, President of the IAPR during 1994-1996.
\end{IEEEbiography}


%\end{thebibliography}


%\end{IEEEbiography}


%\newpage



%\vfill


\end{document}